\documentclass[preprint, 12pt, authoyear, abbrvbib]{article}
\usepackage{jmlr2e} 

\usepackage[english]{babel}
\usepackage[latin1]{inputenc}
\usepackage{amsmath,amssymb,latexsym,amsfonts,thm-restate}
\usepackage{algorithm}
\usepackage{algpseudocode}
\usepackage{bm}
\usepackage{bbm}
\usepackage{hyperref} 
\usepackage{mathtools}
\usepackage{mathrsfs}
\usepackage{graphics}
\usepackage{todonotes}
\usepackage{soul}

\usepackage{color}
\usepackage{float}
\usepackage{floatpag}
\floatpagestyle{empty}
\usepackage{booktabs}

\usepackage{graphicx} \usepackage{verbatim}
\usepackage[font=footnotesize,labelfont=footnotesize]{caption}
\usepackage[font=scriptsize,labelfont=scriptsize]{subcaption}
\usepackage{enumerate}
\usepackage{tabulary}
\usepackage{multirow}
\usepackage{adjustbox}
\allowdisplaybreaks
\usepackage[bb=boondox]{mathalfa}

\DeclareMathOperator*{\esssup}{esssup}
\DeclareMathOperator*{\dist}{dist}
\DeclareMathOperator*{\grad}{\nabla} 
\DeclareMathOperator*{\Hess}{Hess}

\newcommand{\divv}{\mathrm{div}}

\newcommand{\Ex}{\mathbb{E}}

\newcommand{\gmin}{g_{\min}}
\newcommand{\gmax}{g_{\max}}
\renewcommand{\L}{\mathcal{L}}
\newcommand{\Length}{T}

\newcommand{\M}{\mathcal{M}}
\newcommand{\Prob}{\mathbb{P}}
\newcommand{\R}{\mathbb{R}}
\newcommand{\reach}{\mathcal{R}}
\newcommand{\V}{\text{dVol}}
\newcommand{\T}{\log}
\newcommand{\X}{\mathcal{X}}

\makeatletter
\newcommand{\distas}[1]{\mathbin{\overset{#1}{\kern\z@\sim}}}%
\newsavebox{\mybox}\newsavebox{\mysim}
\newcommand{\distras}[1]{%
	\savebox{\mybox}{\hbox{\kern3pt$\scriptstyle#1$\kern3pt}}
	\savebox{\mysim}{\hbox{$\sim$}}%
	\mathbin{\overset{#1}{\kern\z@\resizebox{\wd\mybox}{\ht\mysim}{$\sim$}}}%
}
\makeatother

\newtheorem{thm}{Theorem}[section]

\newtheorem{prop}[thm]{Proposition}
\newtheorem{lem}[thm]{Lemma}

\newtheorem{rems}[thm]{Remark}

\newtheorem{defn}[thm]{Definition}

\newtheorem{assump}[thm]{Assumption}

\newcommand{\red}{\color{red}}

\definecolor{mygreen}{rgb}{0.1,0.75,0.2}

\newcommand{\nc}{\normalcolor}

\begin{document}

\title{Fermat Distances: Metric Approximation, \\Spectral Convergence, and Clustering Algorithms}

 \author{\name Nicol\'as Garc\'ia Trillos \email garciatrillo@wisc.edu \\
       \addr Department of Statistics\\
       University of Wisconsin\\
       Madison, WI 53706, USA 
\AND
\name Anna Little \email little@math.utah.edu \\ \addr Department of Mathematics, Utah Center for Data Science \\ University of Utah \\
Salt Lake City, UT 84112, USA 
\AND 
\name Daniel McKenzie \email dmckenzie@mines.edu \\ \addr Department of Applied Mathematics and Statistics \\ Colorado School of Mines \\ 
Golden, CO 80401, USA 
\AND 
\name James M. Murphy \email jm.murphy@tufts.edu\\ \addr Department of Mathematics\\
       Tufts University\\
       Medford, MA 02155, USA}

 \editor{}
\maketitle	

\begin{abstract}

We analyze the convergence properties of \emph{Fermat distances}, a family of density-driven metrics defined on Riemannian manifolds with an associated probability measure.  Fermat distances may be defined either on discrete samples from the underlying measure, in which case they are random, or in the continuum setting, in which they are induced by geodesics under a density-distorted Riemannian metric.  We prove that discrete, sample-based Fermat distances converge to their continuum analogues in small neighborhoods with a precise rate that depends on the intrinsic dimensionality of the data and the parameter governing the extent of density weighting in Fermat distances.  This is done by leveraging novel geometric and statistical arguments in percolation theory that allow for non-uniform densities and curved domains.  Our results are then used to prove that discrete graph Laplacians based on discrete, sample-driven Fermat distances converge to corresponding continuum operators.  In particular, we show the discrete eigenvalues and eigenvectors converge to their continuum analogues at a dimension-dependent rate, which allows us to interpret the efficacy of discrete spectral clustering using Fermat distances in terms of the resulting continuum limit.  The perspective afforded by our discrete-to-continuum Fermat distance analysis leads to new clustering algorithms for data and related insights into efficient computations associated to density-driven spectral clustering.  Our theoretical analysis is supported with numerical simulations and experiments on synthetic and real image data.

\end{abstract}

\section{Introduction}

Data-driven metrics and related dimensionality reduction methods are a widely used tool in statistics, data science, and machine learning for analyzing point cloud data $\X\subset\mathbb{R}^{D}$.  Seminal methods such as principal \citep{hotelling1933analysis} and independent component analysis \citep{comon1994independent}, Laplacian eigenmaps \citep{Belkin2003laplacian}, diffusion maps \citep{Coifman2005_Geometric, Coifman2006diffusion}, Isomap \citep{Tenenbaum2000global}, locally linear embedding \citep{roweis2000nonlinear}, and tSNE \citep{Maaten2008visualizing} often capture important structural properties in data (e.g., cluster structure, concentration near low-dimensional sets) in a manner that is statistically efficient and robust to noise and outliers.  In many cases, these methods first embed high-dimensional, noisy data into a low-dimensional Euclidean space so that Euclidean distances on the embedded points implicitly define a new metric.  Working in these embedded spaces (or equivalently, analyzing with the induced metrics) has led to methods for unsupervised and semisupervised machine learning \citep{Ng2002}.  

An alternative to this broad class of approaches is to consider weighted shortest path metrics.  The idea is to construct a weighted graph associated to $\X$, and then use shortest path distances in this graph to learn important structures in $\X$.  A particular class of density-weighted path metrics known as \emph{Fermat distances} use powers of the Euclidean distance between points as weights \citep{Bijral2011_Semi, Hwang2016_Shortest, chu2020exact, little2022balancing, Groisman2022nonhomogeneous, fernandez2023intrinsic}; in the discrete setting, we denote these distances as $\ell_{p}$, where $p$ is a parameter that determines the impact of data density.  
Fermat distances and related density-driven path metrics have been successfully applied to a range of problems in unsupervised and semisupervised machine learning \citep{Vincent2003_Density, Bousquet2004_Measure, Sajama2005_Estimating, Chang2008_Robust, Bijral2011_Semi, Moscovich2017_Minimax, alamgir2012shortest, mckenzie2019power, Little2020path}, as well as to topological data analysis for robust computation of persistent homology \citep{fernandez2023intrinsic} and in high-dimensional signal processing \citep{zhang2021_Hyperspectral, manousidaki2021clustering}.
When data points are sampled from a compact Riemannian manifold $\mathcal{M}$, the discrete $\ell_{p}$ converges to a continuum metric $\L_{p}$, which can be interpreted as a density-weighted geodesic distance on $\mathcal{M}$. 

Existing works in the literature establishing discrete-to-continuum convergence of general data-driven distances include \citet{howard2001geodesics, DiazetAl,DavisSethuraman, Hwang2016_Shortest,bungert2022ratio}, among many others. The papers \cite{DiazetAl,DavisSethuraman,bungert2022ratio}, for example, discuss convergence of distances defined on random geometric graphs (RGG), either in the i.i.d. setting or for Poisson point processes. In the RGG setting, admissible paths between two points must consist of consecutive short range (as specified by a connectivity parameter) hops between data, in contrast to Fermat distances, where arbitrarily large hops between points are admissible. The results from \cite{DavisSethuraman} are asymptotic, while the ones in \cite{DiazetAl,bungert2022ratio} provide high probability convergence rates in terms of the RGG's connectivity parameter. The results in \cite{bungert2022ratio}, for example, discuss the convergence of the ratio between certain expectations of distances at different scales. When combined with concentration inequalities, this allows the authors to prove rates of convergence, in sparse settings, for a semisupervised learning procedure known as Lipschitz learning. 
The works \cite{kesten1993speed, howard2001geodesics, Hwang2016_Shortest,Groisman2022nonhomogeneous, little2022balancing, fernandez2023intrinsic} study Fermat distances on point clouds and are the most relevant references for our metric approximation results. Note \cite{Hwang2016_Shortest,Groisman2022nonhomogeneous, fernandez2023intrinsic} only provide asymptotic convergence results, while \cite{howard2001geodesics, little2022balancing} assume a uniform density.  We provide the first local quantitative convergence results for Fermat distance in the manifold setting with a general density.

While there are several mathematical objects that can be constructed over these finite data-driven metric spaces, in this paper we will discuss as particularly important examples the Laplacian operators that these metrics induce on a collection of data points sampled from a distribution over a smooth and compact manifold $\M$. Indeed, the metric $\ell_{p}$ can be used to define graph kernel functions which then lead to embeddings of $\X$ using, for example, the low-frequency eigenvectors of the graph Laplacian. These eigenvectors contain valuable information that can be used in machine learning tasks such as trend filtering, clustering, or dimensionality reduction. The difficulty in analyzing the induced Laplacians and their eigenvectors relies, particularly, on the fact that Fermat distances over data clouds are themselves random (in contrast to the more standard Euclidean metric). This difficulty has impeded full statistical and analytical understanding of the methods that utilize Fermat-based Laplacians.

Discrete-to-continuum convergence of Laplacian spectra on a manifold $\M$ is a well-developed area \citep{belkin2007convergence, burago2015graph, Trillos2019_Error}, at least for random geometric graphs built with the Euclidean distance. The main problem is, given a finite sample $\X\subset\mathcal{M}$, understanding how the sample-based graph Laplacian induced from $\X$ converges as $|\X|\rightarrow\infty$ to an operator defined with respect to the intrinsic geometry of $\mathcal{M}$ (for example, the Laplace-Beltrami operator on $\mathcal{M})$.  This allows to characterize with high probability the behavior of the spectrum of the graph Laplacian, thereby ensuring its good performance in downstream tasks such as clustering. Existing results typically focus on the cases in which the graph Laplacian is constructed using Euclidean distances or the intrinsic geodesic on $\mathcal{M}$.  In the case of Fermat distances, there is a natural continuum analogue of $\ell_{p}$ on $\mathcal{M}$, which we shall denote $\L_{p}$.  The metric $\L_{p}$ is in fact a geodesic distance function on $\M$ with respect to a certain density-dependent Riemannian metric parameterized by $p$.  Unlike in the case of Euclidean distances, the underlying manifold geodesic $\L_{p}$ must itself be estimated from samples using $\ell_{p}$, requiring new tools of analysis.

The computational requirements for computing Fermat distances has also been studied in the literature. Given $n$ data points, Fermat distances can with high probability be computed in a $k$-nearest-neighbors (kNN) graph \citep{little2022balancing, chu2020exact,Groisman2022nonhomogeneous} as long as $k \sim \log(n)$. This implies all pairwise distances can be computed with the Floyd-Warshall algorithm with complexity $O(n^{2}\log(n))$. Furthermore, the Fermat kNN's of all points can be computed with complexity $O((k^{2}+CD)n\log(n))$, where $C$ is a constant that depends exponentially on the intrinsic dimension of the data \citep{mckenzie2019power}.  This allows Fermat distance nearest neighbors (and consequently, graph Laplacians based on Fermat distance nearest neighbors) to be calculated relatively efficiently with quasi-linear complexity. Another quasi-linear approach is to compute Fermat distances on a set of landmarks and then extend quantities of interest using Nystrom-based methods \citep{williams2001using,ghojogh2020multidimensional,platt2005fastmap,yu2012isomap, civril2006ssde,8509134}.

\nc

\subsection{Summary of Contributions}

This paper makes several mathematical, statistical, and algorithmic contributions. 

First, we establish precise local convergence rates of discrete Fermat distances to continuum Fermat distances defined with respect to a non-uniform density $\rho$ on $\M$.  These results are of independent interest for their connections to percolation theory and to a variety of applications in which machine learning tasks rely on the availability of a metric structure over a data point cloud.

Second, we develop spectral convergence results for graph Laplacians constructed with $\ell_{p}$ to operators on $\M$ with geodesic distance given by $\mathcal{L}_{p}$.  These results leverage our metric convergence results and also new geometric results pertaining to the properties of $\M$ when endowed with $\mathcal{L}_{p}$.  Importantly, our results quantify the impact of the underlying geometry of $\M$, and we calculate explicit constants whenever possible. The large sample spectral analysis of Fermat-based graph Laplacians is thus an important application of our metric approximation results.

\begin{figure}[t]
	\centering
\begin{subfigure}[t]{\textwidth}
		\centering
		\includegraphics[width=\textwidth]{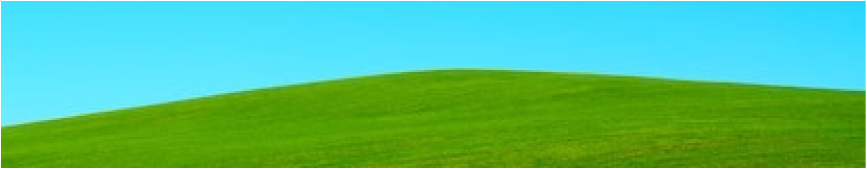}
            \caption{Original Image}
            \label{fig:blue_sky_original_image}
            \vspace{.2cm}
\end{subfigure}
\begin{subfigure}[t]{\textwidth}
		\centering
		\includegraphics[width=\textwidth]{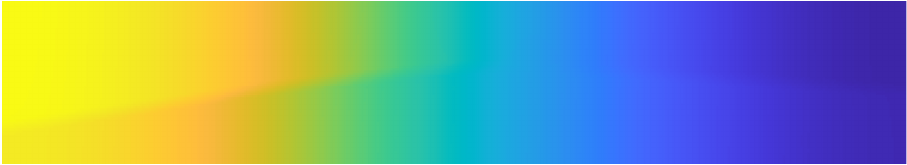}
            \caption{Second eigenvector of Euclidean random walk Laplacian}
            \label{fig:blue_sky_ED_v2}
            \vspace{.2cm}
\end{subfigure}
\begin{subfigure}[t]{\textwidth}
		\centering
		\includegraphics[width=\textwidth]{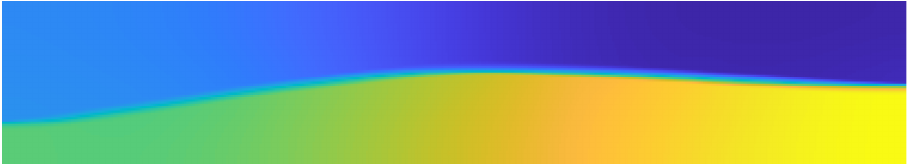}
            \caption{Second eigenvector of Fermat random walk Laplacian ($p=2.5$)}
            \label{fig:blue_sky_FD_v2}
\end{subfigure}
	\caption{\label{fig:landscape}A real image of a landscape that is very elongated; patch-based features are computed which combine color and spatial information.  The second eigenvector of the Euclidean random walk Laplacian is shown in (b) and the second eigenvector for the Fermat Laplacian with $p=2.5$ is shown in (c).  Despite the elongation, the Fermat Laplacian correctly segments background from foreground in a way the Euclidean Laplacian does not.}
\end{figure}

Third, we suggest new spectral clustering algorithms using $\ell_{p}$.  These algorithms enjoy robustness with respect to cluster elongation, a regime where standard spectral clustering fails. We highlight the connection of Fermat Laplacians with a broad family of Laplacian normalizations. Our results provide geometric insight into the choice of normalization parameters and also suggest statistically and computationally efficient methods of practical implementation.  We evaluate our proposed methods on image data, showing the impact of the key parameters in Fermat distance spectral clustering; see Figure \ref{fig:landscape} for an illustration.

\subsubsection{Outline of Paper}

The remainder of the paper is organized as follows. Section \ref{sec:background} provides background on Fermat distances and Laplace operators. Section \ref{sec:setup_mainresults} summarizes our main results. Section \ref{sec:FermatRiemannianMetrics} discusses the Fermat Riemannian metric, including computation of geodesics and discrete to continuum metric approximation results. Section \ref{sec:FermatLaplacians} utilizes the metric approximation to obtain spectral convergence results of Fermat graph Laplacians. Section \ref{sec:NumericsAndNormalizations} discusses some examples and numerical experiments supporting the paper. Section \ref{sec:conclusion} concludes the article.  The Appendix contains proofs of technical results and additional experiments.

\section{Background}
\label{sec:background}

Throughout the paper, we will use the notation in Table \ref{tab:notation}.

\begin{table}[ht!]
\vspace{0.1in}
\begin{center}
\begin{small}
\begin{tabular}{c|c}
\toprule
\textsc{Notation} & \textsc{Definition}  \\

$\| \cdot \|$ & Euclidean 2-norm\\

$| \cdot |$ & norm with respect to the Riemannian metric $g$ \\

$\M$ & data manifold \\

$m$ & intrinsic dimension of data manifold \\

$\X$ & arbitrary point cloud in $\R^D$    \nc\\

$\mathcal{R}$ & reach of $\M$\\

$R$ & injectivity radius of $\M$\\

$K$ & bounds on sectional curvature of $\M$\\

$\omega_m$ & volume of the $m$-dimensional unit ball \\

$\nu$ & probability measure supported on $\M$\\

$\rho$ & density function associated to data measure $\nu$ \\

$\X_n$ & set of $n$ i.i.d. samples from distribution $\nu$ \nc    \\

$H_{n \rho}$ & Poisson point process on $\M$ with intensity $n \rho$ \nc  \\

$N_n$ & Poisson random variable with mean $n$ \\

$p$ & density weight parameter \\

$ \ell_{p}$ & discrete Fermat distance with parameter $p$\\

$\L_{p}$ & continuum Fermat distance with parameter $p$\\

$g$ & arbitrary Riemannian metric\\

$\bar{g}$ & Euclidean metric\\

 $g_p$ & Fermat Riemannian metric \\
 
 $d(x,y)$ & geodesic distance on $(\M, \bar{g})$ \\

$\alpha = 2(p-1)/m$  & reweighting constant \\

$\mu$ & percolation time constant, depending on $p,d$ \\

$a$ & diffusion maps parameter \\

$\eta$ & kernel function, generally taken as $\eta=\mathbb{1}_{[0,1]}$\\

\midrule

\bottomrule
\end{tabular}
\end{small}
\end{center}
\vskip -0.1in
\caption{\label{tab:notation} Notation used throughout the paper.} 
\end{table}

\subsection{Fermat Distances}
\label{sec:bakground_FermatDis}

Discrete Fermat distances can be understood as classical shortest paths on graphs but with edge lengths penalized according to a parameter $p\in [1,\infty)$.

\begin{defn}\label{defn:PWSPD}For $p \in [1,\infty)$, $x,y\in\mathbb{R}^{D}$, and some finite set $\X\subset\mathbb{R}^{D}$, the \emph{(discrete) $p$-weighted Fermat distance} from $x$ to $y$ is:\begin{align}
\label{eqn:DPWSPD}
\ell_p(x,y,\X) =\min_{\pi=\{x_{i}\}_{i=1}^\Length}\left(\sum_{i=1}^{\Length-1} \|x_{i}-x_{i+1}\|^p\right)^{\frac{1}{p}},
\end{align}
where $\pi$ is a path of points in $\X \cup \{x,y\}$ with $x_{{1}}=x$ and $x_{{\Length}}=y$ and $\|\cdot\|$ is the Euclidean norm.
\end{defn} 

Note when there is no ambiguity about $\X$, we will drop it from the notation and simply write $\ell_p(x,y)$. 
The case $p\in(0,1)$ was studied in \cite{alamgir2012shortest} and shown to have counter-intuitive properties; we do not consider this parameter regime. While $\ell_p(x,y)$ depends on the point cloud $\X$, a population formulation is possible, as follows.

\begin{defn}\label{defn:CFD}Let $(\M,g)$ be a compact, $m$-dimensional Riemannian manifold and $\rho:\M \rightarrow \R_{>0}$ a continuous density function on $\M$. 
For $p \in [1,\infty)$ and $x,y\in\M$, the \emph{(continuum) $p$-weighted Fermat distance} from $x$ to $y$ is: \begin{equation}\label{eqn:CFD}\L_p(x,y) = \left(\inf_{\gamma}\int_0^1 \frac{1}{\rho(\gamma(t))^{\frac{p-1}{m}}} \sqrt{g\left(\gamma'(t), \gamma'(t)\right)}dt\right)^{\frac{1}{p}},\end{equation}where $\gamma:[0,1]\rightarrow\M$ is a piecewise $\mathcal{C}^{1}$ path with $\gamma(0)=x, \gamma(1)=y$. 
\end{defn}

Note when $\M$ is embedded in $\R^D$ and $g$ is the Euclidean metric, one simply has $\sqrt{g\left(\gamma'(t), \gamma'(t)\right)} = \| \gamma'(t)\|$. Suppose further that $\rho \in C^{\infty}(\M)$. Then associated to $\L_p^p$ is a Riemannian metric, $g_p$, defined as:
\begin{equation}
    g_{p,x}(X,Y) := \rho^{-\alpha}g_{x}(X,Y),
    \label{eqn:FermatMetric}
\end{equation}
where $\alpha := 2(p-1)/m$ \citep{Hwang2016_Shortest}. Indeed, one easily observes 
\begin{equation*}
    \L_p^{p}(x,y) = \inf_{\gamma} \int_{0}^1\sqrt{g_p(\gamma^\prime(t), \gamma^\prime(t))}\ dt,
\end{equation*}
{\em i.e.}, the distance function associated to $g_p$ is $\L_p^p$. This interpretation of $\L_{p}^{p}$ as a Riemannian metric on $\M$ leads us  to consider $\ell_{p}^{p}$ rather than $\ell_{p}$ in the discrete setting.

Suppose $\X$ is sampled from a distribution $\nu$ with density $\rho$ with respect to the volume form of $\M$. A natural question is to characterize the convergence of $\ell_{p}^{p}$ to $\L_{p}^{p}$ as $|\X|$ tends to infinity. Two concrete probabilistic models for $\X$ that are popular in the literature are:
\begin{itemize}
\item The \emph{i.i.d. setting}, where we consider $\X= \X_n:=\{ x_1, \dots, x_n \}$ to be a collection of i.i.d. draws from $\nu$. 
\item The \emph{Poisson point process (PPP) setting}, where we consider $\X= H_{n\rho}$ to be the realization of a Poisson point process on $\M$ with intensity $n \rho$.
\end{itemize}
In either case, one is naturally interested in the behavior of $\ell_p^p$ as $n$ tends to infinity.  In this paper we obtain convergence results for both PPP models as well as for the i.i.d. setting, the latter being the model that is most often assumed in statistical learning theory.

\begin{rems}
The Poisson point process $H_{n\rho}$ and the i.i.d. model $\X_n$ are closely connected to each other. Indeed, from an infinite sequence $x_1, x_2, \dots$ of i.i.d. samples from $\nu$ and from a Poisson random variable $N_n$ with mean $n$, independent from the $x_i$, one can generate $H_{n \rho}$ by setting $ H_{n \rho}= \{ x_1, \dots, x_{N_n} \}$. In particular, conditioned on $N_n$, $H_{n\rho}$ consists of $N_n$ i.i.d. points drawn from $\nu$.
\label{rem:PPP_to_IID}
\end{rems}

\nc

Informally, one of our main results (Theorem \ref{thm:metric_approx}) states that, with high probability, the discrete Fermat distance \eqref{eqn:DPWSPD} approximates locally the geodesic distance \eqref{eqn:CFD}.  We first derive a metric approximation result for PPPs on manifolds using results from percolation theory.  We then extend the result to the i.i.d. case through a method typically known as \emph{de-Poissonization}.  We then explore some of the geometric implications of this convergence, including quantitative rates for the spectral convergence of normalized graph Laplacian operators induced by discrete Fermat distances toward weighted Laplace-Beltrami operators relative to the family of Riemannian metrics $g_{p}$, a result of interest for its direct application to manifold learning. From the structure of the resulting continuum Laplacian operators, we extract theoretical insights on the role that data density plays in clustering and how, by picking $p$ appropriately, we can accentuate or suppress the effect of density on the resulting data partitioning. 

This paper is not the first to address the discrete-to-continuum convergence of Fermat distances.  Precise statement of such results requires an appropriate normalization, to account for the fact that $\ell_{p}\rightarrow 0^{+}$ as $n\rightarrow\infty$.  It can be shown \citep{howard2001geodesics} that the correct scaling constant for $\ell_{p}$ is $n^{(p-1)/pm}$, so that the relevant convergence question is that of $\tilde{\ell}_{p}^{p}:=n^{(p-1)/m}\ell_{p}^{p}$ to $\L_{p}^{p}$.

\cite{Hwang2016_Shortest} establish that for $\M$ compact and $\rho$ continuous and bounded away from 0, there exists a constant $\mu>0$, depending only on $p$ and $m$, such that for all $b,\epsilon>0$, there exists $\theta_{0}>0$ such that for $n$ large enough,
\[\Prob \left(\sup_{\mathcal{L}_{p}^{p}(x,y)\ge b}\left|\frac{\tilde{\ell}_{p}^{p}(x,y)}{\L_{p}^{p}(x,y)}-\mu\right|>\epsilon\right)\le\exp(-\theta_{0}n^{1/(m+2p)}).\]

The uniform lower bound $b>0$ was removed by \cite{fernandez2023intrinsic}, 
significantly improving the convergence results.  However, this type of convergence in probability is inadequate for our purposes as we need precise characterization of $\epsilon, \theta_{0}$, at least locally.  

\cite{little2022balancing} provide more precise convergence results, albeit in a limited setting.  They show that if $\rho$ is uniform and $\M$ is convex, compact, and has $m$-dimensional unit volume, then for $n$ sufficiently large and for all points $x,y$ sufficiently far from $\partial \M$, $\displaystyle\Ex\left[(\tilde{\ell}_{p}(x,y)-\mu\L_{p}(x,y))^{2}\right]\lesssim n^{-\frac{1}{m}}\log^{2}(n).$  While this result is precise, it applies only to uniform densities, and is thus inadequate for developing spectral convergence results for a large class of densities $\rho$.


\subsection{Discrete Laplace Operators}
\label{subsec:Laplace_Operators}

Let $\X=\{ x_1, \dots, x_n \}$ be a finite set of points on $\M\subset\mathbb{R}^D$. Let $\eta:[0,\infty) \to [0,\infty)$ be any non-increasing function with support contained in $[0,1]$ and $\eta(1/2)>0$.  For ease of exposition in all that follows, we take $\eta = \frac{1}{\omega_m}\mathbf{1}_{[0,1]}$ where the normalizing factor $\omega_m$ denotes the volume of the unit ball in $\mathbb{R}^{m}$ and is chosen so that $\int_{\mathbb{R}^m}\eta(\|x\|)dx = 1$. Choosing a distance function $d_{0}(\cdot,\cdot)$ on $\X$ and a {\em bandwidth parameter} $h$, we define a weighted graph $\Gamma^{d_{0},h} = (\X,W^{d_{0},h})$ by setting
\begin{equation}
    w^{d_{0},h}_{ij} =  \frac{1}{nh^m} \eta\left(\frac{d_{0}(x_i,x_j)}{h}\right).
    \label{eq:WeightsRGG}
\end{equation}
There are several popular definitions of Laplacian operators on $\Gamma^{d_{0},h}$; in this paper we focus on the {\em random-walk Laplacian}, defined for any $u: \X\to\mathbb{R}$ as 
\begin{equation*}
	\left(\Delta_{\Gamma^{d_{0},h}}u\right)(x_i) = \frac{2(m+2)}{h^2} \sum_{j} \frac{w^{d_{0},h}_{ij}}{m_{d_{0},h,i}}\left(u(x_i) - u(x_j)\right).
\end{equation*}
where $m_{d_{0},h,i} := \sum_jw^{d_{0},h}_{ij}$ is the {\em degree} of $x_i$; when the choices of metric $d_{0}$ and scaling parameter $h$ are clear, we will simply write $m_{i}$.  Up to a normalization constant, the random walk Laplacian can be expressed in matrix form as $\Delta_{\Gamma^{d_{0},h}}=I-D^{-1}W$, where $D$ is a diagonal matrix containing the degrees and $W$ is the matrix of weights. As is well-known, (e.g., \citep{Trillos2019_Error}), the eigenvalues of $\Delta_{\Gamma^{d_{0},h}}$ may equivalently be computed using the corresponding \emph{Dirichlet form},
\begin{equation}
	b^{d_{0},h}(u,v) := \frac{m+2}{nh^2}\sum_{i,j} w^{d_{0},h}_{ij}\left(u(x_i) - u(x_j)\right)\left(v(x_i) - v(x_j)\right).
\label{eq:GraphDirichlet}
\end{equation}
Specifically,
\begin{equation*}
	\lambda_k(\Delta_{\Gamma^{d_{0},h}}) = \min_{L_k \, : \, \dim(L_k) = k} \,\max_{u \in L_k\setminus\{0\}} \frac{b^{d_{0},h}_{\Gamma}(u,u)}{\|u\|_{\mathbf{m}}^2},    
\end{equation*}
where the minimization is over $k$-dimensional subspaces $L_k$ and $\|u\|_{\mathbf{m}}^2 := \frac{1}{n}\sum_i m_i u_i^2$. 
When $d_0(x,y)$ is the Euclidean distance $\|x-y\|_2$, the eigenvectors of $\Delta_{\Gamma^{d_0,h}}$ with smallest eigenvalues (which we will hereafter refer to as ``low frequency" in analogy with Fourier analysis) have important applications in unsupervised machine learning.  Indeed, the second lowest frequency eigenvector (and first with eigenvalue greater than 0) is closely connected to normalized graph cuts \citep{Shi2000}.  The problem of finding a partition $(\mathcal{Z}_{*},\mathcal{Z}_{*}^{c})$ of $\X$ which minimizes 
\begin{align}\label{eqn:NCut}\text{Ncut}(\mathcal{Z})=\sum_{i\in\mathcal{Z},j\in\mathcal{Z}^{c}}W_{ij}\bigg/\displaystyle\min\left\{\sum_{i\in\mathcal{Z},j\in\X}W_{ij},\sum_{i\in\mathcal{Z}^{c},j\in\X}W_{ij}\right\}
\end{align}is NP-hard.  One can relax the hard cluster assignments (which correspond to integer constraints in optimization of the Rayleigh quotient) in (\ref{eqn:NCut}), which leads to making partition assignments by thresholding the second lowest frequency eigenvector of $\Delta_{\Gamma}$.  This approach can be extended to more than 2 clusters by running $K$-means or Gaussian mixture modeling on a small number of the lowest frequency eigenvectors of $\Delta_{\Gamma}$.  These procedures---in which low-frequency Laplacian eigenfunctions are used as features in baseline clustering algorithms---are called \emph{spectral clustering} and have been well-studied and extended in recent decades \citep{Ng2002, VonLuxburg2007, Schiebinger2015_Geometry,NGTHoffmannHosseini}.

\subsubsection{Alternative Graph Laplacian Normalizations}
\label{sec:OtherGraphLaplacians}

In the previous sections we have introduced a family of graph Laplacians that are proximity-based relative to a family of data-driven distances. In this section, we discuss another family of graph Laplacians based on density normalizations.  Given $j,q,r \in \R$ and an initial weight matrix $W\in\mathbb{R}^{n\times n}$ (for some $n \in \mathbb{N}$) together with its associated degree matrix $D$, let $W_{q}$ be the weight matrix defined as $(W_{q})_{ij}:=W_{ij}/(D_{i}^{q}D_{j}^{q})$ and $D_{q}$ be its corresponding degree matrix.  We then define the matrix:
\[ L_{j,q,r}:= \begin{cases} D_{q}^{\frac{1-j}{q-1}} (D_{q} -W_{q}) D_{q}^{- \frac{r}{q-1}}, & \text{if } q\not = 1 , \\ D_{q} - W_{q} , & \text{if } q=1. \end{cases} \]
Some popular choices of $(j,r,q)$ are $q=1$, which yields the standard unnormalized graph Laplacian; $(j,q,r)=( 2 ,3 , 1)$, which yields the symmetric normalized Laplacian; and $(j,q,r)=(2,2,0)$, which yields the random walk Laplacian.  We note moreover that the family of operators inducing \emph{diffusion maps} \citep{Coifman2005_Geometric, Coifman2006diffusion} is obtained by taking $(j,q,r)= (2 (1-a), 2(1-a), 0)$ for $a \in [0,1]$. 

In \cite{Hoffmann2019_Spectral}, the base weights $W_{ij}$ over a data set $\X$ sampled from a distribution with density $\rho$ over $\M$ are constructed as in \eqref{eq:WeightsRGG} using the Euclidean distance as $d_0(\cdot, \cdot)$. The resulting graph Laplacians are closely related to the family of differential operators discussed in Remark \ref{rem:ContinuumLimitsHoffmann}. 

\subsection{Continuum Laplace Operators}
\label{sec:Continuum_Laplace_Operators}

There is also a corresponding \emph{continuum} Laplacian that is not based on point clouds, but instead is defined at the population level.  We recall the definition of the $s$-weighted Laplacian from \cite{hein2007graph} (note we have changed the sign in Definition \ref{def:weighted_lap} for consistency with our notation). 


\begin{defn}
	\label{def:weighted_lap}
	Let $(\M,g)$ be a Riemannian manifold with probability measure $\nu$ and associated density $\rho$ defined with respect to $\V$. Let $\Delta$ be the Laplace-Beltrami operator on $(\M,g)$. For $s \in \mathbb{R}$, we define \emph{the $s^{\text{th}}$ weighted Laplacian} $\Delta_s$ as
	\begin{align*}
		\Delta_s &:= -\Delta - \frac{s}{\rho}g^{ij}(\nabla_i\rho)\nabla_j = -\frac{1}{\rho^s}g^{ij}\nabla_i(\rho^s\nabla_j) = -\frac{1}{\rho^s}\divv(\rho^s\grad) \, .
	\end{align*}
\end{defn}
Note that $s=0$ gives the negative Laplace-Beltrami operator under the geometry determined by $g$; $s=2$ is the continuous version of the normalized random walk Laplacian. The diffusion maps framework \citep{Coifman2006diffusion} considers the family of operators $s=2(1-a)$ for $a \in [0,1]$. If, in the construction described in Definition~\ref{def:weighted_lap}, one uses the Fermat metric $g_p$ as defined in \eqref{eqn:FermatMetric} instead of $g$ we arrive at a new operator defined on $\M$, which we denote as $\Delta_{s,p}$.  Explicitly,

\begin{equation}
\Delta_{s,p} := -\frac{1}{\rho_p^s} \divv_p( \rho_p^s \nabla_p )\, ,
\label{eqn:Deltasp}
\end{equation}
where $ \divv_p, \nabla_p$ are the divergence and gradient in the geometry induced by $g_p$, and $\rho_p$ is the density of the measure $\nu$ with respect to the volume form $\V_p$. For the latter, recall that any Riemannian metric $g$ has an associated volume form, $\V$, defined with respect to any local coordinates $x^1,\ldots, x^m$ as $\V = \sqrt{\det(g)}dx^1\cdots dx^m$, or simply $\V = \sqrt{\det(g)}dx$. From this, one sees that
\begin{equation*}
\V_p := \sqrt{\det(g_p)}dx = \sqrt{\det(\rho^{-\alpha}g)}dx = \rho^{-m\alpha/2}\V = \rho^{1-p}\V,
\end{equation*}
where the final equality comes from the definition of $\alpha$. Recall that $\rho$ is the density of the measure $\nu$ with respect to $\V$. That is, for all $U \subset \M$,
\[\nu(U) = \int_{U}\rho\V = \int_{U}\rho\cdot \rho^{p-1}\left(\rho^{1-p}\V\right) = \int_{U} \rho^{p}\V_p  .\]
Thus $\rho_p$, the density of $\nu$ with respect to $\V_p$, is $\rho_p = \rho^{p}$. 

To get a better intuition on how the parameters $p,s$ affect the qualitative properties of the operators $\Delta_{s,p}$, specifically, properties of their spectra and implications for data clustering, it will be convenient to relate geometric quantities in $(\M, g_p)$ with those in $(\M, g)$; we will annotate the objects pertaining to the $(\M,g_p)$-geometry with the subscript $p$.  First, gradients of smooth functions $f : \M \rightarrow \R$ under the different geometries are related according to
$\nabla_p f (x) = \rho^\alpha(x) \nabla f(x), \quad x\in \M.$  This standard identity is deduced by writing the first variation of $f$ in a given direction as an inner product between the gradient of $f$ (in each geometry) and the direction of variation. From the relations between gradients and volume forms for the two geometries $g$ and $g_p$ one can also deduce a relation between the divergences of a smooth vector field under $g$ and $g_p$. For this, we recall the integration by parts formulae:
\begin{align}
\begin{split}
\int_{\M} g(\nabla f, V) \V &= - \int_\M  \divv(V) f  \V,\\
\int_{\M} g_p(\nabla_p f, V) \V_p &= - \int_\M  \divv_p(V) f  \V_p , \quad 
\label{eqn:IntPartsg_p}
\end{split}
\end{align}
which hold for all smooth scalar functions $f$ and all smooth vector fields $V$. From the above, one can readily obtain the following relation: $\divv_p(V) =  \rho^{p-1} \divv(\rho^{1-p} V).$  We can also relate the second order geometries of $(\M, g)$ and $(\M, g_p)$. In particular, \nc the sectional curvatures of $(\M,g_{p})$ can be controlled in terms of those of $(\M,g)$ as stated precisely in Theorem \ref{thm:New_Sec_Curv} in Appendix \ref{App:Sec_Curv}.

With the above identities we can now rewrite $\Delta_{s,p}$ in terms of differential operators in the geometry of $(\M,g)$.

\nc


\begin{prop}[Fermat $s$-Laplacian in Euclidean Coordinates]\label{thm:PM_lap}  When $\M \subseteq \R^D$ is an embedded manifold and $g = \overline{g} = \langle \cdot, \cdot\rangle$ is the Euclidean metric, we have: 
	
	\begin{enumerate}[(a)]
	
	\item $\displaystyle\Delta_{s,p} =-\rho^{\frac{2(p-1)}{m}} \left[\Delta + \left(p(s-1) + 1+\frac{2(p-1)}{m}\right)\frac{\nabla\rho}{\rho}\cdot\nabla \right]\,,$ where $\Delta$ is the Laplace-Beltrami operator on $(\M,\overline{g})$.

 	\item When $s=2$, $\Delta_{2,p} =-\frac{1}{\rho^{2p}}\divv_p(\rho^{2p}\nabla_{p})=-\rho^{\frac{2(p-1)}{m}} \left[\Delta + \left(p+1+\frac{2(p-1)}{m}\right)\frac{\nabla\rho}{\rho}\cdot\nabla \right]$ is a random walk Laplacian.

\end{enumerate}

\end{prop}

\begin{proof}
To see (a), recall that the $s$-weighted Laplacian is $\displaystyle\Delta_{s,p} =-\frac{1}{\rho^{ps}}\divv_p(\rho^{ps}\nabla_{p}).$  The desired formula for $\Delta_{s,p}$ follows by using the fact that $\nabla_p = \rho^\alpha \nabla$, $\divv_{p}(\cdot) = \rho^{p-1}\divv(\rho^{1-p} \,\, \cdot)$ (discussed in Section \ref{sec:FermatRiemannianMetrics}) and the product rule.  Plugging in $s=2$ yields (b).

\end{proof}


	
	
	


The operator $\Delta_{s,p}$ is self-adjoint in an appropriately defined inner product space. For any $s \geq 0$ consider the function space:

\[L^2(\M, \rho^{ps}\V_p):=\left\{f: \M\rightarrow \R \ \bigg| \ \int_{\M}f^{2}\rho^{ps}\V_p <\infty\right\}.\] Define an inner product on this space by
\begin{align*}
\langle f, h\rangle_{p,s} &:= \int_\M f h \, \rho^{ps}\,  \V_p =  \int_\M f h \, \rho^{ps}\,  \rho^{1-p}\V =  \int_\M f h \,  \rho^{p(s-1)+1}\V.
\end{align*}
Also, consider the Dirichlet form $D_{s,p}$ defined as
\begin{align}
\begin{split}
D_{s,p}(f,h) & := \int_{\M}g_{p}\left(\nabla_{ p} f, \nabla_{ p } h\right)\rho^{ps}\V_p \\
& = \int_{\M}\rho^{ \alpha  }g\left(\nabla_{\red } f, \nabla_{ } h\right)\rho^{ps}\left[\rho^{1-p}\V\right] \\
& = \int_{\M}g\left(\nabla f, \nabla h\right)\rho^{1+p(s-1) +  \nc \nc \alpha}\V.
\end{split}
\label{eqn:DirichletEenergiesContinuum}
\end{align}

\noindent Using \eqref{eqn:IntPartsg_p} and the first line in \eqref{eqn:DirichletEenergiesContinuum} we can immediately deduce the following.

\begin{prop}
In its domain of definition, $\Delta_{s,p}$ is self-adjoint with respect to the inner product $\langle\cdot, \cdot \rangle_{p,s}$. Moreover, $D_{s,p}(f,h) = \langle \Delta_{s,p}f, h\rangle_{p,s} = \langle f, \Delta_{s,p} h\rangle_{p,s},$ for all smooth $f,h$.
\end{prop}

\begin{rems}
\label{rmk:diffusion}
To intuitively interpret the role of $\rho$ on the operator $\Delta_{s,p}$, it is helpful to see $\Delta_{s,p} $, at least when we consider $\M= \Omega$ to be a domain in Euclidean space for simplicity,
as the generator of the following diffusion:
\begin{align*}
	dX_t &:= \rho^{\frac{2(p-1)}{m}}\left(p(s-1)+1+\frac{2(p-1)}{m}\right)\frac{\nabla\rho}{\rho}\ dt + \sqrt{2}\rho^{\frac{(p-1)}{m}} dB_t.
\end{align*}
As $p>1$, the scalar coefficients in front of the drift and diffusion terms in the above SDE are large at points of high density. This means that a particle moving according to this SDE explores connected regions with high density very rapidly while being drifted away from regions where $\rho$ is small. Thus, low density barriers will be particularly difficult to cross, effectively inducing a separation of regions (clusters) of high density.  
\end{rems}

The problem of understanding for what choice of parameter scaling $h$ does a discrete operator $\Delta_{\Gamma}$ built over samples from $\nu$ converge to $\Delta_{s}$ as $n\rightarrow\infty$ is by now a classical problem \citep{Belkin2003laplacian, Trillos2019_Error,WormelReich,calder2022improved} when  $g$ is the metric induced by a Euclidean embedding of $\M$. One of our contributions in this paper is to analyze the convergence of the discrete graph Laplacian built from Fermat distances to its appropriate continuum analogue. 

\begin{rems}
\label{rem:ContinuumLimitsHoffmann}
It was informally argued in \cite{Hoffmann2019_Spectral} that the graph Laplacian $L_{j,q,r}$ from Section \ref{sec:OtherGraphLaplacians}, built with base weights as in \eqref{eq:WeightsRGG} for i.i.d. points $x_1, \dots, x_n$ sampled from $ \rho\, \V$ and $d_0$ the Euclidean distance, approximates spectrally the family of differential operators 
\[ \mathcal{L}_{j,q,r}f= -\frac{1}{\rho^j} \divv\left( \rho^q \nabla \left(\frac{f}{\rho^r} \right) \right).\]
On the other hand, based on the relations between different geometric quantities under the geometries $g$ and $g_p$ discussed above, the operator $\Delta_{s,p}$ from \eqref{eqn:Deltasp} can be written as
\[ \Delta_{s,p}f = -\frac{1}{\rho^{(s-1)p + 1}} \divv( \rho^{(s-1) p +1 + \alpha} \nabla f     ), \]
where $\alpha = 2(p-1)/m$. This says that, in principle, the graph Laplacians discussed in \cite{Hoffmann2019_Spectral} can be used to recover the spectrum of $\Delta_{s,p}$, provided one chooses $j= (s-1) p +1$, $q=(s-1)p + 1 + \alpha$, and $r=0$, in which case $\mathcal{L}_{j,q,0}=\Delta_{s,p}$. We will use this observation in our numerics Section \ref{subsec:NormalizationsVsFermat} when comparing Fermat-based graph Laplacians with their degree-reweighted counterparts. 

    
\end{rems}

\section{Set-up and Main Results}
\label{sec:setup_mainresults}
In this section, we introduce some more technical background and state our main results.  Background on Riemannian geometry is given in Appendix \ref{app:prelim}. We assume that $\mathcal{M}$ is normalized so that it has volume $1$ with respect to its canonical measure dVol ({\em i.e.,} the measure given by the volume form associated to $g$).

\subsection{Notation and Assumptions}
\label{subsec:assump}

We shall make the following standing assumptions on the manifold $\M$ and density $\rho$.

\begin{assump}
\label{assump:density}
We assume $\M\subset\mathbb{R}^{D}$ is an embedded, smooth, compact manifold of dimension $m$ and that $\rho$ satisfies:
\begin{enumerate}[(i)]
    \item there exists $\beta>0$ such that $\frac{1}{\beta} \leq \rho(x) \leq \beta$ for all $x\in\mathcal{M}$.
    \item $\rho\in\mathcal{C}^{\infty}(\mathcal{M})$.
    \item $\rho$ is Lipschitz continuous: $\left|\rho(x) - \rho(y)\right| \leq L_1d(x,y) \, \text{ for all } x,y\in\M.$
    \item $\grad\rho$ is Lipschitz continuous: $\|\nabla^2\rho(x)\| \leq L_2$ for all $x \in \M$. 
\end{enumerate}
\end{assump}

Fix $x\in \M$ and let $\T_x = \exp_x^{-1}$ be the inverse of the exponential map at $x$. Let $R=\text{inj}(\M)$ be the largest radius such that $\T_x$ is a diffeomorphism on $\mathcal{B}_x(R)$ for all $x$, where $\mathcal{B}_x(R)$ is a geodesic ball (in terms of geodesic distance $d$) of radius $R$ centered at $x$. Let $K$ be the maximal sectional curvature of $\M$, and let $\reach = \text{reach}(\M)$. For $v, w \in T_x\M$, let $J_x(v)=\sqrt{\det(g)}$ be the Jacobian in the normal coordinates induced by the exponential map.  

For all $v,w\in B_0(R)$, we have:
\begin{align}
	\label{equ:Jacobian_bound}
	1 - CmK\|v\|^2 &\leq J_x(v) \leq 1 + CmK\|v\|^2,
\end{align}
\begin{align}
	\label{equ:tangent_plane_dis}
	\| v - w\| - CK\| v - w\|^3 &\leq d(\exp_x(v),\exp_x(w)) \leq \| v - w\| + CK\| v - w\|^3.
\end{align}
For \eqref{equ:Jacobian_bound}, see (1.34) in \cite{Trillos2019_Error}, and for \eqref{equ:tangent_plane_dis}, see Proposition E.1 in \cite{little2022balancing}.
Furthermore (see Proposition 2 in \cite{Trillos2019_Error}), for all $x,y\in \M$ such that $\|x-y\| \leq \frac{\reach}{2}$, we have
\begin{align}
	\label{equ:Euc_to_geo_dis}
	\|x-y\|  &\leq d(x,y) \leq \|x-y\| + \frac{8}{\reach^2}	\|x-y\|^3.
\end{align}

\subsection{Statements of Main Results}
\label{sec:mainresults}

In this section we state our main results, the first of which quantifies the discrepancy between Fermat distances at the continuum level and Fermat distances built from i.i.d. point clouds.


\begin{thm}[Local Metric Approximation for i.i.d. Point Clouds]
\label{thm:metric_approx_iid}
	Let $x,y\in\M$.  Fix $\epsilon\in(0,1/(8p+6))$ and let $\kappa = \frac{2\epsilon}{3}\min\{\frac{1}{m},\frac{1}{p}\}$. Let Assumption \ref{assump:density} hold and suppose moreover that 
    \begin{equation*}
   2(n\beta/2)^{-\frac{1}{m}\left(\frac{1}{3} - \epsilon\right)}\leq d(x,y) \leq C_{\M,\rho,p} \quad,\quad m\geq 2\, .
\end{equation*}
	Let $\X_n$ consist of $n$ i.i.d. samples from $\rho$. Then for $n$ large enough: 
	\begin{align*}
		|\tilde{\ell}_p^p(x,y,\X_n) - \mu \L_p^p(x,y)| &\leq C_1\L_p^p(x,y)^2 + C_2\L_p^p(x,y)^{3}
	\end{align*}
	with probability at least $1-C_{\epsilon}n\exp\left(-c_{\epsilon}(\frac{n}{4\beta})^{\kappa}\right)$, where
	\begin{align*}
        C_1 &= \beta^{\frac{p-1}{m}}\left(\frac{5\mu}{2}\left(\frac{p-1}{m}\right)L_1\beta + 1\right)\ ,\  C_2 =  C_{p,m,\beta}\left(\reach^{-2} + K(1+L_1) +L_1^2 + L_2 \right) \, .
	\end{align*}
 Here $C_{\epsilon},c_{\epsilon}$ are constants depending on $\epsilon$, $C_{p,m,\beta}$ is a constant depending on $p,m,\beta$, and $C_{\M,\rho,p}$ is a constant depending on $\rho$, $p$, and the geometry of $\M$.
\end{thm}
\begin{proof}
    See Appendix \ref{app:DePPP}. 
\end{proof}

Theorem \ref{thm:metric_approx_iid} follows from Theorem \ref{thm:metric_approx} (metric approximation for PPPs) and a de-Poissonization argument.  As an application of this local distance approximation result we obtain rates for the spectral convergence of Fermat-based graph Laplacians toward their continuum analogues. The graph Laplacians are built over realizations of an i.i.d. point cloud, as we state precisely below. For legibility, we write $\Gamma^{\tilde{\ell},h}$ instead of $\Gamma^{\tilde{\ell}_{p}^p,h}$ and $\Gamma^{\L,h}$ instead of $\Gamma^{\mu\L_p^p,h}$ when $p$ is clear from context. We also denote constants by $C$ or $c$, with subscripts indicating the primitive quantities ($p,m,L_1$\ldots) upon which they depend.

\begin{thm}
\label{thm:Eigenvalue_convergence}
    Let $\Gamma^{\tilde{\ell},h}$ denote the graph constructed using the discrete Fermat distance $\tilde{\ell}^p_p$ and bandwidth $h = O(n^{-\frac{1}{3m} + \frac{\epsilon}{m}})$ for some $\epsilon \in (0, \frac{1}{8p+6})$ over a point cloud of i.i.d. points $\X = \{x_1, \dots, x_n \}$ sampled from $\rho$. Let $\Delta_{\Gamma^{\tilde{\ell},h}}$ denote the random walk Laplacian of this graph and $\lambda_k(\Gamma^{\tilde{\ell},h})$ its $k$-th eigenvalue. Finally, let $\Delta_{2,p}$ denote the $2$-weighted Fermat Laplacian:
    \begin{equation*}
        \Delta_{2,p} u =  - \frac{1}{\rho_p^2} \divv_p\left( \rho^2_p\nabla_p u \right)
    \end{equation*}
    and let $\lambda_k(\Delta_{2,p})$ denote its $k$-th eigenvalue. Then for $n$ large enough
    \begin{equation*}
    \frac{\left|\lambda_k(\Delta_{\Gamma^{\tilde{\ell},h}}) - \lambda_k(\Delta_{2,p})\right|}{|\lambda_k(\Delta_{2,p})|} \leq C_{\beta,p,m,L_1} n^{-\frac{1}{3m} + \frac{\epsilon}{m}} + C_{\beta,p,m,L_1,L_2,K,\reach}n^{-\frac{2}{3m} + \frac{2\epsilon}{m}}
    \end{equation*} 
    with probability at least $1-C_\epsilon n^3\exp\left(-c_\epsilon (\frac{n}{4\beta})^{\frac{\epsilon}{2p+1}\min\{\frac{1}{m},\frac{1}{p}\}} \right)$.
\end{thm}

\begin{rems} Note that the leading error term is $O(n^{-\frac{1}{3m}+ \frac{\epsilon}{m}})$. In contrast, the leading error term in \cite{Trillos2019_Error} is $O(n^{-\frac{1}{2m}})$. This is explained by the fact that in Theorem \ref{thm:Eigenvalue_convergence} the dominating source of error comes from the approximation $\tilde{\ell}_p^p(\cdot, \cdot) \approx \mu\L_p^p(\cdot, \cdot)$ which decays more slowly than the $\infty$-optimal transport distance between the sample measure associated to $\X$ and population measure $\nu$---the dominating source of error in \cite{Trillos2019_Error}.
\end{rems}



We use Theorem \ref{thm:Eigenvalue_convergence} and combine it with some auxiliary estimates discussed in Sections \ref{sec: Discrete Dirichlet energies} and \ref{sec:SpectralConvergenceDiscreteToContinuum} to prove the convergence of eigenvectors of $\Delta_{\Gamma^{\tilde{\ell},h}}$ toward eigenfunctions of $\Delta_{2,p}$. While said results can be derived for arbitrary eigenvectors, we will focus on the consistency of the Fiedler eigenvector (i.e., the one with second smallest eigenvalue) for simplicity and refer the reader to works like \cite{burago2015graph,Trillos2019_Error} for extensions; see also Remark \ref{rem:ExtensionsEigenvectors} below. We will further assume that $\Delta_{2,p}$'s Fiedler eigenvalue is simple. In order to compare the discrete and continuum eigenvectors, we use a \textit{discretization map} that associates a function defined over the data set to every function defined over the manifold $\M$; this operator is introduced in \eqref{eq:Def_Discretization}.
We prove that the difference between the discretization of the continuum Fiedler eigenfunction and its projection onto the graph-Laplacian Fiedler eigenspace tends to zero as $n \rightarrow \infty$; we also quantify the rate of convergence. Other notions of consistency for eigenvectors of discrete graph Laplacians toward manifold counterparts are possible and we refer the reader to \cite{Trillos2019_Error} for more details.

\begin{thm}
\label{thm:eigenvector}
Let the assumptions and notation from Theorem \ref{thm:Eigenvalue_convergence} hold, and furthermore assume that the Fiedler eigenvalue $\lambda_2=\lambda_2(\Delta_{2,p})$ of $\Delta_{2,p}$ is simple.  Let $f$ be a normalized eigenfunction with eigenvalue $\lambda_2(\Delta_{2,p})$  (which by assumption is unique up to sign). Then with probability at least $1-C_\epsilon n^3\exp\left(-c_\epsilon (\frac{n}{4\beta})^{\frac{\epsilon}{2p+1}\min\{\frac{1}{m},\frac{1}{p}\}} \right)$, for $n$ large enough, 
\begin{align*}
\lVert Pf- \psi_2^n \rVert^2_{\textbf{m}} & \leq \frac{1}{\lambda_3 - \lambda_2} \left[C_{\beta,p,m,L_1}n^{-\frac{1}{3m} + \frac{\epsilon}{m}} +  C_{\beta,p,m,L_1,L_2,K,\reach}n^{-\frac{2}{3m} + \frac{2\epsilon}{m}} \right]\lambda_2 \\
    & + C_{\beta,p,m,L_1}n^{-\frac{1}{3m} + \frac{\epsilon}{m}} + C_{m,\beta,p,L_1,L_2,K}n^{-\frac{2}{3m} + \frac{2\epsilon}{m}}  , 
\end{align*}
where $\lambda_3= \lambda_3(\Delta_{2,p}) $, $P$ is the discretization map defined in \eqref{eq:Def_Discretization}, and $\psi_2^n$ is one of the normalized (w.r.t. $\lVert \cdot \rVert_{\textbf{m}}$) Fiedler eigenvectors of the graph Laplacian $\Delta_{\Gamma^{\tilde{\ell},h}}$. 
\end{thm}

Although Theorem \ref{thm:Eigenvalue_convergence} guarantees that for $n$ large enough $\lambda_2(\Gamma^{\tilde{\ell},h})$ is also a simple eigenvalue, in the statement of Theorem \ref{thm:eigenvector} we talk about ``one of the Fiedler eigenvectors of the graph Laplacian" since uniqueness only holds up to sign. In Section \ref{sec:EigenvecConvergenceMain} we provide further remarks on the convergence stated in Theorem \ref{thm:eigenvector} and discuss other approaches in the literature to prove consistency of eigenvectors for Laplacian matrices built over data clouds.

\nc

\section{Fermat Riemannian Metrics}
\label{sec:FermatRiemannianMetrics}

\subsection{Fermat Geodesics and Local Euclidean Equivalence}
\label{sec:Computing Fermat Geodesics}
In this section we derive a concrete, local expression for geodesics in the Fermat distance on a flat domain ({\em i.e.,} an $m$-dimensional, open, connected subset of $\mathbb{R}^{m}$), and then establish a higher order local equivalence between $\L_p$ and Euclidean distance on a flat domain. These results will then be applied in the tangent plane of a general manifold to establish the local metric approximation results given in Section \ref{sec:metric_approx}. To make clear that the domain lacks both intrinsic and extrinsic curvature, we denote it by $\Omega$ instead of $\M$.  

\begin{restatable}{thm}{thmFermatGeodesics}
	\label{thm:FermatGeodesics}
	Let $\Omega \subseteq \mathbb{R}^m$ be a $m$-dimensional, open, connected domain and assume $(\Omega,\rho)$ satisfies Assumption \ref{assump:density} Fix $y\in \Omega$ and let $\gamma_{b}(t)$ denote the unit speed geodesic with respect to $g_p$ originating at $y$ in the direction of unit vector $b\in\mathbb{R}^{m}$. Then:
	 \begin{align}
  \label{equ:fermat_geodesic_expansion}
  \begin{split}
		\gamma_{b}(t) =&  \rho(y)^{\frac{\alpha}{2}} b t + \alpha \rho(y)^{\alpha -1} \left(\frac{1}{2}\langle b, \nabla \rho(y)\rangle b - \frac{1}{4} \nabla \rho(y) \right)t^2 \\+& (C_1' b + C_2' H(y)b + C_3' \nabla \rho(y))t^3 + O(t^4) \, ,
  \end{split}
	\end{align} where $H(y)$ denotes the Hessian matrix of $\rho$ evaluated at $y$ and 
\begin{align*}
	C_1' 
	&= \left(\frac{1}{3}\alpha^2 -\frac{1}{6}\alpha\right)\rho(y)^{\frac{3}{2}\alpha-2}\langle b,\nabla \rho(y)\rangle^2\\
	&+ \frac{\alpha}{6} \rho(y)^{\frac{3}{2}\alpha-1}\langle H(y)b, b\rangle - \frac{\alpha^2}{12}\rho(y)^{\frac{3}{2}\alpha-2}\langle \nabla \rho(y),\nabla \rho(y)\rangle, \\
	C_2' &= -\frac{\alpha}{12}\rho(y)^{\frac{3}{2}\alpha-1}, \\
	C_3' &=\left(\frac{\alpha}{12}-\frac{\alpha^2}{6}\right)\rho(y)^{\frac{3}{2}\alpha-2}\langle \nabla \rho(y),b\rangle.
\end{align*}
\end{restatable}

\begin{proof}
	See Appendix \ref{app:FermatGeo}.
\end{proof}	

\begin{figure}[ht!]
    \begin{center}
	\includegraphics[width=7.5cm]{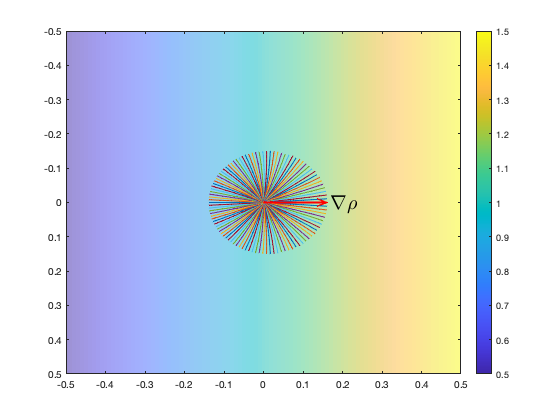}
         \includegraphics[width=7.5cm]{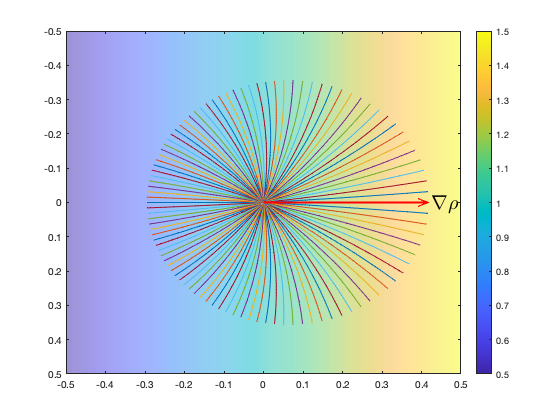}
     \end{center}    
	\caption{\label{fig:FermatBalls} Fermat geodesic balls of radius $T=0.15$ (left) and $T=0.35$ (right). The $\L_{p}^{p}$ ball is wider in directions of high density because geodesics in that direction are less costly.}
\end{figure}

As a numerical illustration of the local Fermat geodesics, consider the density function $\rho(x_1, x_2) = 1+x_1$ for $(x_1, x_2) \in [-\frac{1}{2}, \frac{1}{2}] \times [-\frac{1}{2}, \frac{1}{2}]$, so that $\nabla \rho = (1, 0)$, $\rho(0,0)=1$. We consider $p=3$, and since  $m=2$ we have $\alpha=2$. We solve the geodesic equations appearing in the proof of Theorem \ref{thm:FermatGeodesics} with a numerical solver for a collection of initial directions $b$ and for total time $T$ in order to produce the Fermat geodesic balls appearing in Figure \ref{fig:FermatBalls}; note \eqref{equ:fermat_geodesic_expansion} gives a third order approximation of these curves, valid for small times. 
Intuitively, the $\mathcal{L}_p^p$ ball is elongated in the direction of the gradient because it costs less to travel in that direction.

We can utilize the geodesic computation of Theorem \ref{thm:FermatGeodesics} to establish a higher-order equivalence of Fermat distances with a density-warped Euclidean distance, depending both on $\rho$ and $\nabla \rho$.

\begin{restatable}{thm}{thmHOequiv}
	\label{thm:HO_equiv}
	Let $\Omega \subseteq \mathbb{R}^m$ be an $m$-dimensional, open, connected domain and assume $(\Omega,\rho)$ satisfies Assumption \ref{assump:density}.  Then for $x,y\in\Omega$ with $x\neq y$ and unit vector $u = (y-x)/\|y-x\|$, we can relate Euclidean and Fermat distance as follows:
	\begin{small}
	\begin{align*}
		\|y-x\| &= \rho(x)^{\frac{p-1}{m}}\mathcal{L}_p^p(x,y) + \frac{1}{2}\left(\frac{p-1}{m}\right)\left\langle u, \nabla \rho(x) \right\rangle \rho(x)^{\frac{2(p-1)}{m}-1}\mathcal{L}_p^{2p}(x,y) \\
		&\qquad + C\mathcal{L}_p^{3p}(x,y)+ O(\mathcal{L}_p^{4p}(x,y))
	\end{align*}
\end{small}
for
\begin{small}
\begin{align*}
C &= \rho(x)^{\frac{3}{2}\alpha-2}\left[ \frac{\alpha^2}{96} \langle \nabla\rho(x),\nabla\rho(x)\rangle + \left(\frac{7\alpha^2}{96}-\frac{\alpha}{12}\right)\left\langle u , \nabla \rho(x)\right\rangle^2 + \frac{\alpha}{12}\rho(x)\left\langle H(x) u, u\right\rangle\right].
\end{align*}
\end{small}
Also:
	\begin{small}
	\begin{align*}
		\mathcal{L}_p^p(x,y) &=\frac{1}{\rho(x)^{\frac{p-1}{m}}}\left( \|y-x\| -\frac{1}{2}\left(\frac{p-1}{m}\right)\left\langle u, \frac{\nabla \rho(x)}{\rho(x)} \right\rangle \|y-x\|^2\right) + O\left(\|y-x\|^3\right).
	\end{align*}
\end{small}

\end{restatable}

\begin{proof}
	See Appendix \ref{app:EucEquiv}.
\end{proof}	


\subsection{Discrete-to-Continuum Fermat Distance Approximation}
\label{sec:metric_approx}

We now utilize the results of Section \ref{sec:Computing Fermat Geodesics} to establish that, locally, the discrete, computable metric $\tilde{\ell}_p^p$ is well-approximated by the Fermat geodesic distance $\mathcal{L}_p^p$. Following is the main result of this section, which quantifies the metric approximation for PPPs; we work with PPPs in order to utilize results from percolation theory. 
Recall $H_{n\rho}$ denotes a nonhomogeneous Poisson point process on $\M$ with intensity $n\rho$, and $\tilde{\ell}_p^p(x,y,H_{n\rho})=n^{\frac{p-1}{m}}\ell_p^p(x,y,H_{n\rho})$ denotes the discrete (normalized) Fermat distance computed in $H_{n\rho}\cup \{x,y\}$.  Note although the number of points in $H_{n\rho}$ is random, $\Ex[|H_{n\rho}|]=n$. 

\begin{thm}[Local Metric Approximation for PPP]
	\label{thm:metric_approx}
	Suppose $(\M,\rho)$ satisfies Assumption \ref{assump:density}.  Let $x,y\in\M$, fix $\epsilon\in(0,1/(8p+6))$, and suppose that 
    \begin{equation*}
   2(n\beta)^{-\frac{1}{m}\left(\frac{1}{3} - \epsilon\right)}\leq d(x,y) \leq C_{\M,\rho,p}. 
\end{equation*}
	Then for $n$ large enough: 
	\begin{align*}
		|\tilde{\ell}_p^p(x,y,H_{n\rho}) - \mu \L_p^p(x,y)| &\leq C_1\L_p^p(x,y)^2 + C_2\L_p^p(x,y)^{3} 
	\end{align*}
	with probability at least $1-C_{\epsilon}n\exp\left(-c_\epsilon(\frac{n}{2\beta})^{\frac{2\epsilon}{3}\min\{\frac{1}{m},\frac{1}{p}\}}\right)$, where
	\begin{align*}
        C_1 &= \beta^{\frac{p-1}{m}}\left(\frac{5\mu}{2}\left(\frac{p-1}{m}\right)L_1\beta + 1\right)\ ,\  C_2 =  C_{p,d,\beta}\left(\reach^{-2} + K(1+L_1) +L_1^2 + L_2 \right) \, .
	\end{align*}Here $C_{\epsilon},c_{\epsilon}$ are constants depending on $\epsilon$, $C_{p,d,\beta}$ is a constant depending on $p,d,\beta$, and $C_{\M,\rho,p}$ is a constant depending on $\rho$, $p$, and the geometry of $\M$.
\end{thm}

\subsubsection{Background and Outline for Proof of Theorem \ref{thm:metric_approx}}
We first establish some notation used in the proof of Theorem \ref{thm:metric_approx}. Let $r=d(x,y)$; we assume $r$ is upper bounded by a constant $C_{\M,\rho,p}$ independent of $n$ but depending on $\rho, p$, and the geometry of $\M$, but to simplify already long calculations we do not keep track of it.     
 Let $u=\T_x(y)$ so that $\|u\|=r$.  


Fix any $R >0$ satisfying $4R < \reach$ and define the function $\rho_x(z) = \rho( \exp_x(z))$ for $z\in B_0(R)$, and 0 otherwise. We note that $H_{ng_x}:=\T_x(H_{n\rho} \cap \mathcal{B}_x(R))$ is a PPP on $B_0(R) \subseteq T_x\M$ with intensity $g_x(z) :=  \rho_x(z) J_x(z)$, since for $A\subseteq B_0(R)$, 
\begin{align*}
	\Ex[ H_{ng_x}(A) ] & = \Ex[H_{n \rho}(\exp_x(A))] = n \int_{\exp_x(A)\subseteq\M} \rho(y) dv_y = n \int_{A\subseteq B_0(R)} \rho(\exp_x(z)) J_x(z) dz \, .
\end{align*}
 The function $g_x$ (thought of as the density $\rho$ lifted to the tangent plane) implicitly defines a tangent plane continuum Fermat distance between $x,y$:
\begin{align*}
	\L_p^p(0,u,g_x) = \inf_{\T_x\gamma \subseteq B_0(R)} \int g_x(\T_x\gamma(t))^{(1-p)/m} |(\T_x\gamma)'(t)|\ dt \, 
\end{align*}
where $\T_x\gamma$ is a path in $T_x\M$ connecting 0 and $u$.
Let $\gmin, \gmax$ be the min/max of $g_x(z)$ over $B_0(2r)$. Let $H_{n\gmin}$ be the PPP obtained by replacing $g_x$ with $\gmin$ on $B_0(2r)$ (no change outside of this ball), and similarly for $H_{n\gmax}$. 

Note that we can couple the PPPs so that $H_{n\gmin} \subseteq H_{ng_x} \subseteq H_{n\gmax}.$  Finally, let $\overline{H}_{n\gmin}=H_{n\gmin} \cap B_0(2r)$, $\overline{H}_{n\gmax}=H_{n\gmax} \cap B_0(2r)$.  Note that $\overline{H}_{n\gmin}$, $\overline{H}_{n\gmax}$ are \emph{homogeneous} PPPs on $B_0(2r)$ with intensities $n\gmin$, $n\gmax$, while $H_{ng_x}$ is inhomogeneous and harder to analyze directly.  

Our metric approximation consists of a series of Lemmas:
\begin{enumerate}
	\item \underline{Lemma \ref{lem:local_paths}}:  $\tilde{\ell}_p^p(x,y,H_{n\rho}) = \tilde{\ell}_p^p(x,y,H_{n\rho} \cap \mathcal{B}_x(R))$ holds w.h.p.  Intuitively, this means that optimal discrete Fermat paths between nearby points do not meander too much.  This allows the paths on the manifold to be compared to those on the tangent plane.
	\item \underline{Lemma \ref{lem:curv_pert_dis}}:  $\tilde{\ell}_p^p(x,y,H_{n\rho} \cap \mathcal{B}_x(R)) \approx \tilde{\ell}_p^p(0,u,H_{ng_x})$.  This means that optimal paths on the manifold are close to those on the tangent plane.  The paths on the tangent plane are easier to analyze, because the domain is flat. 
	 \item \underline{Lemma \ref{lem:local_geo_homog_PPP}}: $\ell_p^p(0, u, H_{n\gmin}) = \ell_p^p(0,u,\overline{H}_{n\gmin})$ w.h.p., and same for $H_{n\gmax}$ \\
	 This allows us to ``trap" the Fermat distance for the nonhomogeneous PPP $H_{ng_x}$ as $\ell_p^p(0,u,\overline{H}_{n\gmax} )\leq  \ell_p^p(0,u,H_{ng_x})  \leq \ell_p^p(0,u,\overline{H}_{n\gmin}).$
	  \item \underline{Lemma \ref{lem:metric_approx_tangent_plane}}: $\tilde{\ell}_p^p(0,u,H_{ng_x}) \approx \mu\mathcal{L}_p^p(0,u,g_x)$ w.h.p.  This is a direct discrete-to-continuum metric approximation on the tangent plane.
	  \item \underline{Lemma \ref{lem:curv_pert_cont}}: $\mathcal{L}_p^p(0,u,g_x) \approx \mathcal{L}_p^p(x,y)$. This involves bounding the perturbation due to curvature for $\mathcal{L}_p^p$.
\end{enumerate}

See Appendix \ref{app:MetricApprox} for precise statements and proofs of the above lemmas. An important tool in our analysis is the following result from percolation theory, which is a direct result of Theorem 2.2 in \cite{howard2001geodesics}.

\begin{prop}
	\label{prop:conv_PD_homog_PPP}
	Let $H_{n\lambda}$ be a homogeneous PPP on $\mathbb{R}^m$ with intensity $n\lambda>1$. Fix $q > 1$ and $\epsilon \in (0,1/(8p+6))$. Suppose 
    \begin{equation}
        \|u\| \geq \left(n\lambda\right)^{-\frac{1}{m}\left(\frac{1}{2q-1}-\epsilon\right)}
        \label{eq:u_lower_bound}
    \end{equation}
  Then 
	\begin{equation}
		\label{eq:u_convergence_rate}
		\left| (n\lambda)^{\frac{(p-1)}{m}}\ell_p^p(0,u,H_{n\lambda}) - \mu\|u\| \right| \leq \|u\|^q
	\end{equation}
	with probability at least 
    \begin{equation}
    1-C_\epsilon \exp\left(-c_\epsilon(n\lambda)^{\epsilon\left(\frac{2q-2}{2q-1}\right)\min\{\frac{1}{m},\frac{1}{p}\}}\right).
    \label{eq:u_prob_bound}
    \end{equation}
    for constants $C_\epsilon,c_\epsilon$ depending on $\epsilon$.
\end{prop}

\begin{proof} 
	By Theorem 2.2 in \cite{howard2001geodesics}, for every $\epsilon\in(0,\frac{1}{8p+6})$, $| \ell_p^p(0,z,H_1) - \mu\|z\| | \leq \|z\|^{\frac{1}{2}+\epsilon}$ with probability at least $1-C_\epsilon\exp(-c_\epsilon\|z\|^{\epsilon\min\{1,\frac{m}{p}\}})$.  Via $(n\lambda)^{\frac{1}{m}}u=z$, we obtain
	\begin{align}
		| (n\lambda)^{\frac{p}{m}}\ell_p^p(0,u,H_{n\lambda}) - \mu(n\lambda)^{\frac{1}{m}}\|u\| | &\leq \left((n\lambda)^{\frac{1}{m}}\|u\|\right)^{\frac{1}{2}+\epsilon} \nonumber \\
		\implies 	| (n\lambda)^{\frac{1}{m} +\frac{p-1}{m}}\ell_p^p(0,u,H_{n\lambda}) - \mu(n\lambda)^{\frac{1}{m}}\|u\| | &\leq \left((n\lambda)^{\frac{1}{m}}\|u\|\right)^{\frac{1}{2}+\epsilon} \nonumber \\
		\implies 	| (n\lambda)^{\frac{p-1}{m}}\ell_p^p(0,u,H_{n\lambda}) - \mu\|u\| | &\leq \frac{\|u\|^{\frac{1}{2}+\epsilon}}{(n\lambda)^{\frac{1}{m}(\frac{1}{2}-\epsilon)} } \label{eq:eps_percolation_bound}
	\end{align} 
    with probability
    \begin{equation}
        1-C_\epsilon\exp\left(-c_\epsilon\left((n\lambda)^{\frac{1}{m}}\|u\|\right)^{\epsilon\min\{1,m/p\}}\right).
        \label{eq:pert_prob_bound}
    \end{equation}
    We observe that \eqref{eq:u_lower_bound} implies $\|u\| \geq \left(n\lambda\right)^{\frac{-\left(1 - 2\epsilon\right)}{m\left(2q - 1 - 2\epsilon \right)}}$; thus the right hand side of \eqref{eq:eps_percolation_bound} can be bounded by
    \begin{align*}
        \frac{\|u\|^{\frac{1}{2}+\epsilon}}{(n\lambda)^{\frac{1}{m}(\frac{1}{2}-\epsilon)}} \leq \frac{\|u\|^{\frac{1}{2}+\epsilon}}{\|u\|^{-q+1/2+\epsilon}} = \|u\|^q.
    \end{align*}
    Similarly $\left(n\lambda\right)^{1/m}\|u\| \geq \left(n\lambda\right)^{1/m}\left[\left(n\lambda \right)^{\frac{-(1-2\epsilon)}{m(2q-1-2\epsilon)}}\right] = \left(n\lambda\right)^{\frac{2q-2}{m(2q-1-2\epsilon)}} \geq \left(n\lambda\right)^{\frac{2q-2}{m(2q-1)}}$, assuming $n\lambda > 1$.  Substituting this into \eqref{eq:pert_prob_bound} yields \eqref{eq:u_prob_bound}. 
\end{proof}

\begin{rems}There is an interesting trade-off controlled by $q$. Taking $q$ larger yields a faster rate of convergence in \eqref{eq:u_convergence_rate}, but one that only applies to larger $u$ in \eqref{eq:u_lower_bound}. As $q \to 1^{+}$ we approach a very natural lower bound, $\|u\| \geq \left(n\lambda\right)^{-1/m}$, but the probability bound \eqref{eq:u_prob_bound} becomes constant with respect to $n$, and the metric approximation error \eqref{eq:u_convergence_rate} becomes large.  We shall focus exclusively on the $q=2$ case henceforth, as this choice leads to the best spectral convergence rates. 
\end{rems}

Our PPP metric approximation result (Theorem \ref{thm:metric_approx}) puts together the pieces discussed in the preceding sections to conclude that locally $\mu\L_p^p(x,y)$ is well approximated by $\tilde{\ell}_p^p(x,y,H_{n\rho})$: 

\vspace{10pt}

\begin{proof}[Theorem \ref{thm:metric_approx}]
	Let $\tilde{C}_1, \tilde{C}_2$ be as in Lemma \ref{lem:metric_approx_tangent_plane} and let $\L$ denote $\L_p^p(x,y)$. We have, for $\|u\|\leq C_{\M,\rho,p}$,
	\begin{align*}
		&\tilde{\ell}_p^p(x,y,H_{n\rho}) \\
		=& \tilde{\ell}_p^p(x,y,H_{n\rho} \cap \mathcal{B}_x(R)) \quad (\text{Lemma \ref{lem:local_paths}, prob. }1-p_1) \\
		=&\left(1\pm Cp(mK+\reach^{-2})\|u\|^2\right)\tilde{\ell}_p^p(0,u,H_{ng_x}) \quad (\text{Lemma \ref{lem:curv_pert_dis}}) \\
		=& \left(1\pm Cp(mK+\reach^{-2})\|u\|^2\right)\left( \mu\mathcal{L}_p^p(0,u,g_x) \pm  \tilde{C}_1\L^2 \pm \tilde{C}_2\L^3\right) \quad (\text{Lemma \ref{lem:metric_approx_tangent_plane}, prob. } 1-p_2) \\
		=& \left(1\pm Cp(mK+\reach^{-2})\|u\|^2\right)\left( \left(1\pm Cp(mK+\reach^{-2})\|u\|^2\right)\mu\L \pm  \tilde{C}_1\L^2 \pm \tilde{C}_2\L^3\right) \quad(\text{Lemma \ref{lem:curv_pert_cont}}) \\
		=& \mu\L \pm \tilde{C}_1 \L^2 + \tilde{C}_2'\L^3,
	\end{align*}
where $\tilde{C}_2' = C_{p,d,\beta}\left(\reach^{-2} + K(1+L_1) +L_1^2 + L_2 \right)$
now includes the reach, $p_1 \leq p_2=C_\epsilon n\exp\left(-c_\epsilon(\frac{n}{2\beta})^{\frac{2\epsilon}{3}\min\{\frac{1}{m},\frac{1}{p}\}}\right)$, and we have utilized $\|u\|^2 =\rho(x)^{\frac{2(p-1)}{m}}\L^2 +O (\L^3)$ (see proof of Lemma \ref{lem:metric_approx_tangent_plane}).
\end{proof}

We note Theorem \ref{thm:metric_approx} is closely related to existing results in the Fermat distance literature, especially Theorem 1 in \cite{Hwang2016_Shortest} and Theorem 2.3 in \cite{Groisman2022nonhomogeneous}.  Our results are local (showing $\tilde{\ell}_p^p(x_i, x_j) \approx \mu\L_p^p(x_i, x_j)$ when $x_i,x_j$ are close) but quantitative, giving a sharp third order estimate on the local deviations. 
In contrast, \cite{Hwang2016_Shortest} and \cite{Groisman2022nonhomogeneous} provide macroscale/global convergence results, i.e. $x_i,x_j$ are arbitrary points in $\M$, but of an asymptotic nature.
At a more technical level, we note that from the proof of \cite[Lemma 10]{Hwang2016_Shortest}, a result similar to Lemma \ref{lem:curv_pert_dis} can be deduced:
	\begin{align*}
		\tilde{\ell}_p^p(x,y,H_{n\rho} \cap \mathcal{B}_x(R)) &= \left(1\pm \delta\right)\tilde{\ell}_p^p(0,u,H_{ng_x}).
	\end{align*}

\section{Continuum Limits of Graph Laplace Operators}
\label{sec:FermatLaplacians}

In this section we show that \emph{discrete graph Laplacians built from discrete Fermat distances} converge to \emph{continuum Laplace operators built from continuum Fermat distances}. While this analysis is similar to the arguments of \citet{burago2015graph} and \citet{Trillos2019_Error}, new ideas are needed. Indeed, 
\begin{enumerate}
    \item In \cite{burago2015graph} convergence of eigenvectors of $\Delta_{\Gamma^{d_g,h}}$ to eigenfunctions of $\Delta_{0}$ associated to $(\M,g)$ is studied. Here, $\Gamma^{d_g,h}$ denotes the graph constructed using the kernel $\frac{1}{\omega_{m}}\mathbb{1}_{[0,1]}$, bandwidth parameter $h$, and the geodesic distance$d_g$. In this case the graph weights are degree-normalized and not chosen as in \eqref{eq:WeightsRGG}.
    \item In \cite{Trillos2019_Error} convergence of eigenvectors of $\Delta_{\Gamma^{d_0,h}}$ to eigenfunctions of the random walk Laplacian $\Delta_{2}$ associated to $(\M,g)$ is studied. Here, $\Gamma^{d_0,h}$ denotes the graph constructed using a kernel $\eta$ satisfying certain mild assumptions, bandwidth parameter $h$, and Euclidean distance function $d_0(x,y)=\|x - y\|_2$. 
\end{enumerate}
Like \cite{Trillos2019_Error} our discrete and continuum operators are defined using two different metrics. Unlike \cite{Trillos2019_Error}, the relationship between these two metrics is not straightforward. Indeed, the key new ingredient is the refined approximation bounds $\mu\L_p^p(x,y) \approx \tilde{\ell}_p^p(x,y)$ of Section~\ref{sec:metric_approx}. Nonetheless, we follow the approach outlined in \cite{Trillos2019_Error}. As many of the required technical lemmas can be applied with only minor modifications instead of reproving them we indicate how the proof presented in \cite{Trillos2019_Error} should be modified.

We begin by summarizing some relevant definitions and notation. Let $\nu_{n} := \frac{1}{n}\sum_{i=1}^{n}\delta_{x_i}$ denote the \emph{empirical measure} of $\X_n$. Let $T: \M \to \X$ denote a {\em transportation map} from $\nu$ to $\nu_n$, {\em i.e.,} $T_{\#}\nu = \nu_n$ where the pushforward measure $T_{\#}\nu$ is defined via $T_{\#}\nu(U) = \nu(T^{-1}(U))$. We define 
\begin{equation}
    d_\infty^g(\nu, \nu_n) := \esssup_{x \in \M}d_{g}(x,T(x)),
    \label{eqn:VepsT}
\end{equation}
and later on we use \cite[Theorem 2]{Trillos2019_Error} to quantify $d_\infty^g(\nu, \nu_n)$. For now, we note that we may assume $d_\infty^g(\nu, \nu_n) \ll h$. $T$ induces a partition $\{U_i\}_{i=1}^n$ of $\M$ via $U_i = T^{-1}(x_i)$. Using this partition, we define the discretization operator:
\begin{align}
\begin{split}
    & P: L^2\left(\M,\nu\right) \to L^2\left(\X,\nu_{\X}\right), \\
    & (Pf)(x_i) = n \int_{U_i}f(x)d\nu(x).
    \label{eq:Def_Discretization}
\end{split}
\end{align}
Henceforth we assume $T$, $d_\infty^g(\nu, \nu_n)$, and $P$ are fixed. Finally, we define a non-local energy
\begin{equation*}
    \tilde{E}_r(f) = \int_{\M}\int_{\M} \eta\left(\frac{d_{g}(x,y)}{r}\right)\left|f(y) - f(x)\right|^2d\nu(x)d\nu(y),
\end{equation*}
which will be used to approximate the continuous Dirichlet energy.


\

\subsection{Technical Results for Kernels and Degrees}
We first derive some supporting results regarding Fermat kernels and degrees; see Appendix \ref{app:FermatKernelsAndDegrees} for the proofs. The following corollary applies Theorem \ref{thm:metric_approx_iid} to kernels and is critical for what follows.

\begin{restatable}{cor}{corkernelcomp}
    \label{cor:kernel_comp}
     Let $\delta := 2\mu^{-1} C_1h+4\mu^{-1} C_2h^2 \leq \frac{1}{2}$, where $C_1, C_2$ are as in Theorem \ref{thm:metric_approx_iid}. Define $\widehat{h}_+ := h(1+\delta),\,  \widehat{h}_- := h(1-\delta)$.  Then for $n$ large enough, with probability at least $1-C_\epsilon n^3\exp\left(-c_\epsilon (\frac{n}{4\beta})^{\frac{\epsilon}{2p+1}\min\{\frac{1}{m},\frac{1}{p}\}} \right)$, we have for all $x_i,x_j \in \X$:
    \begin{align}
    \label{equ:kernel_sandwich}
        \eta \left(\frac{\mu\mathcal{L}_p^p(x_i,x_j)}{\widehat{h}_-}\right) \leq \eta \left(\frac{\tilde{\ell}_p^p(x_i,x_j)}{h}\right) &\leq \eta \left(\frac{\mu\mathcal{L}_p^p(x_i,x_j)}{\widehat{h}_+}\right),
    \end{align}
    where $C_{\epsilon},c_{\epsilon}$ are constants depending on $\epsilon$, $\eta = \frac{1}{\omega_m}\mathbb{1}_{[0,1]}$ and $h \geq 4\mu\beta^{\frac{(p-1)}{m}}(n\beta/2)^{-\frac{1}{m}(\frac{1}{3}-\epsilon)}$.
\end{restatable}
The following lemma is a minor modification of Lemma 18 in \cite{Trillos2019_Error} and bounds Fermat degrees.
\begin{restatable}{lem}{degapprox}
\label{lem:Degree_Approx}
    \begin{align*}
       \left| m_i - \rho_p(x_i) \right| &\leq C\beta^p\left[ \left(\frac{L_{1}}{\beta}+C_1m\right)h + m(K_p + C_2\mu)h^2 + m\omega_m\frac{d_\infty^g(\nu, \nu_n)}{h}\right]
   \end{align*}
   for $C_1, C_2$ as in Theorem \ref{thm:metric_approx_iid} and $K_p$ as in Theorem~\ref{thm:New_Sec_Curv}.
\end{restatable}

\subsection{Analysis of Discrete Dirichlet Energies}
\label{sec: Discrete Dirichlet energies}

Next, we consider the relationship between the discrete Dirichlet energies induced by $d_g$ and $d_0$ respectively, where $d_0$ is any distance function approximating $d_g$ as quantified by the following assumption:

\begin{assump}
    \label{assump:Discrete_Compare_Distance}
    Let $d_g$ be the geodesic distance function on $(\M,g)$ while $d_0$ is any distance function on $\X_n$. Suppose that, for appropriate kernel sizes $h > 0$ there  exists $\delta := \delta(h)$ and $\varepsilon_n \to 0^{+}$ such that for all $x_i,x_j \in \X_n$:
    \begin{equation*}
        \eta\left(\frac{d_g(x_i,x_j)}{\widehat{h}_{-}}\right) \leq \eta\left(\frac{d_0(x_i,x_j)}{h}\right) \leq \eta\left(\frac{d_g(x_i,x_j)}{\widehat{h}_{+}}\right) 
    \end{equation*}
    holds with probability $1 - \varepsilon_n$ where $\widehat{h}_{+} = h(1+\delta)$ and $\widehat{h}_{-} = h(1-\delta)$.  
\end{assump}

  For brevity, for $c=0,g$ we write $w^{c,h}_{i,j}$ instead of $w^{d_c,h}_{i,j}$, $b^{c,h}(\cdot)$ instead of $b^{d_c,h}(\cdot)$ and so on; see Section~\ref{subsec:Laplace_Operators} for definitions.
  
\begin{thm}[General Dirichlet Energy Perturbation]
\label{thm:Distance_perturb_Dirichlet}
Suppose Assumption~\ref{assump:Discrete_Compare_Distance} holds for $d_0, d_g$ and $h > 0$. Then, for all $u:\X \to \mathbb{R}$ and with probability $1-\varepsilon_n$,
\begin{equation*}
   \left(1-Cm\delta\right)b^{g,\widehat{h}_{-}}(u) \leq b^{0,h}(u) \leq \left(1+ Cm\delta\right)b^{g,\widehat{h}_{+}}(u).  
\end{equation*}
\end{thm}

\begin{proof}
We simply compare the Dirichlet energies:
\begin{align*}
    b^{g,\widehat{h}_{-}}(u)  &= \frac{m+2}{n\widehat{h}_{-}^2}\sum_{i,j}w^{g,\widehat{h}_{-}}_{i,j}\left(u(x_i) - u(x_j)\right)^2 \\
    &= \frac{m+2}{n^2\widehat{h}_{-}^{m+2}}\sum_{i,j} \eta\left(\frac{d_g(x_i,x_j)}{\widehat{h}_{-}}\right)\left(u(x_i) - u(x_j)\right)^2 \\
    & \stackrel{(a)}{\leq} \left(\frac{1}{1-\delta}\right)^{m+2} \frac{m+2}{n^2h^{m+2}} \sum_{i,j}\eta\left(\frac{d_0(x_i,x_j)}{h}\right)\left(u(x_i) - u(x_j)\right)^2 \\ 
\Rightarrow \left(1 - \delta\right)^{m+2} b^{g,\widehat{h}_{-}}(u)  & \leq \frac{m+2}{nh^2} \sum_{i,j} w_{i,j}^{0,h}\left(u(x_i) - u(x_j)\right)^2 = b^{0,h}(u),
                 \end{align*}
where we use Assumption~\ref{assump:Discrete_Compare_Distance} in (a). For $\delta$ small enough $(1-\delta)^{m+2} \leq 1 - Cm\delta$ and the stated bound follows. The proof of the upper bound is similar. 
\end{proof}
Next, we relate the corresponding eigenvalues. 
\begin{thm}
\label{thm:Eigenvalues_Dirichlet_near}
Suppose Assumption~\ref{assump:Discrete_Compare_Distance} holds for $d_0, d_g$ and $h > 0$. Recall that $\lambda_k(\Delta_{\Gamma^{c,h}})$ denotes the $k$-th eigenvalue of the random walk Laplacian of the graph $\Gamma^{c,h} := (\X,W^{c,h})$ for $c=0,g$. Then with probability $1-\varepsilon_n$
 \begin{align*}
\lambda_k(\Delta_{\Gamma^{2,h}})  & \leq \left(1 + Cm\delta + C\beta L_{1}h + C\beta^2\frac{d_\infty^g(\nu, \nu_n)}{h} + C_{\beta,m,L_{1}}h^2 + C_{m,\beta, L_{1}}\delta h\right)\lambda_k(\Delta_{\Gamma^{1,\widehat{h}_{+}}}), \\
\lambda_k(\Delta_{\Gamma^{2,h}}) & \geq 
 \left(1 - Cm\delta - C\beta L_{1}h - C\beta^2\frac{d_\infty^g(\nu, \nu_n)}{h} - C_{\beta,m,L_{1}}h^2 - C_{m,\beta, L_{1}}\delta h\right)\lambda_k(\Delta_{\Gamma^{1,\widehat{h}_{-}}}), 
 \end{align*}
 where $d_\infty^g(\nu, \nu_n)$ is as defined in \eqref{eqn:VepsT}.
\end{thm}

\begin{proof}
   As discussed in Section~\ref{subsec:Laplace_Operators},
   \begin{equation*}
       \lambda_k(\Delta_{\Gamma^{c,h^\prime}}) = \min_{L_k} \max_{u \in L_k\setminus\{0\}} \frac{b^{c,h^\prime}(u)}{\|u\|^2_{c,\mathbf{m}}} \quad \text{ for } c=0,g \text{ and } h^\prime > 0.
   \end{equation*}
First, we compare degrees, which we explicitly decorate with a value of $h$. From Assumption~\ref{assump:Discrete_Compare_Distance} we deduce that
\begin{align*}
m_{g,\widehat{h}_{-},i} &:= \frac{1}{n\widehat{h}_{-}^m}\sum_{j=1}^n \eta\left(\frac{d_g(x_i,x_j)}{\widehat{h}_{-}}\right) \leq \left(\frac{1}{1-\delta}\right)^{m+2}\frac{1}{nh^m} \sum_{j=1}^n \eta\left(\frac{d_0(x_i,x_j)}{h}\right) \\
&\leq \left(1 + Cm\delta\right)\frac{1}{nh^m}\sum_{j=1}^n \eta\left(\frac{d_0(x_i,x_j)}{h}\right) = \left(1 + Cm\delta\right) m_{0,h,i},
\end{align*}
and similarly $m_{g,\widehat{h}_{+},i} \geq \left(1 - Cm\delta\right) m_{0,h,i}$, where both hold with probability $1-\varepsilon_{n}$. 
Using \cite[Lemma 18]{Trillos2019_Error} we get
\begin{align*}
\max_{i=1, \dots, n} | m_{g, \widehat{h}_-,i } - m_{g, \widehat{h}_+,i } | & \leq \max_{i=1, \dots, n} | m_{g, \widehat{h}_-,i } - \rho(x_i) | + \max_{i=1, \dots, n} | m_{g, \widehat{h}_+,i } - \rho(x_i) |     
\\& \leq CL_{1} h + C\beta\frac{d_\infty^g(\nu, \nu_n)}{h} + C\beta mKh^2.
\end{align*}
Also, letting $i^{\star} := \min_{i=1,\ldots,n}m_{g, \widehat{h}_{\pm},i}$,
\begin{align*}
    \min_{i=1,\ldots,n}m_{g, \widehat{h}_{\pm},i} &= \rho(x_{i^{\star}}) + \left(m_{g, \widehat{h}_{\pm},i^{\star}} - \rho(x_{i^{\star}})\right) \\
    & \geq \frac{1}{\beta} - CL_1 h - C\beta\frac{d_\infty^g(\nu, \nu_n)}{h} - C\beta mKh^2.
\end{align*}
for $d_\infty^g(\nu, \nu_n)$ as defined in \eqref{eqn:VepsT}. So,
\begin{align*}
    m_{0,h,i} & \geq \left(1 - Cm\delta\right)\left(m_{g,\widehat{h}_+,i} - CL_1 h - C\beta\frac{d_\infty^g(\nu, \nu_n)}{h} - C\beta mKh^2  \right)\\
        & \geq \left(1 - Cm\delta\right)\left(m_{g,\widehat{h}_+,i} -  \frac{m_{g,\widehat{h}_+,i}}{\min_{i=1,\ldots,n}m_{g, \widehat{h}_{\pm},i}}\left(CL_1 h + C\beta\frac{d_\infty^g(\nu, \nu_n)}{h} + C\beta mKh^2\right)\right) \\
        & \geq \left(1 - \left(Cm\delta + C\beta L_1h + C\beta^2\frac{d_\infty^g(\nu, \nu_n)}{h} + C_{\beta,m,L_1}h^2 + C_{m,\beta, L_1}\delta h\right)\right)m_{\widehat{h}_{+},i} \\
        & := \left(1 - E\right)m_{\widehat{h}_{+},i}
\end{align*}
and similarly, $m_{0,h,i} \leq (1+E)m_{\widehat{h}_{-},i}$, where we have ignored certain higher order terms.
We now compare norms.
\begin{align*}
   \|u\|_{0,\mathbf{m},h}^2 & = \sum_{i=1}^nm_{0,h,i}u_i^2 \leq \sum_{i=1}^n \left(1 + E\right)m_{\widehat{h}_{-},i}u_i^2 = (1+E) \|u\|_{g,\mathbf{m},\widehat{h}_{-}}^2
\end{align*}
and similarly $\|u\|_{0,\mathbf{m},h}^2 \geq \left(1 - E\right)\|u\|_{g,\mathbf{m},\widehat{h}_{+}}^2$.
Now fix any subspace $L_k$ of dimension $k$. 
\begin{align*}
    \max_{u \in L_k\setminus\{0\}} \frac{b^{0,h}(u)}{\|u\|^2_{0,\mathbf{m},h}} 
    & \leq (1+ E)\max_{u \in L_k\setminus\{0\}} \frac{b^{0,h}(u)}{\|u\|^2_{g,\mathbf{m},\widehat{h}_{+}}} \leq (1+E)(1+Cm\delta)\max_{u \in L_k\setminus\{0\}} \frac{b^{g,\widehat{h}_+}(u)}{\|u\|^2_{g,\mathbf{m},\widehat{h}_{+}}}.
\end{align*}
But $L_k$ was arbitrary, and so
\begin{align*}
 \lambda_k(\Delta_{\Gamma^{2,h}}) &= \min_{L_k \, : \, \dim(L_k) = k} \, \max_{u \in L_k\setminus\{0\}} \frac{b^{0,h}(u)}{\|u\|^2_{0,\mathbf{m},h}} \\
 & \leq (1+E)(1+Cm\delta)\min_{L_k \, : \, \dim(L_k) = k}\,\max_{u \in L_k\setminus\{0\}} \frac{b^{g,\widehat{h}_{+}}(u)}{\|u\|^2_{g,\mathbf{m},\widehat{h}_{+}}} \\
 &= (1+E)(1+Cm\delta)\lambda_k(\Delta_{\Gamma^{2,\widehat{h}_{+}}}) \, .
\end{align*}
Simplifying $(1+E)(1+Cm\delta)$, ignoring higher order terms {\em e.g.} $\delta h^2, \delta d_{\infty}^g(\nu, \nu_n)$, and adjusting constants as necessary yields the stated upper bound. A similar argument yields the lower bound.
\end{proof}

\red

\nc

\subsection{Discrete to Continuum Convergence of Eigenvalues}
\label{sec:SpectralConvergenceDiscreteToContinuum}

With $d_0$ as in Section \ref{sec: Discrete Dirichlet energies} we now show the eigenvalues of $\Delta_{\Gamma^{d_0,h}}$ converge to those of $\Delta_2$ provided Assumption~\ref{assump:Discrete_Compare_Distance} holds.
Later, we specialize this to $d_{0} = \tilde{\ell}_p^p$ and $d_{g} = \mu\L_p^p$.  

\begin{lem}[cf. Lemma 13, part (ii) \cite{Trillos2019_Error}]
\label{lemma_13_replacement}
Suppose Assumption~\ref{assump:Discrete_Compare_Distance} holds. Then with probability $1-\varepsilon_n$, for all $f \in H^1(\M)$,
\begin{equation*}
    b^{d_0}_h(Pf) \leq \left(1+C\beta L_1(1+\delta)h + Cm \frac{d_\infty^g(\nu, \nu_n)}{(1+\delta)h} + CmK_p(1+3\delta)h^2\right)D(f)\, .
\end{equation*}

\end{lem}

\begin{proof}
    We indicate how to modify the proof. Recalling $\sigma_{\eta}$ in their notation is $\frac{1}{m+2}$:
    \begin{align}
        b^{d_0}_h(Pf) &\leq \frac{m+2}{h^{m+2}} \sum_i\sum_j \int_{U_i}\int_{U_j} \eta\left(\frac{d_0(x_i, x_j)}{h}\right)\left|f(y) - f(x)\right|^2d\nu(y)d\nu(x) \label{eq:32}\\
        & \leq \frac{m+2}{h^{m+2}} \sum_i\sum_j \int_{U_i}\int_{U_j} \eta\left(\frac{d_g(x_i, x_j)}{\widehat{h}^+}\right)\left|f(y) - f(x)\right|^2d\nu(y)d\nu(x) \label{eq:Use_metric_closeness}\\
        & \leq \frac{m+2}{h^{m+2}} \int_{\M}\int_{\M} \eta\left(\frac{(d_g(x,y) - 2d_\infty^g(\nu, \nu_n))_{+}}{\widehat{h}^+}\right)\left|f(y) - f(x)\right|^2d\nu(y)d\nu(x) \label{eq:use_Ui_Eps} \\
        & = \frac{m+2}{h^{m+2}} \int_{\M}\int_{\M} \eta\left(\frac{d_g(x,y)}{\widehat{h}^+ + 2d_\infty^g(\nu, \nu_n)} \right)\left|f(y) - f(x)\right|^2d\nu(y)d\nu(x) \label{eq:eta_ineq_NGT} \\
        & = \frac{m+2}{h^{m+2}} \tilde E_{\widehat{h}^+ + 2d_\infty^g(\nu, \nu_n)}, \nonumber
    \end{align}
    where \eqref{eq:32} is shown in the proof of Lemma 13, part (ii) \cite{Trillos2019_Error}, \eqref{eq:Use_metric_closeness} follows from Assumption~\ref{assump:Discrete_Compare_Distance}, \eqref{eq:use_Ui_Eps} follows as the $U_i$ have diameter at most $2d_\infty^g(\nu, \nu_n)$, and \eqref{eq:eta_ineq_NGT} follows from the identity $\eta\left(\frac{(t-s)_{+}}{r}\right) = \eta\left(\frac{t}{r+s}\right)$ for $r,t,s > 0$, see \cite[Lemma 10]{Trillos2019_Error}. From \cite[Lemma 5]{Trillos2019_Error}:
    \begin{align*}
        \tilde{E}_{\widehat{h}^+ + 2d_\infty^g(\nu, \nu_n)}(f)
        \leq & \left(1 + L_1\beta(\widehat{h}^+ \!+\! 2d_\infty^g(\nu, \nu_n))\right)\left(1 + CmK\left(\widehat{h}^+ \!+\! 2d_\infty^g(\nu, \nu_n)\right)^2\right) \\
        & \left(\widehat{h}^+ \!+\! 2d_\infty^g(\nu, \nu_n)\right)^{m+2}\frac{D(f)}{m+2}.
    \end{align*}
    Simplifying as in \cite{Trillos2019_Error} and using $d_\infty^g(\nu, \nu_n) \ll h < \widehat{h}^{+} := h(1+\delta)$ yields the stated bound.
\end{proof}
\begin{proof}[Proof of Theorem \ref{thm:Eigenvalue_convergence}] By Corollary~\ref{cor:kernel_comp}, as long as
\begin{equation}\label{eq:eps_lower_bound_h}
h \geq 4\mu\beta^{\frac{(p-1)}{m}}(n\beta/2)^{-\frac{1}{m}(\frac{1}{3}-\epsilon)},
\end{equation}
Assumption~\ref{assump:Discrete_Compare_Distance} holds with $d_g = \mu\L_p^p$ and $d_0 = \tilde{\ell}_p^p$ where 
\begin{equation}
 \delta = 2\mu^{-1} C_1h+4\mu^{-1} C_2h^2.
 \label{eq:formula_for_delta}
\end{equation}
From Theorem~\ref{thm:Eigenvalues_Dirichlet_near}, 
\begin{equation}
\lambda_k(\Gamma^{\tilde{\ell},h}) - \lambda_k(\Gamma^{\L,\widehat{h}_{-}}) \geq -E\lambda_k(\Gamma^{\L,\widehat{h}_{-}}) \text{ and } \lambda_k(\Gamma^{\tilde{\ell},h}) - \lambda_k(\Gamma^{\L,\widehat{h}_{+}}) \leq E\lambda_k(\Gamma^{\L,\widehat{h}_{+}}) \label{eq:lambda_discrete_to_cont}
\end{equation}
where 
\begin{equation}
    E = Cm\delta + C\beta L_1h + C\beta^2\frac{d_\infty^g(\nu, \nu_n)}{h} + C_{\beta,m,L_1}h^2 + C_{m,\beta, L_1}\delta h.
    \label{eq:formula_for_E}
\end{equation}
From \cite[Theorem 4, part 1]{Trillos2019_Error} we have
\begin{equation}
\label{eq:lambda_Trillos_1}
    \lambda_k(\Gamma^{\L,\widehat{h}_{+}}) -  \lambda_k(\Delta_{2,p}) \leq \delta_1\lambda_k(\Delta_{2,p}),
\end{equation}
where
\begin{align}
    \delta_1 = \tilde{C}\left[L_1\widehat{h}_{+} + \frac{d_\infty^g(\nu, \nu_n)}{\widehat{h}_{+}} + \sqrt{\lambda_k(\Delta_{2,p})}d_\infty^g(\nu, \nu_n) + K_p\widehat{h}_{+}^2 + \|\mathbf{m} - \rho_p\|_{\infty}\right].
    \label{eq:formula_for_delta_1}
\end{align}
where $\tilde{C}$ depends only on $m,\beta,p$ and $L_1 $. We do not retain the term proportional to $h^2/R^2$ in \cite[eq. (1.16)]{Trillos2019_Error}, as $\Gamma^{\L,\widehat{h}_{+}}$ is constructed using geodesic distances between points, not Euclidean distances. More precisely, this discrepancy arises from using Lemma~\ref{lemma_13_replacement} in place of \cite[Lemma 13]{Trillos2019_Error}. Similarly, from \cite[Theorem 4, part 2]{Trillos2019_Error} we have
\begin{equation}
\lambda_k(\Gamma^{\L,\widehat{h}_{-}}) -  \lambda_k(\Delta_{2,p}) \geq - \delta_2\lambda_k(\Delta_{2,p}),
    \label{eq:lambda_Trillos_2}
\end{equation}
where
\begin{align}
    \delta_2 &= \tilde{C}\left[L_1\widehat{h}_{-} + \frac{d_\infty^g(\nu, \nu_n)}{\widehat{h}_{-}} + \sqrt{\lambda_k(\Delta_{2,p})}d_\infty^g(\nu, \nu_n) + K_p\widehat{h}_{-}^2 + \|\mathbf{m} - \rho_p\|_{\infty}\right].
    \label{eq:formula_for_delta_2}
\end{align}
 Combining \eqref{eq:lambda_Trillos_1} and \eqref{eq:lambda_discrete_to_cont} yields:
 \begin{align}
 \begin{split}
\lambda_k(\Gamma^{\tilde{\ell},h}) - \lambda_k(\Delta_{2,p}) &= \lambda_k(\Gamma^{\tilde{\ell},h}) - \lambda_k(\Gamma^{\L,\widehat{h}_{+}}) + \lambda_k(\Gamma^{\L,\widehat{h}_{+}}) - \lambda_k(\Delta_{2,p}) \\
    & \leq E\lambda_k(\Gamma^{\L,\widehat{h}_{+}}) + \delta_1\lambda_k(\Delta_{2,p}) \\
    & \leq E(1+\delta_1)\lambda_k(\Delta_{2,p}) + \delta_1\lambda_k(\Delta_{2,p}) \\
    &= \left(E + \delta_1 + E\delta_1\right)(\Delta_{2,p}).
    \label{eq:lambda_error_upper_bound}
    \end{split}
 \end{align}
Similarly, combining \eqref{eq:lambda_Trillos_2} and \eqref{eq:lambda_discrete_to_cont}
 \begin{equation}
      \lambda_k(\Gamma^{\tilde{\ell},h}) - \lambda_k(\Delta_{2,p}) \geq -\left(E + \delta_2 + E\delta_2\right)\lambda_k(\Delta_{2,p}) \, .
      \label{eq:lambda_error_lower_bound}
 \end{equation}
 Using \eqref{eq:formula_for_delta}, \eqref{eq:formula_for_E}, \eqref{eq:formula_for_delta_1} and \eqref{eq:formula_for_delta_2} the error terms in \eqref{eq:lambda_error_upper_bound} and \eqref{eq:lambda_error_lower_bound} can be expressed as
 \begin{align*}
     E + \delta_1 + E\delta_1 &= C_{\beta,p,L_1,m}h + C\beta^2 \frac{d_{\infty}^g(\nu,\nu_n)}{h} + C_{\beta,m,L_1, L_2,p, K_p,\reach}h^2 + O(h^3 + d_{\infty}^g(\nu,\nu_n)), \\
     E + \delta_2 + E\delta_2 &= C_{\beta,p,L_1,m}h + C\beta^2 \frac{d_{\infty}^g(\nu,\nu_n)}{h} + C_{\beta,m,L_1, L_2,p, K_p,\reach}h^2 + O(h^3 + d_{\infty}^g(\nu,\nu_n)).
 \end{align*}
 From \cite[Theorem 2]{Trillos2019_Error} we have
 \begin{equation*}
     d_\infty^g(\nu, \nu_n) = O\left(\frac{\log(n)^{p_m}}{n^{1/m}}\right)
 \end{equation*}
 w.h.p, where $p_m = 3/4$ if $m=2$ and $p_m = 1/m$ if $m \geq 3$. In \cite{Trillos2019_Error} contributions to the approximation error ($\delta_1$ and $\delta_2$ in our notation) from $h$ and $d_\infty^g(\nu, \nu_n)$ are optimally balanced by setting $h = \sqrt{d_\infty^g(\nu, \nu_n)} = \mathcal{O}\left(\log(n)^{p_m}n^{-\frac{1}{2m}}\right)$. We cannot do so here as $h$ must satisfy the lower bound of \eqref{eq:eps_lower_bound_h}. Instead, we take $h = 4\mu\beta^{\frac{(p-1)}{m}}(n\beta)^{-\frac{1}{m}(\frac{1}{3}-\epsilon)}= O(n^{-\frac{1}{3m} + \frac{\epsilon}{m}})$. Noting that $K_p$ may be bounded by an expression involving $K,L_1,L_,m$, and $p$ (see Appendix \ref{App:Sec_Curv}) we may write $C_{\beta,m,L_1, L_2,p, K,\reach}$ instead of $C_{\beta,m,L_1, L_2,p, K_p,\reach}$. We now identify the leading and second-order terms in \eqref{eq:lambda_error_upper_bound}  and \eqref{eq:lambda_error_lower_bound}. Observe that 
 \begin{equation*}
     h^2 = O(n^{-\frac{2}{3m} + \frac{2\epsilon}{m}}) \gg d_\infty^g(\nu, \nu_n) \text{ and } h^2 \gg O(n^{-\frac{2}{3m} - \frac{\epsilon}{m}}) = \frac{d_\infty^g(\nu, \nu_n)}{h}\, .
 \end{equation*}
 retaining only terms proportional to $h$ and $h^2$ yields the stated error bounds.
\end{proof}

\begin{rems}
The dependence of $h$ on $n$ is determined by Corollary~\ref{cor:kernel_comp}, which is in turn determined by Proposition~\ref{prop:conv_PD_homog_PPP}. One might enquire as to whether a better choice of $q$ in Proposition~\ref{prop:conv_PD_homog_PPP} could result in a better error bound in Theorem~\ref{thm:Eigenvalue_convergence}. In fact it is straightforward (but tedious) to trace the dependency on $q$ through Theorem~\ref{thm:metric_approx} and Corollary~\ref{cor:kernel_comp} to the proof of  Theorem~\ref{thm:Eigenvalue_convergence}, whence a straightforward computation reveals $q=2$ is optimal.
\end{rems}

\begin{rems}
Although we state the above theorem in terms of (discrete and continuous) Fermat distance, with superficial modifications it can be applied to any Riemannian distance $d_{g}(\cdot, \cdot)$ and any discrete approximation $d_0(\cdot,\cdot)$ satisfying Assumption~\ref{assump:Discrete_Compare_Distance}.
\end{rems}

\subsection{Discrete to Continuum Convergence of Eigenvectors}
\label{sec:EigenvecConvergenceMain}

In this section we relate eigenvectors of the Fermat graph Laplacian $\Delta_{\Gamma^{\tilde{\ell},h}}$ and the eigenfunctions of the operator $\Delta_{2,p}$ by proving Theorem \ref{thm:eigenvector}. The proof of this result uses standard arguments in the literature and relies on the relationship between Dirichlet energies that we established in previous sections as well as on the consistency of eigenvalues from Theorem \ref{thm:Eigenvalue_convergence}. 

\begin{proof}[Proof of Theorem \ref{thm:eigenvector}]  Throughout this proof let $\psi^n_1,\ldots, \psi^n_n$ denote the  eigenvectors of $\Delta_{\Gamma^{\tilde{\ell},h}}$ normalized according to $\lVert \cdot \rVert^2_{\textbf{m}}$. Let $f$ be an eigenfunction of $\Delta_{2,p}$ with eigenvalue $\lambda_2 := \lambda_2(\Delta_{2,p})$, normalized according to $\lVert \cdot \rVert^2_{L^2(\M, \rho_p^2  \V_p)}$ . Recall that, by assumption, this eigenfunction is unique up to sign. From Lemma \ref{lemma_13_replacement} we know that
\begin{equation}
    b^{\tilde{\ell}}_h(Pf) \leq \left(1+C\beta L_1h + Cm \frac{d_\infty^g(\nu, \nu_n)}{h} + CmK_ph^2\right)\lambda_2,
    \label{eq:Dirichlet_Pf}
\end{equation}
and from the proof of Lemma 13 in \cite{Trillos2019_Error} we know that
\begin{align}
\begin{split}
  \left| \lVert P f \rVert^2_{\textbf{m}} - 1 \right| & =  \left| \lVert P f \rVert^2_{\textbf{m}}  - \lVert f \rVert^2_{L^2(\M, \rho_p^2  \V_p)} \right|  \\ & \leq C ( \lVert \textbf{m} - \rho_p \rVert_\infty  + d_\infty^g(\nu, \nu_n)  ) 
   + C d_\infty^g(\nu, \nu_n)\lambda_2^{1/2}. 
  \end{split}
  \label{aux:Eigen55}
\end{align}
Without loss of generality we can take $\psi_1^n = \|\mathbf{1}_{\X}\|_{\textbf{m}}^{-1} \mathbf{1}_{\X}$. Let $\langle \cdot,\cdot\rangle_{\textbf{m}}$ be the inner product associated to $\|\cdot\|_{\textbf{m}}$. Using the definition of the discretization operator $P$ and recalling that $U_i= T^{-1}(x_i)$, it follows
\begin{align*}
 \langle Pf,\mathbf{1}_{\X}\rangle_{\textbf{m}} &= \frac{1}{n}\sum_{i=1}^n Pf (x_i) \textbf{m}_i \\   
 &= \sum_{i=1}^n  \textbf{m}_i  \int_{U_i} f (x)  \rho_p(x) \V_p(x)  
 \\& = \sum_{i=1}^n \int_{U_i}  f(x)  \rho_p^2(x) \V_p(x)  +   \sum_{i=1}^n \int_{U_i}  f(x)  ( \textbf{m}_i - \rho_p(x)  )\rho_p(x) \V_p(x) 
 \\& =:  \sum_{i=1}^n \int_{U_i}  f(x)  \rho_p^2(x) \V_p(x) + c_n
 \\&= c_n,
\end{align*}
where we have used the fact that $f$ is orthogonal to $\mathbf{1}_{\M}$---the first eigenfunction of $\Delta_{2,p}$---w.r.t. the inner product $\langle \cdot , \cdot  \rangle_{L^2(\M, \rho_p^2  \V_p  )}$:
\[   \sum_{i=1}^n \int_{U_i}  f(x)  \rho_p^2(x) \V_p(x) =  \int_\M f(x) \rho_p^2(x) \V_p(x)  = \langle f , \mathbf{1}_{\M}  \rangle_{L^2(\M, \rho_p^2  \V_p  )} = 0 . \]
In turn, the error term $c_n$ can be bounded as
\begin{align}
    |c_n| &\leq \lVert  \textbf{m} - \rho_p\rVert_\infty \int_\M |f(x)| \rho_p(x) \V_p \leq  \beta^{p} \lVert f  \rVert_{L^2(\M,\rho_p^2\V_p)} \lVert  \textbf{m} - \rho_p\rVert_\infty  \nonumber \\
    & = \beta^{p} \lVert  \textbf{m} - \rho_p\rVert_\infty \leq C_{\beta,p,m,L_1}n^{-\frac{1}{3m} + \frac{\epsilon}{m}} + C_{m,\beta,p,L_1,L_2,K}n^{-\frac{2}{3m} + \frac{2\epsilon}{m}} \label{eq:simplifying_c_n}
\end{align}
where we have used Lemma \ref{lem:Degree_Approx}, the value of $h$ given in the assumptions, simplified as in the proof of Theorem \ref{thm:Eigenvalue_convergence}, and ignored higher order terms ({\em i.e.} those $\gg n^{-\frac{2}{3m} + \frac{2\epsilon}{m}}$).
\noindent Now, let $v : \X \rightarrow \R$ be the function $v:=  \left(Pf - c_n\|\mathbf{1}_{\X}\|_{\textbf{m}}^{-1}\psi_1^n\right)$,
which is orthogonal to $\psi_1^n$, and define $\hat{v} = v/\|v\|_{\textbf{m}}$.
Appealing to \eqref{eq:Dirichlet_Pf} it follows that
\begin{align}
   b_h^{\tilde{\ell}}(v) &= b_h^{\tilde{\ell}}(Pf) \leq \left(1+C\beta L_1h + Cm \frac{d_\infty^g(\nu, \nu_n)}{h} + CmK_ph^2\right)\lambda_2 \label{eq:Preliminary_est_b} \\
   & \leq \lambda_2 + C_{\beta,L_1}\lambda_2n^{-\frac{1}{3m} + \frac{\epsilon}{m}} + C_{m,\beta,p,L_1,L_2,K}\lambda_2n^{-\frac{2}{3m} + \frac{2\epsilon}{m}} \label{eq:Prelim_est_b_plug_in_n}
\end{align}
where \eqref{eq:Preliminary_est_b} follows from the invariance of $b_h^{\tilde{\ell}}(\cdot)$ to shifts while \eqref{eq:Prelim_est_b_plug_in_n} follows similarly to \eqref{eq:simplifying_c_n}.
Expanding in the eigenbasis,
\begin{align*}
  \lambda_3(\Gamma^{\tilde{\ell},h}) \lVert \hat{v} - P_2 \hat{v} \rVert^2_{\textbf{m}}  & = \lambda_3(\Gamma^{\tilde{\ell},h}) \sum_{i=3}^n ( \langle  \hat{v} , \psi_i^n \rangle_{\textbf{m}} )^2 
   \leq   \sum_{i=3}^n \lambda_i(\Gamma^{\tilde{\ell},h})  ( \langle  v , \psi_i^n \rangle _{\textbf{m}} )^2 \\ 
   &= b_h^{\tilde{\ell}}(\hat{v}) -  \lambda_2(\Gamma^{\tilde{\ell},h})  \lVert P_2 \hat{v} \rVert^2_{\textbf{m}}.
\end{align*}
It follows that 
\begin{equation}
( \lambda_3(\Gamma^{\tilde{\ell},h})-  \lambda_2(\Gamma^{\tilde{\ell},h})) \lVert \hat{v}- P_2 \hat{v} \rVert^2_{\textbf{m}} +   \lambda_2(\Gamma^{\tilde{\ell},h}) \leq  b_h^{\tilde{\ell}}(\hat{v}). \label{eq:rearrange_for_hat_v} 
\end{equation}
Now observe that 
\begin{align*}
    \lVert \hat{v}- P_2 \hat{v} \rVert^2_{\textbf{m}} &= \frac{1}{\|v\|_{\textbf{m}}^2} \lVert v- P_2v \rVert^2_{\textbf{m}} \\
    & = \frac{1}{\|v\|_{\textbf{m}}^2} \left\|Pf - P_2Pf -  c_n\|\mathbf{1}_{\X}\|_{\textbf{m}}^{-1}\psi_1^n\right\|^2 \\
    & \geq \frac{1}{2\|v\|_{\textbf{m}}^2} \left(\left\|Pf - P_2Pf\right\|^2 - 2c_n^2\|\mathbf{1}_{\X}\|_{\textbf{m}}^{-2} \right)
\end{align*}
Substituting in to \eqref{eq:rearrange_for_hat_v} and rearranging:
\begin{equation}
    \left\|Pf - P_2Pf\right\|^2 \leq \frac{1}{\lambda_3(\Gamma^{\tilde{\ell},h})-  \lambda_2(\Gamma^{\tilde{\ell},h})} \left(2b_h^{\tilde{\ell}}(v) - 2\|v\|_{\textbf{m}}^2\lambda_2(\Gamma^{\tilde{\ell},h}) \right) + 2c_n^2\|\mathbf{1}_{\X}\|_{\textbf{m}}^{-2}
    \label{eq:Proto_evec_theorem}
\end{equation}
where we have used the $2$-homogeneity of $b_h^{\tilde{\ell}}(\cdot)$:
\begin{equation*}
    \|v\|_{\textbf{m}}^2 b_h^{\tilde{\ell}}(\hat{v}) = b_h^{\tilde{\ell}}(\|v\|_{\textbf{m}}\hat{v}) = b_h^{\tilde{\ell}}(v).
\end{equation*}
To conclude we bound the terms in \eqref{eq:Proto_evec_theorem} in terms of powers of $n$. 
From the error estimates of Theorem \ref{thm:Eigenvalue_convergence}
\begin{equation*}
  \lambda_2(\Gamma^{\tilde{\ell},h}) \leq \lambda_2\left(1 + C_{\beta,p,m,L_1} n^{-\frac{1}{3m} + \frac{\epsilon}{m}} + C_{\beta,p,m,L_1,L_2,K_p,\reach}n^{-\frac{2}{3m} + \frac{2\epsilon}{m}}\right)  
\end{equation*}
A similar computation to that done for $\langle Pf,\mathbf{1}_{\X}\rangle_{\textbf{m}}$ reveals that 
\begin{align*}
   \|\mathbf{1}_{\X}\|_{\textbf{m}}^2 &= \frac{1}{n}\sum_{i=1}^n\mathbf{m}_i  \geq  \int_{\M} \rho_p^2 \V_p  - \lVert \textbf{m} - \rho_p \rVert_\infty  \stackrel{(a)}{\geq} \frac{1}{2} \int_{\M} \rho_p^2 \V_p \geq \frac{1}{2}\beta^{-2p},
\end{align*}
where in (a) we have used the fact that $\lVert \textbf{m} - \rho_p \rVert_\infty$ can be assumed to be small enough. Similarly, $\|\mathbf{1}_{\X}\|_{\textbf{m}}^2 \leq \frac{1}{2}\beta^{2p}$ and so
\begin{align*}
    \|v\|_{L^2(\X,\textbf{m}\nu_n)} &= \left\| P f - c_n\|\mathbf{1}_{\X}\|_{L^2(\X, \textbf{m} \nu_n)}^{-1}\psi_1^n\right\|_{L^2(\X, \textbf{m} \nu_n)} \\
    &\geq \lVert Pf \rVert_{L^2(\X, \textbf{m} \nu_n)} - c_n \|\mathbf{1}_{\X}\|_{L^2(\X, \textbf{m} \nu_n)}^{-1} \\
    & \geq 1 - C ( \lVert \textbf{m} - \rho_p \rVert_\infty  + d_\infty^g(\nu, \nu_n)) - C d_\infty^g(\nu, \nu_n)\lambda_2^{1/2} - \frac{c_n}{2} \beta^{2p} \\
    & \geq 1 - C_{\beta,p,m,L_1}n^{-\frac{1}{3m} + \frac{\epsilon}{m}} - C_{m,\beta,p,L_1,L_2,K}n^{-\frac{2}{3m} + \frac{2\epsilon}{m}}
\end{align*}
Using the error estimates of Theorem \ref{thm:Eigenvalue_convergence} again and letting $\lambda_3 := \lambda_3(\Delta_{2,p})$,
\begin{align*}
    \frac{1}{\lambda_3(\Gamma^{\tilde{\ell},h}) -  \lambda_2(\Gamma^{\tilde{\ell},h})} &= \frac{1}{( \lambda_3(\Gamma^{\tilde{\ell},h}) - \lambda_3) + (\lambda_3 - \lambda_2) + (\lambda_2 - \lambda_2(\Gamma^{\tilde{\ell},h})) } \\
    & \leq \frac{1}{\lambda_3 - \lambda_2} - \frac{\lambda_3}{(\lambda_3 - \lambda_2)^2}\left[ C_{\beta,p,m,L_1} n^{-\frac{1}{3m} + \frac{\epsilon}{m}} + C_{\beta,p,m,L_1,L_2,K,\reach}n^{-\frac{2}{3m} + \frac{2\epsilon}{m}} \right] \\
    & \ - \frac{\lambda_2}{(\lambda_3 - \lambda_2)^2}\left[ C_{\beta,p,m,L_1} n^{-\frac{1}{3m} + \frac{\epsilon}{m}} + C_{\beta,p,m,L_1,L_2,K_p,\reach}n^{-\frac{2}{3m} + \frac{2\epsilon}{m}} \right]
\end{align*}
Combining these bounds, and ignoring higher order terms,
\begin{align*}
    \lVert Pf- P_2 (Pf) \rVert^2_{\textbf{m}} \leq & \frac{1}{\lambda_3 - \lambda_2} \left[C_{\beta,p,m,L_1}n^{-\frac{1}{3m} + \frac{\epsilon}{m}} +  C_{\beta,p,m,L_1,L_2,K,\reach}n^{-\frac{2}{3m} + \frac{2\epsilon}{m}} \right]\lambda_2 \\
    & + C_{\beta,p,m,L_1}n^{-\frac{2}{3m} + \frac{2\epsilon}{3m}}.
\end{align*}
Finally, as for $n$ large $\psi_2^n$ is simple, $P_2(Pf) = \|P_2(Pf)\|_{\textbf{m}}\psi_2^n$ and so
\begin{align*}
    \left\|Pf - \psi_2^n\right\|_{\textbf{m}} &\leq \left\|Pf - P_2(Pf)\right\|_{\textbf{m}} + \left\|P_2(Pf) - \psi_2^n\right\|_{\textbf{m}} \\
    & \leq \left\|Pf - P_2(Pf)\right\|_{\textbf{m}} + \left|1 - \|P_2(Pf)\|_{\textbf{m}}\right| \\
    & \leq \left\|Pf - P_2(Pf)\right\|_{\textbf{m}} + \left|1 - \|Pf\|_{\textbf{m}} + \|Pf - P_2(Pf)\|_{\textbf{m}}\right| \\
    & \leq 2\left\|Pf - P_2(Pf)\right\|_{\textbf{m}} + \left|1 - \|Pf\|_{\textbf{m}}\right| \\
    & \leq  \frac{1}{\lambda_3 - \lambda_2} \left[C_{\beta,p,m,L_1}n^{-\frac{1}{3m} + \frac{\epsilon}{m}} +  C_{\beta,p,m,L_1,L_2,K,\reach}n^{-\frac{2}{3m} + \frac{2\epsilon}{m}} \right]\lambda_2 \\
    & + C_{\beta,p,m,L_1}n^{-\frac{1}{3m} + \frac{\epsilon}{m}} + C_{m,\beta,p,L_1,L_2,K}n^{-\frac{2}{3m} + \frac{2\epsilon}{m}}
\end{align*}
\end{proof}

\nc
\begin{rems}
\label{rem:ExtensionsEigenvectors}
It is possible to consider alternative metrics to capture the convergence of graph Laplacian eigenvectors toward their continuum counterparts; e.g., see \cite{Trillos2019_Error}. We also want to remark that extending the convergence result from Theorem \ref{thm:eigenvector} to higher eigenfunctions is straightforward but slightly cumbersome as one needs to introduce additional notation and operators. The rationale of the proof is however the same and relies on the strong convexity of the discrete Raleigh quotient when restricted to eigenspaces. In this interpretation, the strong convexity constants are determined by the spectral gaps between consecutive eigenvalues. We refer the reader to \cite{Trillos2019_Error} for some related discussion. 
\end{rems}

\begin{rems}

It is interesting to observe that the proof of convergence of eigenvectors presented in \cite{calder2022improved}, which ultimately produces faster rates of convergence than the ones based on energy considerations (i.e., the approach used in \cite{Trillos2019_Error,burago2015graph} as well as in here) cannot be used in our setting, since the approach considered in  \cite{calder2022improved} relies on \textit{pointwise consistency} of graph Laplacians. Pointwise consistency results  for Fermat-based graph Laplacians do not follow from similar considerations as in other works in the literature given that the discrete Fermat distances are random. Obtaining pointwise consistency results for Fermat-based graph Laplacians is thus left as an interesting open problem.
\end{rems}

\subsection{Extensions and Discussion}
\label{rem:NormalizedGraphs}


It is possible to adapt the proofs of our theorems in Sections \ref{sec: Discrete Dirichlet energies}- \ref{sec:EigenvecConvergenceMain} to obtain similar spectral convergence results (with rates) for more general normalizations of graph Laplacians such as the ones discussed in Section \ref{sec:OtherGraphLaplacians}. Let the base weights $\tilde{W_{ik}}$ be defined according to $\tilde{W}_{ik} = w_{i,k}^{\tilde{\ell},h} $ as in \eqref{eq:WeightsRGG} for $d_0=\tilde{\ell}_p^p$ the discrete $p$-Fermat distance. We choose $r=0, j=q=s$ in the construction from Section \ref{sec:OtherGraphLaplacians} and denote by $L_{p,s}$ the resulting normalized graph Laplacian. It is straightforward to see that $L_{p,s}$ is self-adjoint with respect to the inner product $\langle \cdot, \cdot \rangle_D $, where the matrix $D$ is the degree matrix for the weights $W_{i,k}= \frac{\tilde{W}_{ij}}{\tilde D ^s_i \tilde D _k^s }$.

The (properly scaled) Dirichlet energy associated to the operator $L_{p,s}$  takes the form
\[ \frac{2}{n(m+2)h^{m+2}} \sum_{i,k} \frac{\eta ( \frac{\tilde{\ell}_p^p(x_i, x_k)}{h} )}{\tilde D _i^s \tilde D _k^s} \left(u(x_i) - u(x_k)\right)^2.  \]
Note that the only difference with the Dirichlet energy $b_h^{d_0}(u,u)$ from \eqref{eq:GraphDirichlet} is the extra normalization terms $\tilde D_i^s $, where
\[\tilde D_i:= \frac{1}{n}\sum_{k=1}^n \eta \left( \frac{\tilde{\ell}_p^p(x_i, x_k)}{h} \right)  .\]
Using Lemma \eqref{lem:Degree_Approx}, we can reduce our analysis to studying the Dirichlet energy  
\[ \frac{2}{n(m+2)h^{m+2}} \sum_{i,k} \frac{\eta ( \frac{\tilde{\ell}_p^p(x_i, x_k)}{h} )}{\rho^{sp}(x_i) \rho^{sp}(x_k)} \left(u(x_i) - u(x_k)\right)^2,  \]
which, in turn, can be related to the continuum Dirichlet energy 
\[ \int_{\M} |\nabla_p u|^2 \rho_p^{2s}(x) \V_p(x),   \]
after following very similar steps as in the proofs of our results in Sections \ref{sec: Discrete Dirichlet energies} and \ref{sec:SpectralConvergenceDiscreteToContinuum}. 

The operators $L_{p,s}$ described above can thus be shown to converge spectrally toward the family of $s$-weighted operators (see Theorem \ref{thm:PM_lap})):
\[  \Delta_{s,p} u = - \frac{1}{\rho_p^{2s}} \divv_p\left( \rho^{2s}_p\nabla_p u \right).  \]
Compare with the discussion in Remark \ref{rem:ContinuumLimitsHoffmann}. See also our discussion in Section \ref{sec:NumericsAndNormalizations}.



\subsubsection{Dynamic Perspective and Comparison with Diffusion Maps}
\label{sec:DynamicPerspective}

As we discuss below, every graph Laplacian operator
$L_{p,s}$ induces a family of transformations of the original data set $\{ x_1, \dots, x_n\}$ (which we recall is embedded in $\R^m$) into $\R^n$. The idea of using these transformations is to help ``disentangle" the original data set $\X$ by finding a representation for it that more clearly reveals its intrinsic geometric structure. These transformations are in many instances effective preprocessing steps for tasks such as clustering or dimensionality reduction. 

To introduce the family of transformations associated to $L_{p,s}$, let us denote by $Q_{s,p}:= -L_{p,s}$ the transition rate matrix associated to the Laplacian $L_{p,s}$. Following Section 2 in \cite{Craig2022}, we consider the family of evolution equations:
 \begin{align} \label{eqn:Markovchain}
     \begin{cases}
     \partial_t {u}_t(y) = \sum_{x \in \X } u_t(x) Q_{s,p}(x,y) , \quad t >0 \\
     u_0 = u^0, 
     \end{cases}
 \end{align}
 where each $u_t$ is interpreted as a function $u_t :\X \rightarrow \R$; in the above, we use the notation $Q_{s,p}(x_i,x_j)$ to denote the $ij$ entry of $Q_{s,p}$.  Fixing a time horizon $T>0$, for an arbitrary data point $x_i$ we set $u^0:= \delta_{x_i}$ and define the transformation
 \begin{equation}
    x_i \mapsto u_T^{x_i} \in L^2(\{ x_1, \dots, x_n\}) \cong \R^n,
    \label{eq:TransformationPoints}
 \end{equation}
where we have used the notation $u^{x_i}$ to denote the solution to equation \eqref{eqn:Markovchain} when its initial condition is $\delta_{x_i}$, the function over $\X$ which is one at $x_i$ and $0$ everywhere else.

In what follows we discuss the similarities and differences between the family of transformations induced by the operators $\{ L_{p,s} \}_{s}$ for different values of $p$ and the family of diffusion maps from \cite{Coifman2006diffusion}. To compare these families, we use the connection between discrete and continuum operators that we developed in the previous sections and draw conclusions from the comparison between the continuum analogues of each of these families.
 
 First, as in Section 2 in \cite{Craig2022}, we can argue that the (discrete) evolution equation \eqref{eqn:Markovchain} has the following continuum counterpart:
\begin{align*} 
 \partial_t f_{t,p} =  \divv_p \left(\rho_p^{2s}  \nabla_p \left(  \frac{\ f_{t,p}}{\rho_p^{2s}}\right)\right),
 \end{align*}
 where the function $f_{t,p}$ must be interpreted as a density function with respect to the volume form $\V_p$. In terms of density functions and operators in the base geometry $(\M, g)$, the above equation can be rewritten as: 
 \begin{equation}
     \partial_t f_{t}=  \divv(\rho^\alpha \nabla f_t) - ((2s-1)p +1) \divv( f_t \rho^{\alpha-1} \nabla \rho ) 
    \label{eq:FermatDistance_Evolution}
 \end{equation}
 where the function $f_{t}= f_{t,p}\rho^{1-p} $ is interpreted as a density function with respect to the base volume form $\V$. If $\alpha >0$, the above can be 
 written as
 \[ \partial_t f_{t}=  \divv(\rho^\alpha \nabla f_t) - \frac{((2s-1)p +1)}{\alpha} \divv( f_t \nabla \rho^\alpha ),  \]
 whereas for $\alpha =0$ (which happens when $p=1$) it can be written as
 \[ \partial_t f_{t}=  \divv( \nabla f_t) - 2s \divv( f_t \nabla \log(\rho) ). \]
 Notice that regardless of the value of $\alpha \geq 0$ chosen, the second term on the right hand side of \eqref{eq:FermatDistance_Evolution} is a ``mean shift term", {\em i.e.,} it is a term that pushes the distribution $f_t$ towards the local maxima of the function $\rho$ (which are the same as the local maxima of $\log(\rho)$ or $\rho^\alpha$ for $\alpha>0$). The first term on \eqref{eq:FermatDistance_Evolution}, on the other hand, is a non-homogeneous diffusion term with a diffusion coefficient that is larger at points with larger values of $\rho$. The larger the value of $\alpha$, the more dramatic the difference between the diffusion rate at points with large values of $\rho$ and at points with small values of $\rho$. Notice that $\alpha$ gets modulated by the value of $p$ only, and in particular $s$ does not play any role in determining it. As we will see below, being able to tune this effect in the diffusion term in \eqref{eq:FermatDistance_Evolution} is one of the main differences between the family of maps represented by \eqref{eq:FermatDistance_Evolution} and the diffusion maps of \cite{Coifman2006diffusion}. 
 
 Let us recall that, as discussed in Section 4.2 in \cite{Craig2022}, diffusion maps are connected to the following family of evolution equations:
 \begin{align} \label{CL_varyingalpha}
 \partial_t f_t =  \divv_\M \left(\rho^{2(1-a)}  \nabla_\M \left(  \frac{\ f_t}{\rho^{2(1-a)}}\right)\right)=   \Delta_\M f_t - 2(1-a) \divv_\M ( f_t \nabla_\M \log(\rho)),
 \end{align}
 for $a \in (-\infty, 1]$. We see that \eqref{eq:FermatDistance_Evolution} and \eqref{CL_varyingalpha} coincide when $s=1-a$ and $p=1$. This is expected, since the distance function $\mathcal{L}_p$ is precisely the original metric on $\M$ when $p=1$. In general, however, with the family of Fermat based Laplacians we have one extra degree of freedom that can be used to accelerate the diffusion in regions where the density $\rho$ is larger, as was mentioned earlier. Numerical examples illustrating this effect are presented in Section \ref{sec:NumericsAndNormalizations}. In settings like the blue sky problem from Figure \ref{fig:landscape}, considering larger values of $\alpha$ in the maps \eqref{eq:FermatDistance_Evolution} should be beneficial, as in that case the map \eqref{eq:TransformationPoints} is expected to induce agglomeration of points at each of the clusters more quickly while still inducing the mean-shift force pushing clusters apart. However, we highlight that in order to translate our insights from the continuum model to the discrete setting, it is important to have enough samples to justify the approximation of the continuum model by the discrete one, and in this regard, larger values of $\alpha$ require a larger number of samples.


\nc

\section{Numerical Simulations}
\label{sec:NumericsAndNormalizations}

The discrete Fermat distance spectral clustering algorithm is summarized in Algorithm \ref{alg:FD_SC}. The algorithm depends on two key parameters: $p$, which determines the metric geometry, and $s$, which determines the Laplacian normalization. In Section \ref{subsec:Role_ps} we explore the impact of these parameters.  In Section \ref{subsec:NormalizationsVsFermat} we explore the equivalence of Algorithm \ref{alg:FD_SC}, where Fermat distances are computed explicitly, with Algorithm \ref{alg:DegNorm_SC}, a Euclidean spectral clustering algorithm with positive degree normalization. 


\begin{algorithm}[t!]
	\caption{Fermat Distance Spectral Clustering}\label{alg:FD_SC}
	\begin{algorithmic}[1]
		\State \textbf{Input:} data points $x_1, \ldots, x_n$, density parameter $p\geq 1$, normalization parameter $s\geq0$, embedding dimension $r$, kernel scale $\epsilon$
		\State \textbf{Output:} Laplacian $L_{\epsilon,s}(\ell_p^p)$, FD-SC spectral embedding $[v_1, \ldots, v_r]\in \mathbb{R}^{n\times r}$ \\

		\State $W_p(x_i,x_k) \gets \eta(\ell_{p}(x_i,x_k)/\epsilon)$		\Comment Compute weights
		\State $d_p(x_i) \gets \sum_k W_p(x_i,x_k)$ \Comment Compute normalization factor
		\State $W_{p, s}(x_i,x_k) \gets \displaystyle \frac{W_p(x_i,x_k)}{d_p(x_i)^{(1-\frac{s}{2})}d_p(x_k)^{(1-\frac{s}{2})}}$ \Comment Compute normalized weight matrix\\
		\State $D_{p, s}(x_i,x_i) \gets \sum_k W_{p, s}(x_i,x_k)$ \Comment Compute degree matrix
		\State $L_{p,s} \gets D_{p, s}^{-1}\left(D_{p, s} - W_{p, s}\right)$ \Comment Compute random walk Laplacian
            \State $[v_1, \ldots, v_r]\gets$ bottom $r$ eigenvectors of $L_{p,s}$ 
	\end{algorithmic}
\end{algorithm}

\begin{algorithm}[t!]
	\caption{Degree Normalized Spectral Clustering}\label{alg:DegNorm_SC}
	\begin{algorithmic}[1]
		\State \textbf{Input:} data points $x_1, \ldots, x_n$, parameters $q,j$, embedding dimension $r$, kernel scale $\epsilon$
		\State \textbf{Output:} Laplacian $L_{q,j}$, DN-SC embedding $[v_1, \ldots, v_r]\in \mathbb{R}^{n\times r}$ \\
		\State $W(x_i,x_k) \gets \eta(\| x_i -x_k\|/\epsilon)$ \Comment Compute weights 
		\State $d(x_i) \gets \sum_k W(x_i,x_k)$  \Comment Compute normalization factor
		\State $W_{q}(x_i,x_k) \gets  \displaystyle\frac{W(x_i,x_k)}{d(x_i)^{(1-\frac{q}{2})}d(x_k)^{(1-\frac{q}{2})}}$ \Comment Compute normalized weight matrix\\
		\State $D_{q}(x_i,x_i) \gets \sum_k W_{q}(x_i,x_k)$ \Comment Compute degree matrix
		\State $L_{q,j} \gets D_{q}^{\frac{1-j}{q-1}}\left(D_{q} - W_{q}\right)$ \Comment Compute normalized Laplacian
            \State $[v_1, \ldots, v_r]\gets$ bottom $r$ eigenvectors of $L_{q,j}$
	\end{algorithmic}
\end{algorithm}

\subsection{Role of $p$ and $s$ in Fermat Spectral Clustering}
\label{subsec:Role_ps}

\begin{figure}[h]
	\centering
	\begin{subfigure}[t]{0.48\textwidth}
		\centering
		\includegraphics[width=\textwidth]{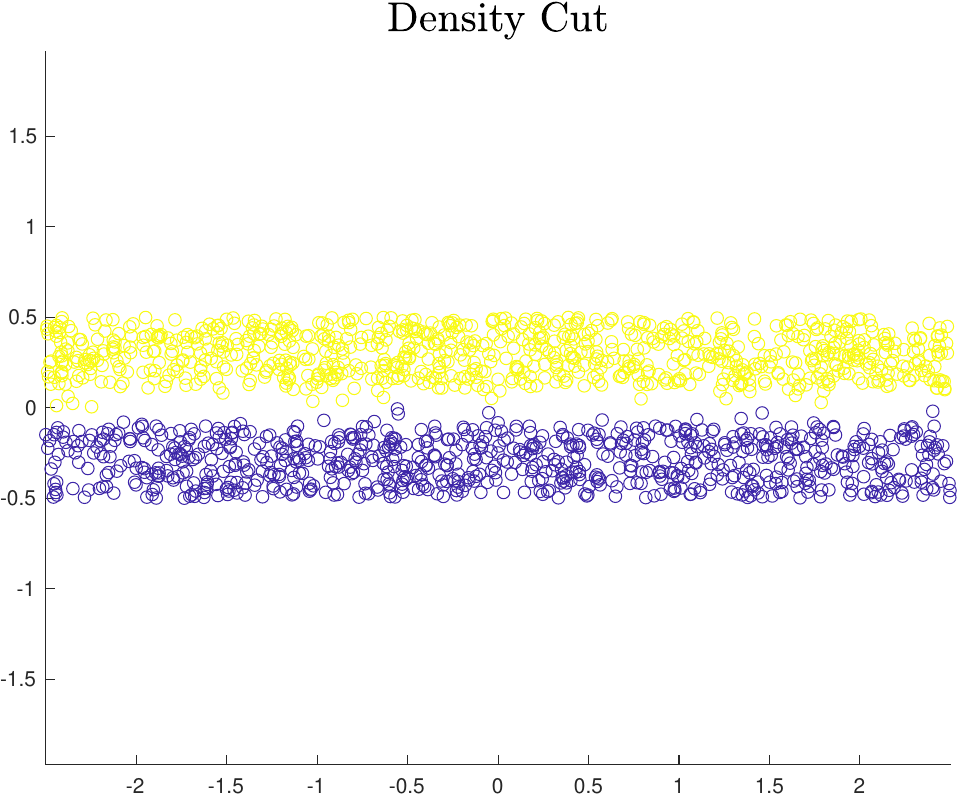}
	\end{subfigure}
	\begin{subfigure}[t]{0.48\textwidth}
		\includegraphics[width=\textwidth]{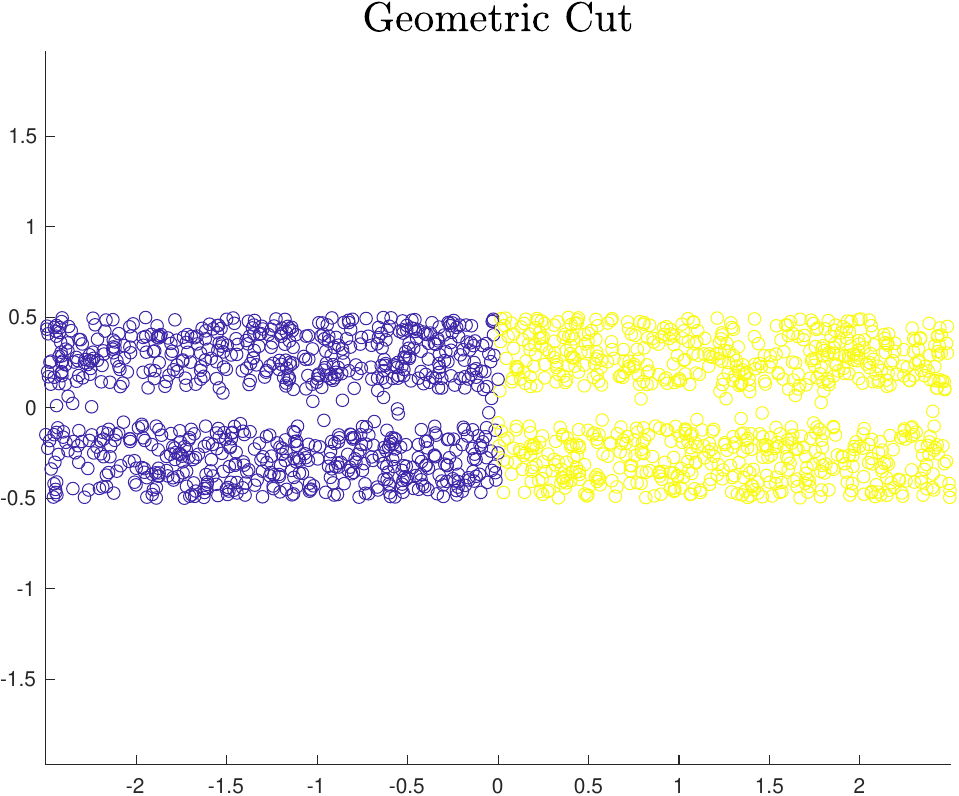}
	\end{subfigure}
 	\begin{subfigure}[t]{0.48\textwidth}
		\includegraphics[width=\textwidth]{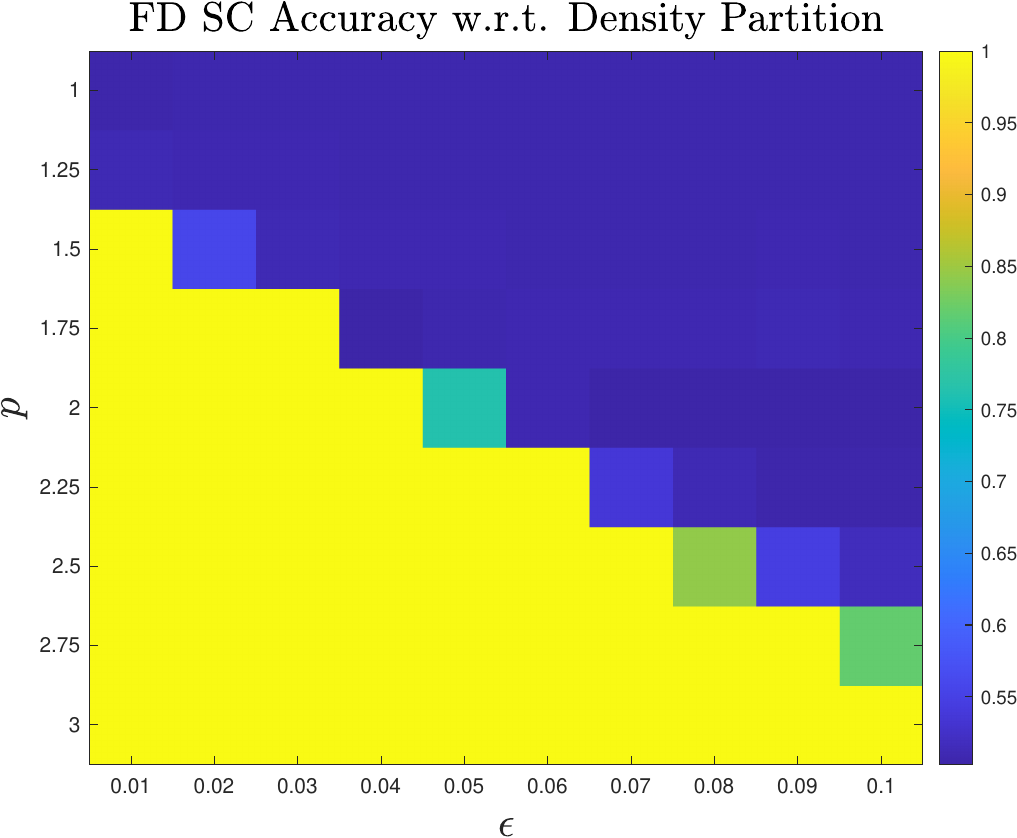}
	\end{subfigure}
	\begin{subfigure}[t]{0.48\textwidth}
		\centering
		\includegraphics[width=\textwidth]{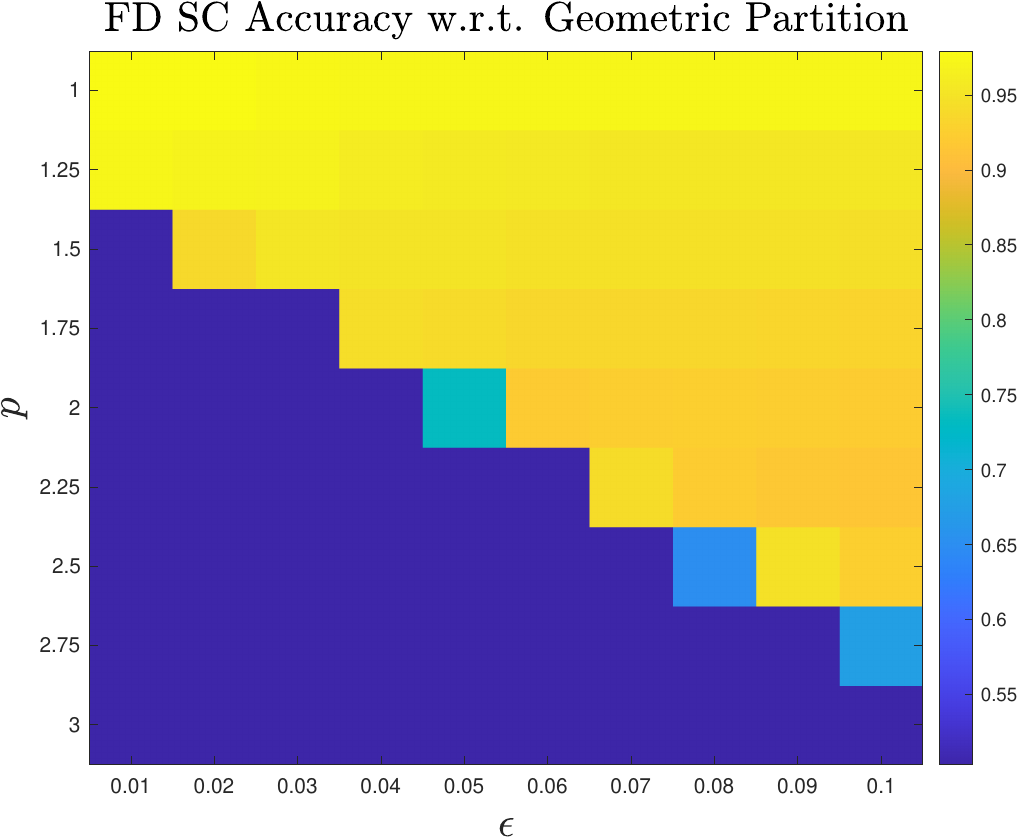}
	\end{subfigure}
	\caption{Fermat distance SC on elongated, density separated data.  Two partitions are reasonable.  As $p$ increases, the learned clustering transitions from the geometric partition to the density partition.}
	\label{fig:TwoPartitions}
\end{figure}

In Figure \ref{fig:TwoPartitions}, we consider an elongated data set with a density gap.  There are two natural partitions: one that cuts ``long" through the region of low density (which we call ``Density Cut") and one that cuts ``short" (which we call ``Geometric Cut").  We see that as $p$ increases, the density cut is eventually learned by Algorithm \ref{alg:FD_SC} by applying $k$-means to the spectral embedding to produce labels, which are then compared with the two natural partitions via an \emph{accuracy} score that computes the proportion of points correctly assigned.  The behavior in $p$ of Figure \ref{fig:TwoPartitions} corroborates our analysis that low $p$ will favor geometry-driven cuts and large $p$ density-driven cuts.

\begin{figure}[h]
	\centering
\begin{subfigure}[t]{0.32\textwidth}
		\centering
		\includegraphics[width=\textwidth]{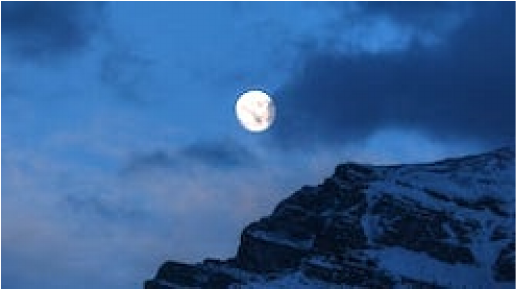}
            \caption{Original Image}
            \label{fig:moon_original_image}
\end{subfigure}
\begin{subfigure}[t]{0.32\textwidth}
		\centering
		\includegraphics[width=\textwidth]{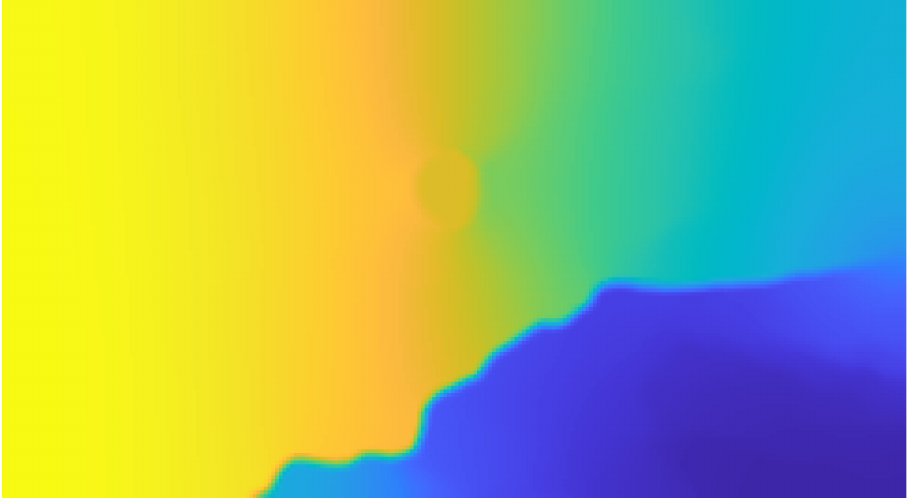}
            \caption{Euclidean RW $v_2$ ($s=2$)}
            \label{fig:moon_reg_v2}
\end{subfigure}
\begin{subfigure}[t]{0.32\textwidth}
		\centering
		\includegraphics[width=\textwidth]{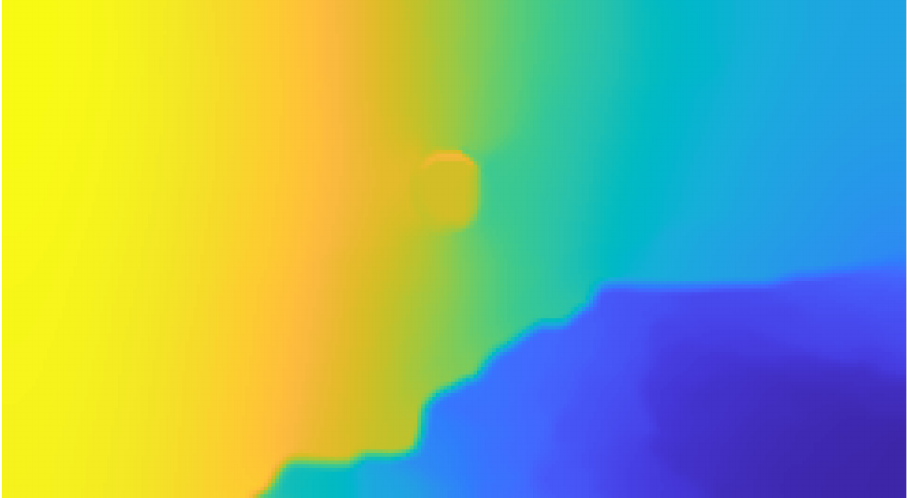}
            \caption{Fermat $v_2$ ($p=2$, $s=1$)}
            \label{fig:moon_Fermatp2s1_v2}
\end{subfigure}
\begin{subfigure}[t]{0.32\textwidth}
		\centering
		\includegraphics[width=\textwidth]{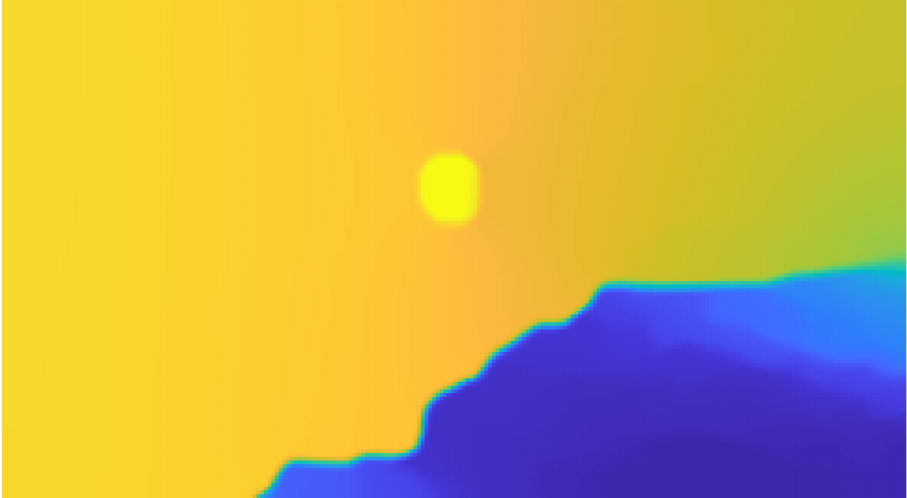}
            \caption{Fermat $v_2$ ($p=2$, $s=2$)}
            \label{fig:moon_Fermatp2s2_v2}
\end{subfigure}
\begin{subfigure}[t]{0.32\textwidth}
		\centering
		\includegraphics[width=\textwidth]{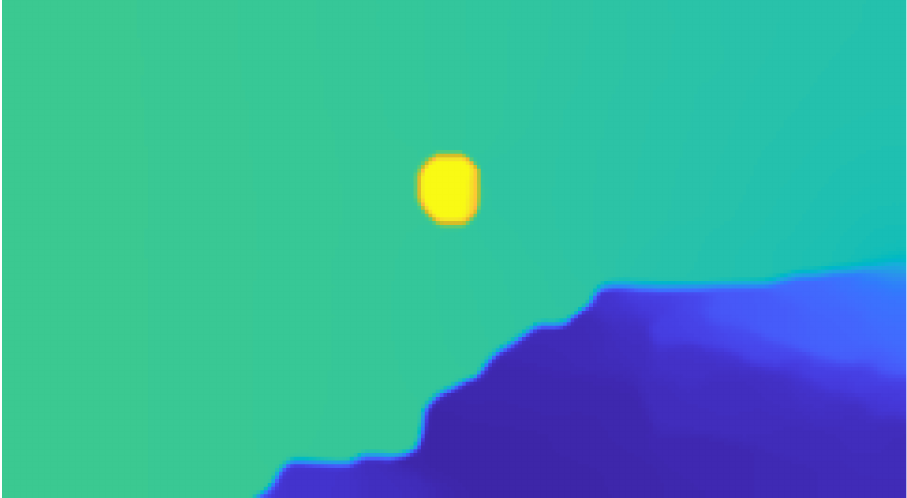}
            \caption{Fermat $v_2$ ($p=2$, $s=2.5$)}
            \label{fig:moon_Fermatp2s25_v2}
\end{subfigure}
\begin{subfigure}[t]{0.32\textwidth}
		\centering
		\includegraphics[width=\textwidth]{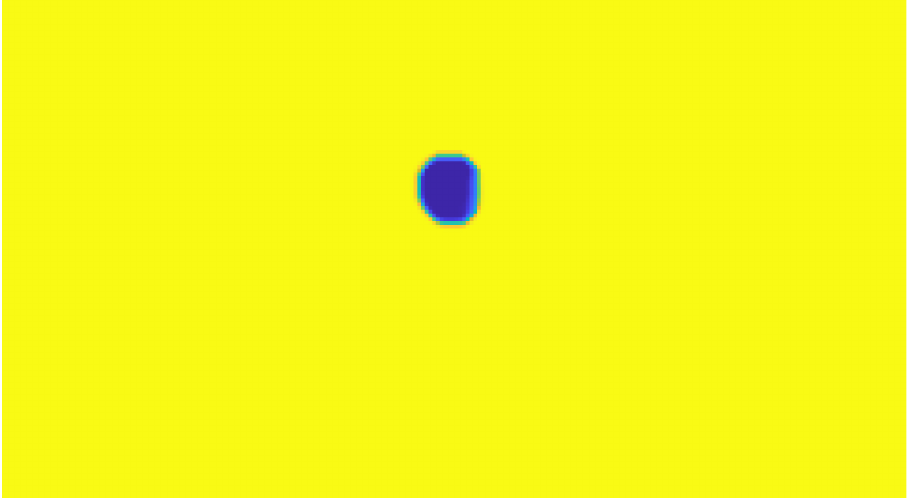}
            \caption{Fermat $v_2$ ($p=2$, $s=3$)}
            \label{fig:moon_Fermatp2s3_v2}
\end{subfigure}
	\caption{Moon example: effect of normalization. For small $s$, the Fermat $v_2$ cuts along the horizon, but for large $s$ the cut produced by Fermat $v_2$ may yield very imbalanced clusters.}
	\label{fig:moon}
\end{figure}

 In Figure \ref{fig:moon}, we demonstrate the impact of the normalization parameter $s$ while keeping $p$ fixed.  As $s$ increases, the moon---which is small but very separated from the rest of the pixels by intensity---is increasingly emphasized in the second Fermat eigenvector. Choosing a large $s$ allows the Fiedler eigenvector to concentrate on a set of very small volume, while a smaller $s$ prevents this behavior and recovers the more balanced cut of the horizon.  
 
\subsection{Asymptotic Equivalence with Degree Normalized Euclidean Laplacians}
\label{subsec:NormalizationsVsFermat}

We compare to \cite{Hoffmann2019_Spectral} by sampling data from a non-uniform density and constructing (i) a discrete Fermat graph Laplacian (e.g., Algorithm \ref{alg:FD_SC}) and (ii) a Euclidean distances graph Laplacian, with density rescaling so that in the large sample limit, the discrete Laplacians convergence to the same continuum operator (e.g., Algorithm \ref{alg:DegNorm_SC}).  As a baseline, we also compare with (iii) the random walk Laplacian build with Euclidean distances but without density rescaling.  

We consider 2 example datasets in $\mathbb{R}^{2}$:
\begin{enumerate}
    \item Data sampled from a ball of radius 1 with a density valley running down the vertical axis, specifically we consider for some $\tau>0$ the density $\rho(x_{1},x_{2})\propto(\tau+x_{1}^{2})^{-1}\cdot\mathbb{1}_{[0,1]}(x_{1}^{2}+x_{2}^{2})$.
    \item A mixture of 2 Gaussians with uniform background noise, whose intensity is governed by a parameter $\tau>0$.
\end{enumerate}  

\begin{figure}[t]
	\centering
\begin{subfigure}[t]{0.32\textwidth}
		\centering
		\includegraphics[width=\textwidth]{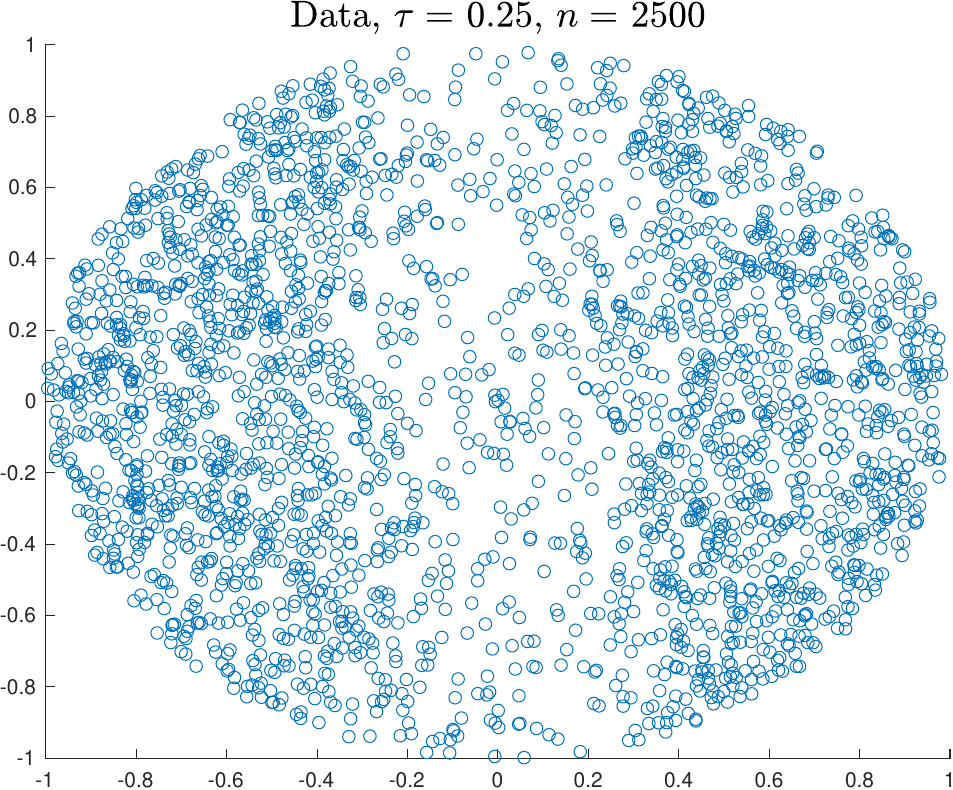}
\end{subfigure}
\begin{subfigure}[t]{0.32\textwidth}
		\centering
		\includegraphics[width=\textwidth]{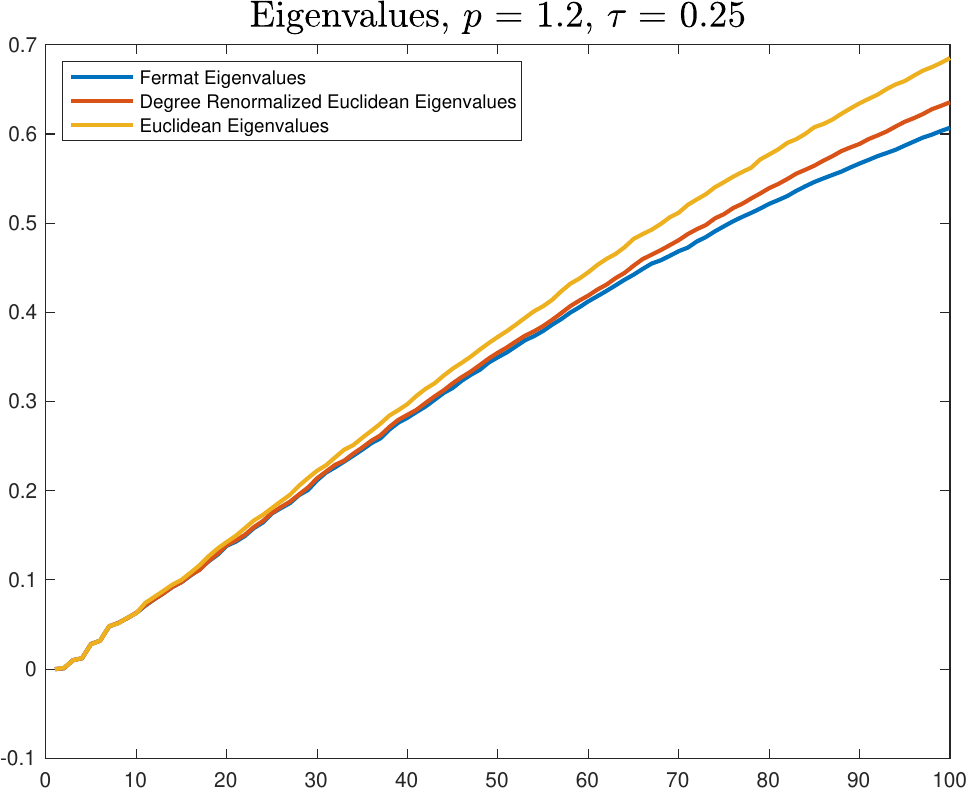}
\end{subfigure}
\begin{subfigure}[t]{0.32\textwidth}
		\centering
		\includegraphics[width=\textwidth]{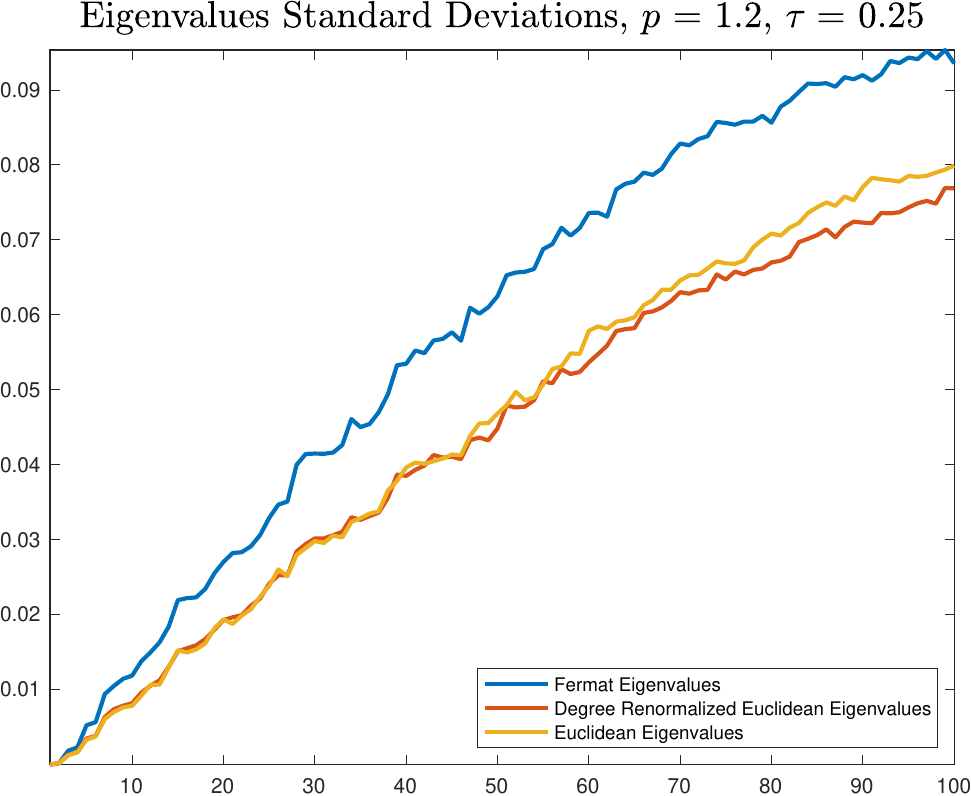}
\end{subfigure}

     \begin{subfigure}[t]{0.32\textwidth}
		\centering
		\includegraphics[width=\textwidth]{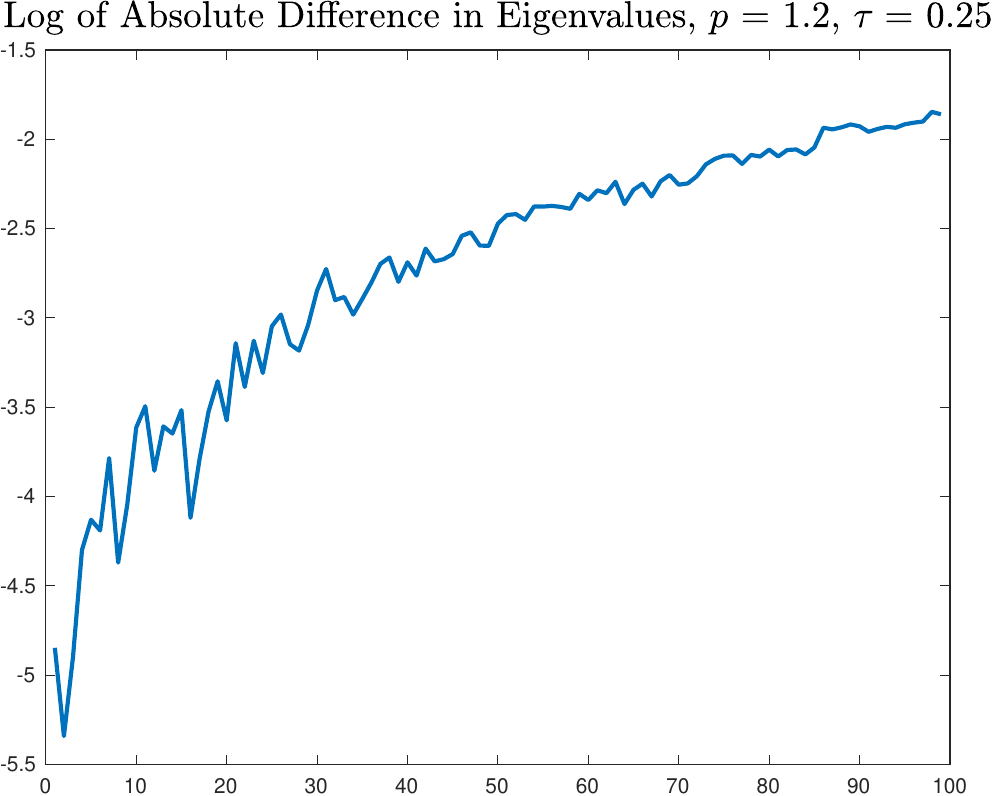}
\end{subfigure}
  \begin{subfigure}[t]{0.32\textwidth}
		\centering
		\includegraphics[width=\textwidth]{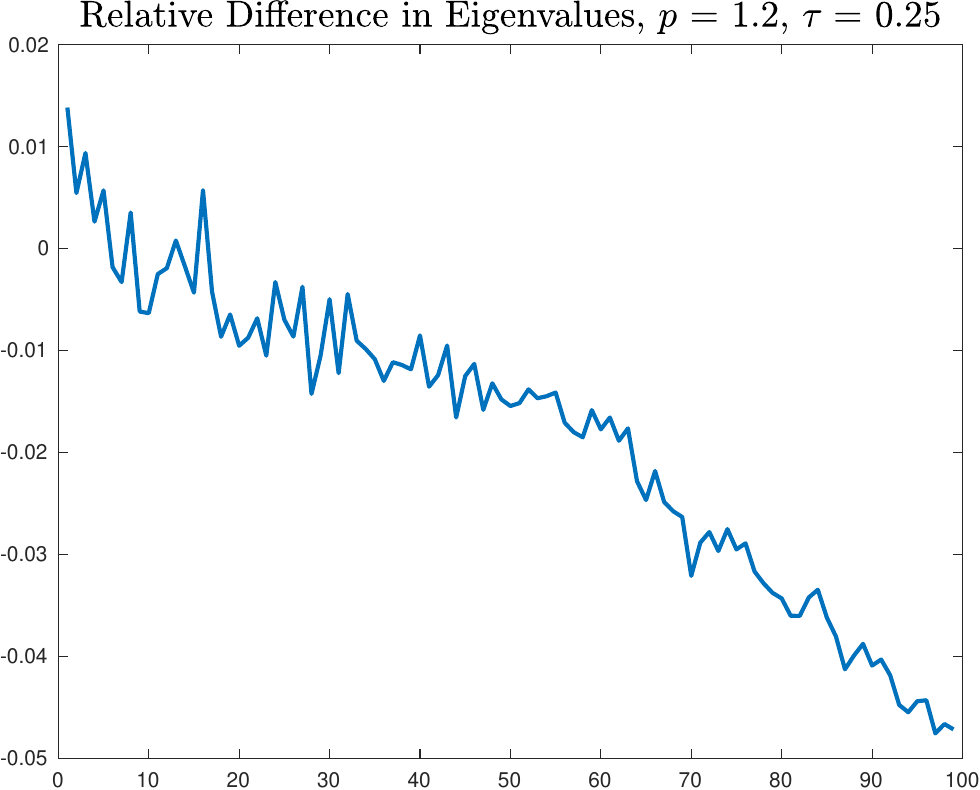}
\end{subfigure}
\begin{subfigure}[t]{0.32\textwidth}
		\centering
		\includegraphics[width=\textwidth]{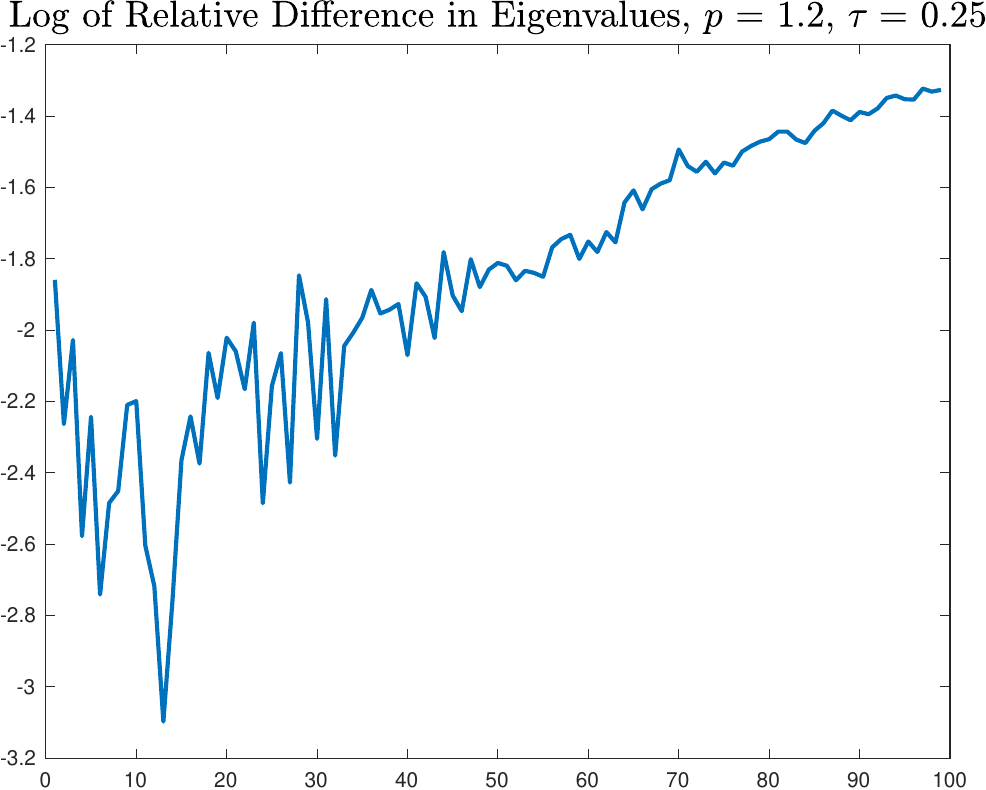}
\end{subfigure}
	\caption{$p=1.2$, $\tau=.25$.  Runtime for Fermat Laplacian: $168.46\pm 10.70$s.  Runtime for Rescaled Euclidean Laplacian: $6.64\pm .63$s.}
	\label{fig:BallGap_p=1.2_thresh=.25}
\end{figure}

\begin{figure}[t]
	\centering
\begin{subfigure}[t]{0.32\textwidth}
		\centering
		\includegraphics[width=\textwidth]{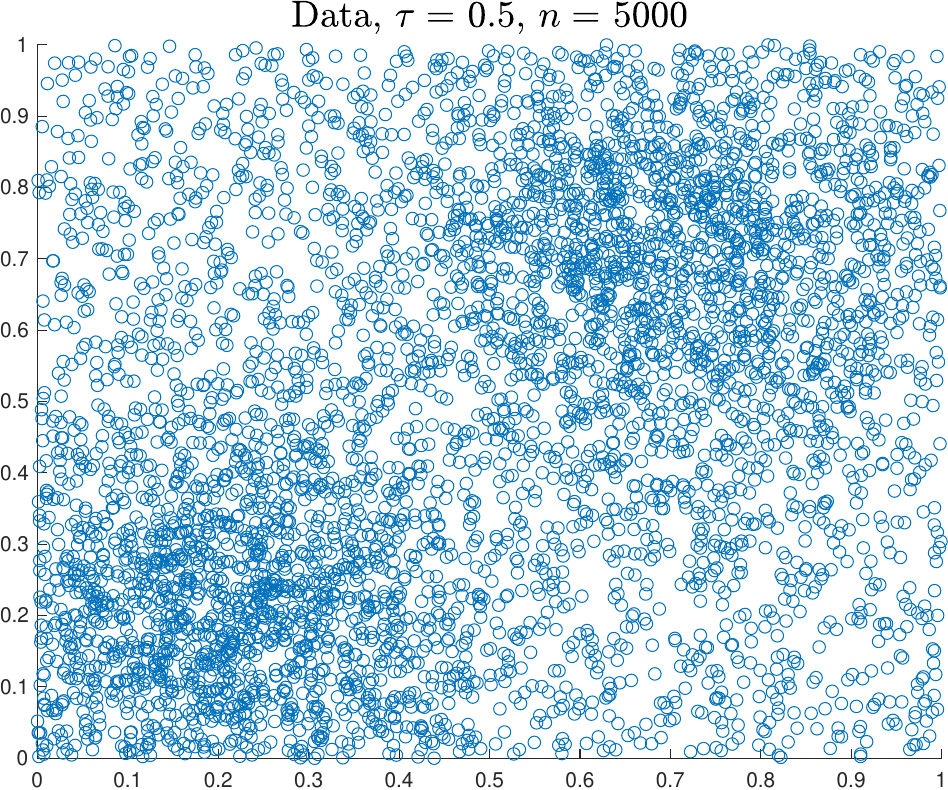}
\end{subfigure}
\begin{subfigure}[t]{0.32\textwidth}
		\centering
		\includegraphics[width=\textwidth]{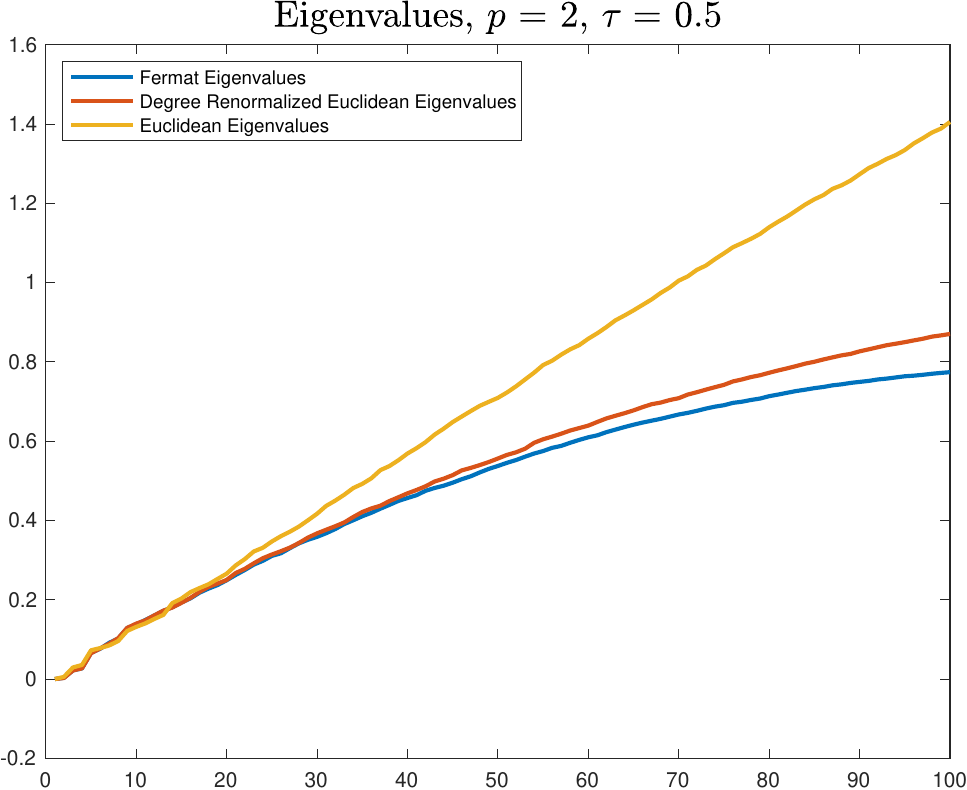}
\end{subfigure}
\begin{subfigure}[t]{0.32\textwidth}
		\centering
		\includegraphics[width=\textwidth]{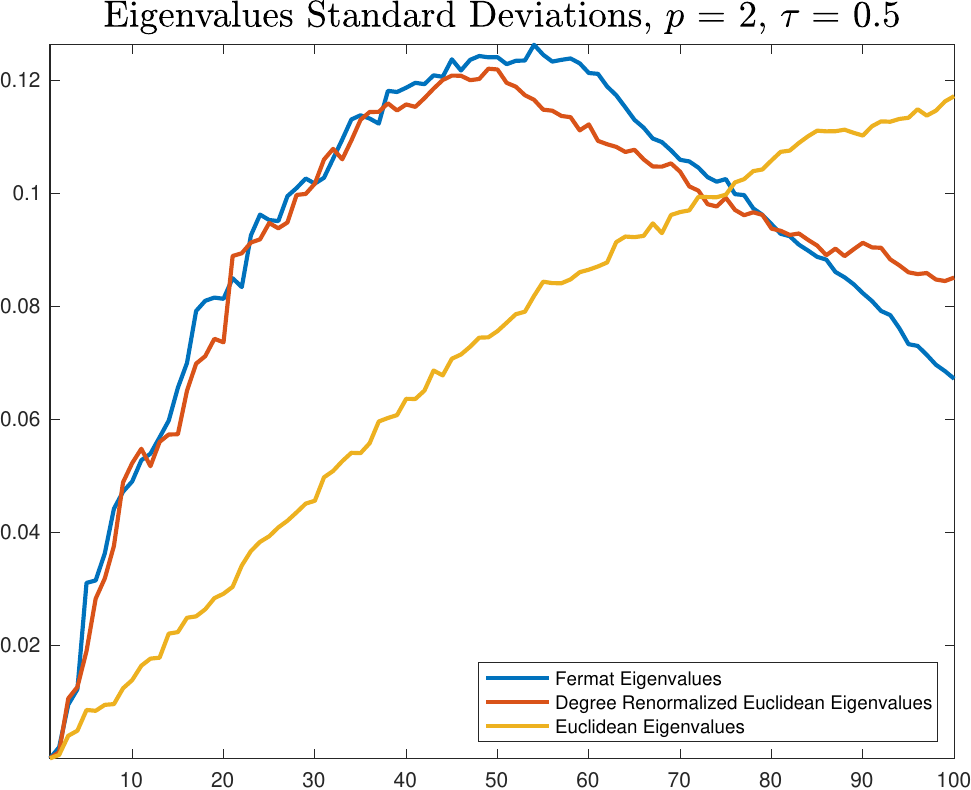}
\end{subfigure}
\begin{subfigure}[t]{0.32\textwidth}
		\centering
		\includegraphics[width=\textwidth]{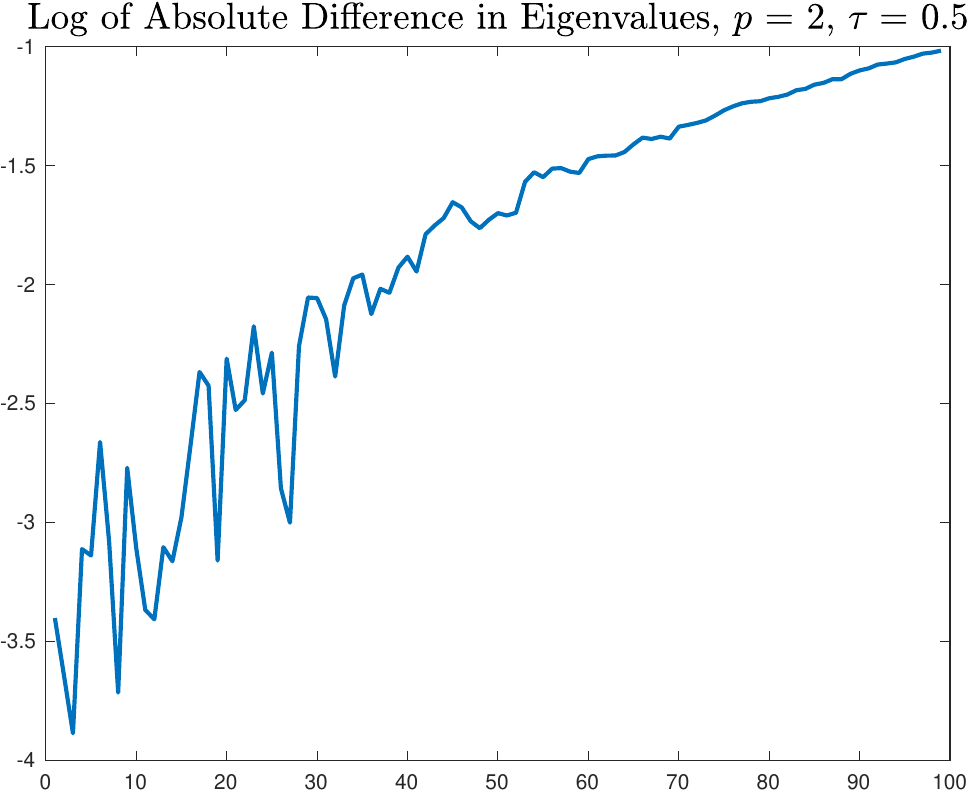}
\end{subfigure}
  \begin{subfigure}[t]{0.32\textwidth}
		\centering
		\includegraphics[width=\textwidth]{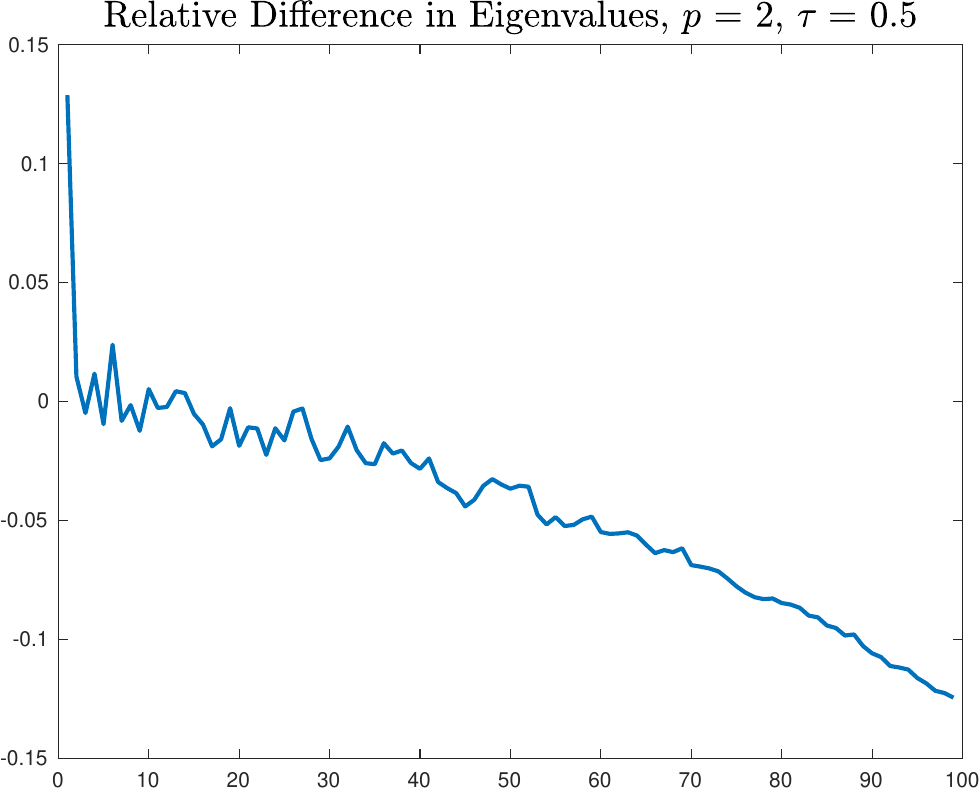}
\end{subfigure}
\begin{subfigure}[t]{0.32\textwidth}
		\centering
		\includegraphics[width=\textwidth]{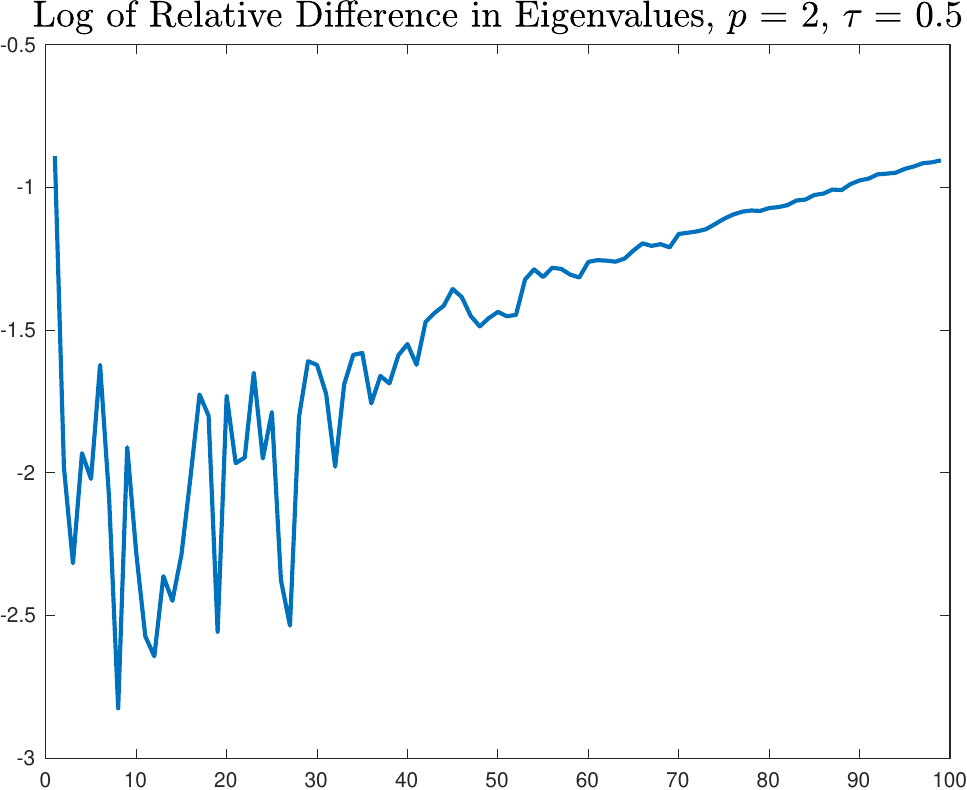}
\end{subfigure}
	\caption{$p=2$, $\tau=.5$.  Runtime for Fermat Laplacian: $271.85\pm96.75$s.  Runtime for Rescaled Euclidean Laplacian: $1.45\pm .23$s.}
	\label{fig:GaussianMixture_p=2_thresh=.5}
\end{figure}

Figures \ref{fig:BallGap_p=1.2_thresh=.25}-\ref{fig:GaussianMixture_p=2_thresh=.5} illustrate the convergence of the Fermat Laplacian eigenvalues to their density normalized counterparts.  The bottom row of these plots show comparisons between (i) and (ii).  In general, smaller $p$ and larger $\tau$ lead to larger regions of convergence.  It is also relevant that the Fermat eigenvalues have larger variance, especially the higher frequency ones.  Additional experiments appear in Appendix \ref{sec:Additional Plots}.

This suggests that in practice and for sufficiently large sample size $n$, density-normalization provides a more computationally efficient (because Fermat distances do not need to be calculated) and statistically consistent alternative to constructing the Fermat distance graph Laplacian.  Indeed, the complexity of computing a Fermat distances graph Laplacian, with each point connected to its $k$ Fermat nearest neighbors, is $O((k^{2}+CD)n\log(n))$ for a constant $C$ that depends exponentially on the intrinsic dimension of the data \citep{mckenzie2019power}.  This is done via a modified Dijkstra's algorithm that leverages fast Euclidean nearest neighbor algorithms such as cover trees \citep{beygelzimer2006cover}.  In contrast, computing a $k$ Euclidean nearest neighbors graph is essentially $O((k+CD)n\log(n))$.  In practice, the cost of computing the Fermat nearest neighbor distances is significantly larger than computing the Euclidean nearest neighbors, even if they have the same asymptotic complexity in $n$ ({\em i.e.,} $k^2$ is fixed to be relatively small compared to $C$ and $D$). 
 The spectral decomposition of the sparse Fermat distance Laplacian and density-reweighted Euclidean Laplacian have essentially the same complexity (since the sparsity level is the same), so the benefit of avoiding Fermat Laplacians is at the level of avoiding expensive distance calculations that suffer from relatively high statistical variance \citep{little2022balancing}.


\section{Conclusions and Future Work}

By computing Fermat geodesics and applying percolation results in the manifold tangent plane, we have developed the first quantitative local Fermat metric convergence results in the literature. We apply these results to develop a continuum limit theory for Fermat graph Laplacians, and prove convergence results of the eigenvalues and eigenvectors of the discrete operators to those of their continuum analogues.  An interesting consequence of this analysis is that it establishes the similarity of (i) Fermat Laplacians and (ii) density-reweighted Euclidean Laplacians in the large sample limit.  The geometric framework we develop thus leads to new computational schemes that leverage the theoretical benefits of Fermat spectral clustering (for example robustness with respect to cluster elongation as illustrated in Figure \ref{fig:landscape}) without the need for onerous calculation of pairwise Fermat distances.  

In future work, it is of interest to investigate whether our local metric convergence results can be extended to apply globally, whether it is possible to obtain point-wise consistency for Fermat graph Laplacians, how our results are impacted by noise, and also to theoretically investigate the role of the normalization parameter $s$, which we have fixed in this work.  
\label{sec:conclusion}

\vspace{10pt}

\noindent\textbf{Acknowledgements:}  NGT was supported by NSF DMS-2005797.  AL thanks NSF DMS-2136198 and NSF DMS-2309570.  JMM acknowledges support from NSF DMS-1912737 and DMS-1924513. 

\clearpage
\newpage
\bibliography{FD.bib}

\appendix

\section{Preliminaries from Differential Geometry}
\label{app:prelim}

 In this appendix, we shall use the Einstein summation convention: when an index appears in an upper and lower position, it means summation over that index. In contrast to the main body, we shall use $\operatorname{grad}$ (resp. $\operatorname{Hess}$) to denote the Riemannian gradient (resp. Hessian) so that $\nabla$ (resp. $H$) can be reserved for Euclidean gradients (resp. Hessians) computed in the ambient space or on a tangent space.

\begin{defn}Let $\dist(x,\M)=\min_{y\in\M}\|x-y\|$.  The \emph{reach} of $\M\subset\mathbb{R}^D$ is
	\begin{equation*}
		\reach:= \sup\{ t > 0: \forall x \in \mathbb{R}^{D} \text{ with } \dist(x,\M) \leq t, \exists ! y \in \M \text{ s.t. } \dist(x,\M) = \|x-y\|\}.
	\end{equation*}
	
\end{defn}
As $\reach$ depends only on the embedding, not on the specific metric chosen on $\M$, the reach of $(\M, g_{p})$ equals the reach of $(\M, g)$.

\begin{defn}For any linearly independent $X,Y \in T_x\M$ the {\em sectional curvature}, with respect to $g$, of the plane spanned by $X,Y$ is defined as 
	
	\begin{equation*}
		K_x(X,Y) = \frac{\mathbf{R}_x(X,Y,Y,X)}{g_x(X,X)g_x(Y,Y) - g_x(X,Y)^2},
	\end{equation*}
	where $\mathbf{R}$ is the Riemannian curvature tensor.
	
\end{defn}
We are interested in $(\M,g)$ with globally bounded sectional curvature:
\begin{equation*}
	\left|K_x(X,Y)\right| \leq K \quad \text{ for all } x \in \M, \ X,Y\in T_{x}\M.
\end{equation*}

\begin{lemma}
    Suppose that $\rho \in C^2\M$. Fix any $x \in \M$ and consider the function
    \begin{align}
    \begin{split}
        & \rho^{*}_x: T_x\M \to \mathbb{R} \\
        & \rho^{*}_x(u) = \rho(\exp_x(u))J_x(u) \label{eq:Def_of_rho_star}
        \end{split}
    \end{align}
    where $J_x = \sqrt{\det(g_x)}$. Then $\|\nabla \rho^{*}_x(0)\| = \|\nabla\rho(x)\|$ and
    \begin{equation*}
        \|H\rho^{*}_x(0)\| \lesssim \|H\rho(x)\| + mK\rho(x) + Kd^{\frac{3}{2}} \| \nabla \rho(x)\|.
    \end{equation*}
\label{lem:Bound_Hess_expansion}
\end{lemma}
\begin{proof}
    Starting from the normal coordinate expansion of the metric (see, {\em e.g.} \cite{brewin2009riemann}), taking determinants, and using $\sqrt{1 + x} = 1 + \frac{1}{2}x + O(x^2)$ yields:
    \begin{equation}
	J_x(z) = \sqrt{\det(g_x)} = 1 - \frac{1}{6}z^i z^j R_{i j}-\frac{1}{12}z^i z^j z^k\nabla_i R_{j k} + O(\|z\|^4) \, \label{eq:appendix_Jac_expansion}
    \end{equation} 
    where $R_{ij}$ is the Ricci curvature tensor, expressed in normal coordinates. As all sectional curvatures are bounded by $K$, $R_{ij} \leq (m-1)Kg_{ij}$. Differentiating \eqref{eq:appendix_Jac_expansion}:
		\begin{align*}
			\| \nabla J_x(z) \| &\lesssim (m-1)K \| z\| + O(\|z\|^2)\quad, \quad  \| H J_x(z) \| \lesssim (m-1)K + O(\|z\|)
		\end{align*}
            As $J_x(0)=1, \nabla J_x(0)=0$ differentiating \eqref{eq:Def_of_rho_star} yields
		\begin{align*}
			\nabla \rho^{*}_x(0) &=J_x(0) \nabla \rho_x(\exp_x(u)) = \nabla \rho(\exp_x(z)) |_{z=0} = \nabla \rho (x) \cdot \text{Jac}(\exp_x(0)) = \nabla \rho (x).
		\end{align*}
		Using $\text{Jac}(\exp_x(0)) = \text{Id}$. Taking norms yields the first claim. From the product rule for Hessians:
		\begin{align*}
			H\rho^{*}_x &= J_x(H\rho\circ\exp_x) + (\nabla \rho\circ\exp_x)^T \nabla J_x + (\nabla J_x)^T\nabla \rho\circ\exp_x + \rho\circ\exp_x (H J_x)
		\end{align*}
		evaluating at $0$ and recalling $J_x(0)=1, \nabla J_x(0) = 0$
		\begin{align*}
			H\rho^{*}_x(0) =H\rho\circ\exp_x(0) + \rho(x) H J_x(0)
		\end{align*}
		Since $\|H J_x(0)\| \lesssim (m-1)K$, it remains to bound $H\rho_x(0)$. 
   From \cite{skorski2019chain} we have the following chain rule for Hessians:
		\begin{align*}
			H(\rho\circ\exp_x) &= (\text{Jac} \exp_x)^T \cdot H\rho(\exp_x) \cdot (\text{Jac} \exp_x) + \sum_{k=1}^D \frac{\partial \rho}{\partial x^k} \cdot H(\exp_x^k)
		\end{align*}
		where $\exp_x^k$ is the $k^{\text{th}}$ coordinate of $\exp_x$ and $x_1, \ldots, x_D$ are full-dimensional Euclidean coordinates. Since $\|H(\exp_x^k)\| \lesssim Km$, evaluating at $z=0$ and taking norms gives:
		\begin{align*}
			\| H\rho\circ\exp_x(0)\| &\leq \| H\rho(x)\| + CKm \| \nabla \rho(x)\|_1 \leq \| H\rho(x)\| + CKm^{\frac{3}{2}} \| \nabla \rho(x)\|
		\end{align*}
		(here we have used the fact that we can choose coordinates so that $\nabla\rho(x)$ is zero except in the first $m$ coordinates, and once again we have ignored the lower order term). Thus we obtain:
		\begin{align*}
			\|H\rho_x^{*}(0)\| &\lesssim \| H\rho(x)\| + mK\rho(x) + Km^{\frac{3}{2}} \| \nabla \rho(x)\|.
		\end{align*}
\end{proof}
\section{Fermat Sectional Curvature}
\label{App:Sec_Curv}
In this section we bound the sectional curvatures of $(\M,g_p)$ in terms of those of $(\M,g)$. This will follow from standard results on conformally equivalent metrics. 

\begin{lem}
Suppose $\rho$ is a density function on $(\M,g)$ satisfying Assumptions~\ref{assump:density}. Let $\varphi = \frac{\alpha}{2}\log(\rho)$ where $\alpha= 2(p-1)/m$. Then:
\begin{enumerate}
    \item $\|\operatorname{grad}\varphi\| \leq \beta\left(\frac{p-1}{m}\right)L_1$
    \item $\|\operatorname{Hess}\varphi\| \leq \beta\left(\frac{p-1}{m}\right)(\beta L_1^2 + L_2)$
\end{enumerate}
\label{lem:Density_phi_bounds}
\end{lem}

\begin{proof}
We compute:
\begin{equation*}
    \operatorname{grad} \varphi = \frac{\alpha}{2}\frac{1}{\rho}\operatorname{grad}\rho = \frac{\alpha}{2\rho}\operatorname{grad}\rho = \frac{p-1}{m\rho}\operatorname{grad} \rho
\end{equation*}
From {\em e.g.} \cite[Proposition 10.43]{boumal2020introduction}, we obtain $\|\operatorname{grad} \rho\| \leq L_1$. Hence:
\begin{equation*}
    \|\operatorname{grad} \varphi\| =  \frac{p-1}{m\rho}\|\operatorname{grad} \rho\| \leq  \beta\left(\frac{p-1}{m}\right)L_1
\end{equation*}
For part 2, a simple calculation reveals
\begin{equation*}
    \Hess\varphi = -\frac{p-1}{d\rho^2}\operatorname{grad}\rho\otimes\operatorname{grad} \rho + \frac{p-1}{d\rho}\Hess\rho .
\end{equation*}
By assumption $\|\Hess\rho\| \leq L_2$, and $\|\operatorname{grad}\rho\otimes\operatorname{grad}\rho\| = \|\operatorname{grad}\rho\|^2 \leq L_1^2$, so
\begin{equation*}
    \left\|\Hess\varphi\right\| \leq  \frac{p-1}{m\rho^2}L_1^2 + \frac{p-1}{m\rho} L_2 \leq \beta\left(\frac{p-1}{m}\right)(\beta L_1^2 + L_2).
\end{equation*}
\end{proof}

\begin{theorem}
	\label{thm:New_Sec_Curv}
	Assume $\rho$ satisfies Assumption \ref{assump:density} and $(\M,g)$ has sectional curvature bounded by $K$. Then $(\M,g_p)$ has sectional curvature bounded by $K_p$ where 
	
	\begin{equation*}
	K_p := \beta^{\alpha}\left(K + \frac{3\beta^{2}(p-1)^2L_1^2}{m^2} + \frac{2\beta^{2}(p-1)L_1^2}{m} + \frac{\beta(p-1)L_2}{m}\right).
	\end{equation*}
\end{theorem}

\begin{proof}
   Let $R$ (resp. $R_p$) denote the $(0,4)$ Riemannian curvature tensor of $(\M,g)$ (resp. $(\M,g_p)$). Let $\phi := -\frac{\alpha}{2}\log\rho$. From \cite[pp. 345]{kuhnel2015differential} (see also \cite[pp. 58]{besse2007einstein}) we have the identity:
   \begin{equation*}
       e^{2\phi} R_p = R - \frac{1}{2}\langle \operatorname{grad}\varphi, \operatorname{grad}\varphi \rangle g\bullet g + \left(\operatorname{Hess}\varphi\right)\bullet g + \left(\operatorname{grad}\varphi\right)^2\bullet g
   \end{equation*}
   where $\bullet$ is the Kulkarni-Nomizu product between symmetric two tensors defined as
   \begin{equation*}
       A\bullet B(X,Y,Z,W) = A(X,Z)B(Y,W) + A(Y,W)B(X,Z) - A(X,W)B(Y,Z) - A(Y,Z)B(X,W),
   \end{equation*}
   see \cite{kuhnel2015differential}[Def. 8.20], and $\left(\operatorname{grad}\varphi\right)^2$ is the symmetric (0,2) tensor defined as
   \begin{equation*}
       \left(\operatorname{grad}\varphi\right)^2(X,Y) = \left(X\varphi\right)\left(Y\varphi\right).
   \end{equation*}
    By the definition of sectional curvature,
   \begin{equation*}
       K_{p,x}(X,Y) = \frac{R_{p,x}(X,Y,Y,X)}{g_{p,x}(X,X)g_{p,x}(Y,Y) - g_{p,x}(X,Y)^2} 
   \end{equation*}
   for any linearly independent $X,Y \in T_x\M$.  It suffices to consider only orthonormal $X,Y$, in which case $K_x(X,Y) = R_x(X,Y,Y,X)$. As $g$ and $g_p$ are conformally equivalent, $X,Y$ are orthonormal with respect to $g$ if and only if they are orthonormal with respect to $g_p$. Assuming orthonormality, some straightforward calculations reveal
   \begin{align*}
       & g\bullet g(X,Y,Y,X) = -2, \\
       & \left(\operatorname{grad}\varphi\right)^2\bullet g (X, Y, Y, X) = -\left(g(\operatorname{grad}\varphi,X)\right)^2 -  \left(g(\operatorname{grad}\varphi,Y)\varphi\right)^2, \\
       & \left(\operatorname{Hess}\varphi\right)\bullet g(X, Y, Y, X) = -\operatorname{Hess}\varphi(X,X) - \operatorname{Hess}\varphi(Y,Y).
   \end{align*}
   Now,
   \begin{align*}
       \left|K_{p,x}(X,Y)\right| &=  \left|R_{p,x}(X,Y,Y,X)\right|\\
                                     & \leq e^{2\varphi}\left(|R(X,Y,Y,X)| + \left|\frac{1}{2}\langle \operatorname{grad}\varphi, \operatorname{grad}\varphi \rangle g\bullet g (X,Y,Y,X)\right| \right. \\
                                     & \left. + \left|\left(\operatorname{Hess}\varphi\right)\bullet g(X,Y,Y,X)\right| + \left|\left(\nabla\varphi\right)^2\bullet g(X,Y,Y,X)\right|\right) \\
                                     & \leq e^{2\varphi}\left( K + \frac{1}{2}\|\operatorname{grad}\varphi\|^2\left|-2\right| + \right. \\
                                     & \left. + \left|\operatorname{Hess}\varphi(X, X) \right| + \left|\operatorname{Hess}\varphi(Y, Y)\right| + \left(g(\operatorname{grad}\varphi,X)\right)^2 +  \left(g(\operatorname{grad}\varphi,Y)\right)^2 \right) \\
                                     & \stackrel{(a)}{\leq} (\rho^{\alpha}) \left(K + \frac{(p-1)^2\beta^2 L_1^2}{m^2} + 2 \beta\left(\frac{p-1}{m}\right)(\beta L_1^2 + L_2)  +  \frac{2(p-1)^2\beta^2 L_1^2}{m^2} \right) \\
                                     & = \rho^{\alpha}\left(K + \frac{3(p-1)^2\beta^2 L_1^2}{m^2} + \frac{2(p-1)\beta^2 L_1^2}{m} + \frac{(p-1)\beta L_2}{m}\right) \\
                                     & \leq \beta^{\alpha}\left(K + \frac{3(p-1)^2\beta^2 L_1^2}{m^2} + \frac{2(p-1)\beta^2 L_1^2}{m} + \frac{(p-1)\beta L_2}{m}\right).
   \end{align*}  
\end{proof}
where in (a) we use various bounds from Lemma~\ref{lem:Density_phi_bounds}.
\section{Fermat Geodesics}
\label{app:FermatGeo}

This appendix contains the proof of Theorem \ref{thm:FermatGeodesics}.

\thmFermatGeodesics*

\begin{proof}
	We compute the geodesic equations for $(\Omega,g_p)$ for geodesics starting at some $y\in\Omega$. Without loss of generality, assume $y=0$ and for notational convenience, let $\rho_{0}$, $\nabla \rho_{0}$, and $H_{0}$ denote $\rho(y)=\rho(0)=\rho_{0}$, $\nabla \rho(y)=\nabla \rho(0)=\nabla \rho_{0}$, and $H(y)=H(0)=H_{0}$, respectively; also let $g^p_{ij}$ denote $(g_p)_{ij}$.  We have global coordinates on $\Omega$ which are just the standard coordinates $(x^i)$.  The function $x(t)=(x^1(t),\ldots,x^m(t))$ is a $g_p$ geodesic if and only if it satisfies for all $k=1,\dots, d$ the geodesic equation  
	\begin{align*}
		\ddot{x}^{k}(t) + \dot{x}^i(t)\dot{x}^j(t) \Gamma^k_{ij}(x(t)) &=0,
	\end{align*}
	where
	\[\Gamma^k_{ij}= \frac{1}{2}g_p^{k\ell}(\partial_i g^p_{j\ell}+\partial_jg^p_{i\ell}-\partial_\ell g^p_{ij})\]
	are the Christoffel symbols associated to $g_p$ \citep{lee2006riemannian}.   
	Now, $g^p_{ij} = \rho^{-\alpha} g_{ij}$, where $g_{ij} = \langle \partial_i, \partial_j \rangle = \delta_{ij}$ is the Euclidean metric tensor on $\mathbb{R}^m$. In addition, since $g_{ij}$ is the identity matrix, $g_p^{ij} = (g^p_{ij})^{-1} = \rho^{\alpha}  (g_{ij})^{-1} =\rho^{\alpha}\langle \partial_i, \partial_j \rangle=\rho^{\alpha}\delta^{ij}$ and $\partial_i g^p_{j\ell}=\partial_i (\rho^{-\alpha} g_{j\ell})= \delta_{j\ell}\partial_i \rho^{-\alpha}$. Thus:
	\begin{align*}
		\Gamma^k_{ij} &= \frac{1}{2}g_p^{k\ell}(\partial_i g^p_{j\ell}+\partial_jg^p_{i\ell}-\partial_\ell g^p_{ij}) \\
		&= \frac{1}{2} \rho^{\alpha}\delta^{k\ell}\left( \delta_{j\ell}\partial_i \rho^{-\alpha} + \delta_{i\ell}\partial_j \rho^{-\alpha} - \delta_{ij}\partial_\ell \rho^{-\alpha}\right) \\
		&= \frac{1}{2} \rho^{\alpha}\left( \delta^{k}_{j}\partial_i \rho^{-\alpha} + \delta^{k}_{i}\partial_j \rho^{-\alpha} - \delta_{ij}\delta^{k\ell}\partial_{\ell} \rho^{-\alpha}\right) \, .
	\end{align*}
	So, for $k$ fixed:
	\begin{align*}
	\dot{x}^i \dot{x}^j \Gamma^k_{ij} &= \dot{x}^i \dot{x}^j  \frac{1}{2} \rho^{\alpha}\left( \delta^{k}_{j}\partial_i \rho^{-\alpha} + \delta^{k}_{i}\partial_j \rho^{-\alpha} - \delta_{ij}\delta^{k\ell}\partial_{\ell} \rho^{-\alpha}\right) \\
		&=\frac{1}{2}\rho^{\alpha} \dot{x}^i\dot{x}^k \partial_i\rho^{-\alpha} + \frac{1}{2}\rho^{\alpha} \dot{x}^k\dot{x}^j \partial_j\rho^{-\alpha} - \frac{1}{2}\rho^{\alpha}\dot{x}^i\dot{x}_i \delta^{k\ell}\partial_{\ell}\rho^{-\alpha} \\
        &= \frac{1}{2}\rho^{\alpha}\left(\dot{x}^i\partial_i\rho^{-\alpha} + \dot{x}^j\partial_j\rho^{-\alpha}\right)\dot{x}^k - \frac{1}{2}\rho^{\alpha}\dot{x}^i\dot{x}_i \delta^{k\ell}\partial_{\ell}\rho^{-\alpha} \\
	&= \rho^{\alpha} \langle \dot{x}, \nabla \rho^{-\alpha} \rangle \dot{x}^k - \frac{1}{2} \rho^{\alpha} \langle \dot{x}, \dot{x} \rangle \partial_k\rho^{-\alpha}.
	\end{align*}
	Where in the last line we have switched to vector notation and identified $d\rho^{-\alpha}$ with $\nabla \rho^{-\alpha}$ (which is valid as the metric is the Euclidean one). Thus, in vector notation the geodesic equation becomes:
	\begin{align}
		\label{equ:geo_equ}
		\ddot{x} + \rho^{\alpha} \langle \dot{x}, \nabla \rho^{-\alpha} \rangle \dot{x} - \frac{1}{2} \rho^{\alpha} \langle \dot{x}, \dot{x} \rangle \nabla \rho^{-\alpha} &= 0.
	\end{align}
	For a unit Euclidean norm vector $b\in\mathbb{R}^{m}$, consider the initial value problem 
	\begin{align*}
		x(0) &= 0, \\
		\dot{x}(0) &= \rho_{0}^{\frac{\alpha}{2}} b, 
	\end{align*}
	which has a unique geodesic solution $\gamma_b(t)$ in some local neighborhood \citep{lee2006riemannian}. 
	Note $|\dot{x}(0)|_p =   \rho_{0}^{-\frac{\alpha}{2}} \|  \rho_{0}^{\frac{\alpha}{2}} b \| = \| b\| = 1$ where $|\cdot|_p$ denotes the norm with respect to $g_p$, so $\gamma_b(t)$ is the unit speed geodesic in the direction of $b$.  Because $\rho$ is assumed $\mathcal{C}^{\infty}$ and bounded away from 0, we have that each $g^p_{ij}$ is $C^{\infty}$, hence the Christoffel symbols are $\mathcal{C}^{\infty}$.  This implies the geodesic $\gamma_{b}(t)$ is also $\mathcal{C}^{\infty}$, and hence we can Taylor expand it about $t=0$ to obtain:
	\begin{align*}
		\gamma_b(t) &= \rho_{0}^{\frac{\alpha}{2}} b t + vt^2 + qt^3 + O(t^4), \\
		\dot{\gamma}_b(t) &=  \rho_{0}^{\frac{\alpha}{2}} b + 2vt + 3qt^2 + O(t^3), \\
		\ddot{\gamma}_b(t) &=  2v + 6qt + O(t^2),
	\end{align*}
	for some vectors $v,q$ depending on $b,\rho$, which can be computed from the leading order terms in \eqref{equ:geo_equ}.
	
	We have the following Taylor expansions for $\rho, \nabla \rho, \rho^{\alpha}, \nabla \rho^{-\alpha}$:
	\begin{align*}
		\rho(x) &= \rho_{0} + \langle \nabla \rho_{0}, x\rangle + \frac{1}{2}x^{T}H_{0}x+O(x^3), \\
		\nabla \rho(x) &= \nabla \rho_{0} + H_{0}x + O(x^2), \\
		\rho(x)^{\alpha} &= \rho_{0}^{\alpha} + \alpha \rho_{0}^{\alpha-1} \langle \nabla \rho_{0}, x\rangle + O(x^2), \\
		\nabla \rho(x)^{-\alpha} &= -\alpha \rho(x)^{-\alpha -1} \nabla \rho(x)  \\ 
		&= -\alpha(\rho_{0} + \langle \nabla \rho_{0}, x\rangle +O(x^2) )^{-\alpha -1}( \nabla \rho_{0} + H_{0}x + O(x^2)) \\
		&=-\alpha \rho_{0}^{-\alpha-1}(1 + \rho_{0}^{-1}\langle \nabla \rho_{0}, x\rangle +O(x^2) )^{-\alpha -1}( \nabla \rho_{0} + H_{0}x + O(x^2)) \\
		&=-\alpha \rho_{0}^{-\alpha-1}(1 - (\alpha+1)\rho_{0}^{-1}\langle \nabla \rho_{0}, x\rangle +O(x^2) )( \nabla \rho_{0} + H_{0}x + O(x^2)) \\
		&= -\alpha \rho_{0}^{-\alpha-1}(\nabla \rho_{0} - (\alpha+1)\rho_{0}^{-1}\langle \nabla \rho_{0}, x\rangle \nabla \rho_{0}+H_{0}x + O(x^2)) \\
		&= -\alpha \rho_{0}^{-\alpha-1}\nabla \rho_{0} + \alpha(\alpha+1)\rho_{0}^{-\alpha-2}\langle \nabla \rho_{0}, x\rangle \nabla \rho_{0}-\alpha \rho_{0}^{-\alpha-1}H_{0}x + O(x^2).
	\end{align*}
	Substituting $x=\gamma_b(t)$, we obtain for some vectors $z,w$ the following Taylor expansions in terms of $t$:
	\begin{align*}
		\rho^{\alpha} &= \rho_{0}^{\alpha}+\alpha \rho_{0}^{\frac{3}{2}\alpha-1} \langle \nabla \rho_{0} , b\rangle t + O(t^2) \\
		&:= \rho_{0}^{\alpha}+z t + O(t^2),\\
		\nabla \rho^{-\alpha} &=  -\alpha \rho_{0}^{-\alpha-1}\nabla \rho_{0} + \alpha(\alpha+1)\rho_{0}^{-\frac{\alpha}{2}-2}\langle \nabla \rho_{0}, b\rangle \nabla \rho_{0} t-\alpha \rho_{0}^{-\frac{\alpha}{2}-1}H_{0}bt + O(t^2) \\
		&:= -\alpha \rho_{0}^{-\alpha-1}\nabla \rho_{0} + wt + O(t^2), \\
		\langle \dot{\gamma}_b, \nabla \rho^{-\alpha} \rangle &= \langle  \rho_{0}^{\frac{\alpha}{2}} b + 2vt , -\alpha \rho_{0}^{-\alpha-1}\nabla \rho_{0} + wt \rangle + O(t^2) \\
		&= -\alpha \rho_{0}^{-\frac{\alpha}{2}-1}\langle b,\nabla \rho_{0}\rangle + \rho_{0}^{\frac{\alpha}{2}} \langle b,w\rangle t - 2\alpha \rho_{0}^{-\alpha-1}\langle v, \nabla \rho_{0}\rangle t + O(t^2),\\
		\langle \dot{\gamma}_b,\dot{\gamma}_b\rangle &= \langle  \rho_{0}^{\frac{\alpha}{2}} b + 2vt ,\rho_{0}^{\frac{\alpha}{2}} b + 2vt \rangle +O(t^2) \\
		&= \rho_{0}^{\alpha} + 4\rho_{0}^{\frac{\alpha}{2}}\langle b,v\rangle t + O(t^2).
	\end{align*}
	To solve for $v$, we only need to compute the leading order term in the geodesic equation \eqref{equ:geo_equ}. 
	We have:
	\begin{align*}
		\rho^{\alpha} \langle \dot{\gamma}_b, \nabla \rho^{-\alpha} \rangle \dot{\gamma}_b &= \rho_{0}^{\alpha} (-\alpha \rho_{0}^{-\frac{\alpha}{2}-1}\langle b,\nabla \rho_{0}\rangle) \rho_{0}^{\frac{\alpha}{2}} b+ O(t) \\
		&= -\alpha \rho_{0}^{\alpha-1}\langle b, \nabla \rho_{0}\rangle b +O(t)
	\end{align*}
	as well as
	\begin{align*}
		- \frac{1}{2} \rho^{\alpha} \langle \dot{\gamma}_b, \dot{\gamma}_b \rangle \nabla \rho^{-\alpha} 
		&= -\frac{1}{2}\rho_{0}^{\alpha}\left(\rho_{0}^{\alpha}\right)\left(-\alpha \rho_{0}^{-\alpha-1}\nabla \rho_{0}\right)+O(t) \\
		&=\frac{\alpha}{2} \rho_{0}^{\alpha-1} \nabla \rho_{0} + O(t).
	\end{align*}
	Since $\ddot{\gamma}_b(t) = 2v + O(t)$, plugging into \eqref{equ:geo_equ} gives:
	\begin{align*}
		2v - \alpha \rho_{0}^{\alpha-1} \langle b, \nabla \rho_{0} \rangle b +\frac{\alpha}{2} \rho_{0}^{\alpha-1} \nabla \rho_{0} + O(t) &= 0.
	\end{align*}
	Since the above must hold for arbitrarily small $t$, we obtain
	\begin{align*}
		v &= \alpha \rho_{0}^{\alpha -1} \left(\frac{1}{2}\langle b, \nabla \rho_{0}\rangle b-\frac{1}{4} \nabla \rho_{0}\right).
	\end{align*}
	We have thus established that the unit geodesic in direction $b$ has form:
	\begin{align*}
		\gamma_b(t) &=  \rho_{0}^{\frac{\alpha}{2}} b t + \alpha \rho_{0}^{\alpha -1} \left(\frac{1}{2}\langle b, \nabla \rho_{0}\rangle b-\frac{1}{4} \nabla \rho_{0}\right)t^2 + O(t^3) \, .
	\end{align*}
	We can now compute $q$ to obtain a higher order expansion; we need to compute the linearization of each term in the geodesic equation. We have three terms:
	\begin{align*}
		(\text{I}) &:= \ddot{\gamma}_b = 2v + 6qt + O(t^2) \\
		(\text{II})&:= \rho^{\alpha} \langle \dot{\gamma}_b, \nabla \rho^{-\alpha} \rangle \dot{\gamma}_b \\
		&= ( \rho_{0}^{\alpha}+z t ) (-\alpha \rho_{0}^{\frac{-\alpha}{2}-1}\langle b,\nabla \rho_{0}\rangle + \rho_{0}^{\frac{\alpha}{2}} \langle b,w\rangle t - 2\alpha \rho_{0}^{-\alpha-1}\langle v, \nabla \rho_{0}\rangle t) (\rho_{0}^{\frac{\alpha}{2}} b + 2vt) + O(t^2) \\
		&= -\alpha \rho_{0}^{\alpha-1} \langle b, \nabla \rho_{0} \rangle b - \alpha \rho_{0}^{-1}z\langle b, \nabla \rho_{0}\rangle b t  + \rho_{0}^{\frac{3}{2}\alpha}b\left[\rho_{0}^{\frac{\alpha}{2}}\langle b, w\rangle t - 2\alpha \rho_{0}^{-\alpha-1}\langle v, \nabla \rho_{0}\rangle t \right] \\
		&\qquad -2\alpha \rho_{0}^{\frac{\alpha}{2}-1} \langle b, \nabla \rho_{0}\rangle v t + O(t^2)\\
		&=  -\alpha \rho_{0}^{\alpha-1} \langle b, \nabla \rho_{0} \rangle b - \alpha \rho_{0}^{-1}z\langle b, \nabla \rho_{0}\rangle b t  + \rho_{0}^{2\alpha}\langle b, w\rangle bt - 2\alpha \rho_{0}^{\frac{\alpha}{2}-1}\langle v, \nabla \rho_{0}\rangle bt \\
		&\qquad -2\alpha \rho_{0}^{\frac{\alpha}{2}-1} \langle b, \nabla \rho_{0}\rangle v t + O(t^2) \\
		&= -\alpha \rho_{0}^{\alpha-1} \langle b, \nabla \rho_{0} \rangle b + \left(-\alpha \rho_{0}^{-1}z\langle b, \nabla \rho_{0}\rangle  + \rho_{0}^{2\alpha}\langle b, w\rangle -2\alpha \rho_{0}^{\frac{\alpha}{2}-1}\langle v, \nabla \rho_{0}\rangle\right) bt \\
		&\qquad -2\alpha \rho_{0}^{\frac{\alpha}{2}-1} \langle b, \nabla \rho_{0}\rangle v t + O(t^2) \\
		&= -\alpha \rho_{0}^{\alpha-1} \langle b, \nabla \rho_{0} \rangle b + \left(-\alpha \rho_{0}^{-1}z\langle b, \nabla \rho_{0}\rangle  + \rho_{0}^{2\alpha}\langle b, w\rangle- 2\alpha \rho_{0}^{\frac{\alpha}{2}-1}\langle v, \nabla \rho_{0}\rangle\right) bt \\
		&\qquad -2\alpha \rho_{0}^{\frac{\alpha}{2}-1} \langle b, \nabla \rho_{0}\rangle -\alpha \rho_{0}^{\alpha -1} \left(\frac{1}{4} \nabla \rho_{0} - \frac{1}{2}\langle b, \nabla \rho_{0}\rangle b\right) t  + O(t^2) \\
		&=  -\alpha \rho_{0}^{\alpha-1} \langle b, \nabla \rho_{0} \rangle b + C_1 bt + \frac{\alpha^2}{2}\rho_{0}^{\frac{3}{2}\alpha-2}\langle b,\nabla \rho_{0}\rangle \nabla \rho_{0} t  + O(t^2) 
	\end{align*}
	where 
	\begin{align*}
		C_1 &= -\alpha \rho_{0}^{-1}z\langle b, \nabla \rho_{0}\rangle  + \rho_{0}^{2\alpha}\langle b, w\rangle- 2\alpha \rho_{0}^{\frac{\alpha}{2}-1}\langle v, \nabla \rho_{0}\rangle - \alpha^2 \rho_{0}^{\frac{3}{2}\alpha-2} \langle b, \nabla \rho_{0}\rangle ^2 \\
		&= -\alpha^2 \rho_{0}^{\frac{3}{2}\alpha-2}\langle b, \nabla \rho_{0}\rangle^2 + \rho_{0}^{2\alpha}\left( -\alpha(-\alpha-1)\rho_{0}^{-\frac{\alpha}{2}-2}\langle \nabla \rho_{0}, b\rangle^2 - \alpha \rho_{0}^{-\frac{\alpha}{2}-1}\langle H_{0}b, b\rangle\right) \\
		&\qquad -2\alpha \rho_{0}^{\frac{\alpha}{2}-1}\left(-\frac{\alpha}{4}\rho_{0}^{\alpha-1} \langle \nabla \rho_{0}, \nabla \rho_{0} \rangle +\frac{\alpha}{2}\rho_{0}^{\alpha-1}\langle b, \nabla \rho_{0}\rangle^2\right) - \alpha^2\rho_{0}^{\frac{3}{2}\alpha-2}\langle b,\nabla \rho_{0}\rangle^2 \\
		&= -2\alpha^2 \rho_{0}^{\frac{3}{2}\alpha-2}\langle b,\nabla \rho_{0}\rangle^2 + (\alpha^2+\alpha)\rho_{0}^{\frac{3}{2}\alpha-2}\langle b,\nabla \rho_{0}\rangle^2 - \alpha \rho_{0}^{\frac{3}{2}\alpha-1}\langle H_{0}b, b\rangle \\
		&\qquad+ \frac{\alpha^2}{2}\rho_{0}^{\frac{3}{2}\alpha-2}\langle \nabla \rho_{0},\nabla \rho_{0}\rangle - \alpha^2 \rho_{0}^{\frac{3}{2}\alpha-2}\langle b,\nabla \rho_{0}\rangle^2 \\
		&= (-2\alpha^2 +\alpha)\rho_{0}^{\frac{3}{2}\alpha-2}\langle b,\nabla \rho_{0}\rangle^2 - \alpha \rho_{0}^{\frac{3}{2}\alpha-1}\langle H_{0}b, b\rangle + \frac{\alpha^2}{2}\rho_{0}^{\frac{3}{2}\alpha-2}\langle \nabla \rho_{0},\nabla \rho_{0}\rangle.
	\end{align*}
	Similarly:
	\begin{align*}
		(\text{III})&:=- \frac{1}{2} \rho^{\alpha} \langle \dot{\gamma}_b, \dot{\gamma}_b \rangle \nabla \rho^{-\alpha} \\
		&= - \frac{1}{2}\left( \rho_{0}^{\alpha} + zt\right)\left(\rho_{0}^{\alpha} + 4\rho_{0}^{\frac{\alpha}{2}}\langle b,v\rangle t \right) \left(-\alpha \rho_{0}^{-\alpha-1}\nabla \rho + wt  \right)+O(t^2) \\
		&=- \frac{1}{2}\left( -\alpha \rho_{0}^{\alpha-1}\nabla \rho_{0} + \rho_{0}^{2\alpha}wt - 4\alpha \rho_{0}^{\frac{\alpha}{2}-1}\langle b,v\rangle \nabla \rho_{0} t - \alpha \rho_{0}^{-1} z\nabla \rho_{0} t \right)+O(t^2) \\
		&= \frac{\alpha}{2} \rho_{0}^{\alpha-1}\nabla \rho_{0} - \frac{1}{2} \rho_{0}^{2\alpha}wt +2\alpha \rho_{0}^{\frac{\alpha}{2}-1}\langle b,v\rangle \nabla \rho_{0} t +\frac{\alpha}{2} \rho_{0}^{-1} z\nabla \rho t +O(t^2) \\
		&= \frac{\alpha}{2} \rho_{0}^{\alpha-1}\nabla \rho_{0} - \frac{1}{2} \rho_{0}^{2\alpha}\left( (\alpha^2+\alpha)\rho_{0}^{-\frac{\alpha}{2}-2}\langle \nabla \rho_{0}, b\rangle \nabla \rho_{0} - \alpha \rho_{0}^{-\frac{\alpha}{2}-1}H_{0}b\right)t \\
		&\qquad 2\alpha \rho_{0}^{\frac{\alpha}{2}-1}\langle b,v\rangle \nabla \rho_{0} t +\frac{\alpha}{2} \rho_{0}^{-1} (\alpha \rho_{0}^{\frac{3}{2}\alpha-1})\langle \nabla \rho_{0},b\rangle\nabla \rho_{0} t +O(t^2) \\
		&= \frac{\alpha}{2} \rho_{0}^{\alpha-1}\nabla \rho_{0} + \left(-\frac{\alpha}{2}-\frac{\alpha^2}{2}\right)\rho_{0}^{\frac{3}{2}\alpha-2}\langle\nabla \rho_{0}, b\rangle\nabla \rho_{0} t + \frac{\alpha}{2}\rho_{0}^{\frac{3}{2}\alpha-1}H_{0}b t \\
		&\qquad + 2\alpha \rho_{0}^{\frac{\alpha}{2}-1}\langle b, v\rangle \nabla \rho_{0} t + \frac{\alpha^2}{2}\rho_{0}^{\frac{3}{2}\alpha-2}\langle\nabla \rho_{0}, b\rangle\nabla \rho_{0} t+O(t^2) \\
		&= \frac{\alpha}{2} \rho_{0}^{\alpha-1}\nabla \rho_{0} +C_2\nabla \rho_{0} t + \frac{\alpha}{2}\rho_{0}^{\frac{3}{2}\alpha-1}H_{0}b t +O(t^2)
	\end{align*}
	where
	\begin{align*}
		C_2 &=  -\frac{\alpha}{2}\rho_{0}^{\frac{3}{2}\alpha-2}\langle\nabla \rho_{0}, b\rangle+ 2\alpha \rho_{0}^{\frac{\alpha}{2}-1}\langle b, v\rangle \\
		&=-\frac{\alpha}{2}\rho_{0}^{\frac{3}{2}\alpha-2}\langle\nabla \rho_{0}, b\rangle+ 2\alpha \rho_{0}^{\frac{\alpha}{2}-1}\left(-\frac{\alpha}{4}\rho_{0}^{\alpha-1}\langle\nabla \rho_{0}, b\rangle + \frac{\alpha}{2}\rho_{0}^{\alpha-1}\langle b,\nabla \rho_{0}\rangle\right) \\
		&= \left(\frac{\alpha^2}{2}-\frac{\alpha}{2}\right)\rho_{0}^{\frac{3}{2}\alpha-2}\langle \nabla \rho_{0},b\rangle.
	\end{align*}
	The linear terms in the geodesic equation must sum to zero, and we obtain:
	\begin{align*}
		6qt + C_1 bt + \frac{\alpha^2}{2}\rho_{0}^{\frac{3}{2}\alpha-2}\langle b,\nabla \rho_{0}\rangle \nabla \rho_{0} t +C_2\nabla \rho_{0} t + \frac{\alpha}{2}\rho_{0}^{\frac{3}{2}\alpha-1}H_{0}b t &=0 \\
		\implies \quad 6q + C_1 b  +C_3\nabla \rho_{0}  + \frac{\alpha}{2}\rho_{0}^{\frac{3}{2}\alpha-1}H_{0}b &=0
	\end{align*}
	where 
	\begin{align*}
		C_3 &= C_2 + \frac{\alpha^2}{2}\rho_{0}^{\frac{3}{2}\alpha-2}\langle b,\nabla \rho_{0}\rangle = \left(\alpha^2-\frac{\alpha}{2}\right)\rho_0^{\frac{3}{2}\alpha-2}\langle b, \nabla\rho_0 \rangle.
	\end{align*}
	We thus obtain:
	\begin{align*}
		q &=- \frac{1}{6}C_1 b- \frac{\alpha}{12}\rho_{0}^{\frac{3}{2}\alpha-1}H_{0}b  -\frac{C_3}{6}\nabla \rho_{0}, \\
		&= C_1' b + C_2' H_{0}b + C_3' \nabla \rho_{0},
	\end{align*}
	where
	\begin{align*}
		C_1' 
		&= \left(\frac{1}{3}\alpha^2 -\frac{1}{6}\alpha\right)\rho_{0}^{\frac{3}{2}\alpha-2}\langle b,\nabla \rho_{0}\rangle^2 + \frac{\alpha}{6} \rho_{0}^{\frac{3}{2}\alpha-1}\langle Hb, b\rangle - \frac{\alpha^2}{12}\rho_{0}^{\frac{3}{2}\alpha-2}\langle \nabla \rho_{0},\nabla \rho_{0}\rangle, \\
		C_2' &= -\frac{\alpha}{12}\rho_{0}^{\frac{3}{2}\alpha-1}, \\
		C_3' &= \left(\frac{\alpha}{12}-\frac{\alpha^2}{6}\right)\rho_{0}^{\frac{3}{2}\alpha-2}\langle \nabla \rho_{0},b\rangle.
	\end{align*}
	In summary we have the following geodesic expansion depending on $b, \rho_{0}, \nabla \rho_{0}, H_{0}$:
	\begin{align*}
		\gamma_b(t) &=  \rho_{0}^{\frac{\alpha}{2}} b t + \alpha \rho_{0}^{\alpha -1} \left(\frac{1}{2}\langle b, \nabla \rho_{0}\rangle b-\frac{1}{4} \nabla \rho_{0}\right)t^2 + (C_1' b + C_2' H_{0}b + C_3' \nabla \rho_{0})t^3 + O(t^4) \, .
	\end{align*}
	
\end{proof}

\section{Local Euclidean Equivalence}
\label{app:EucEquiv}

This appendix contains the proof of Theorem \ref{thm:HO_equiv}.

\thmHOequiv*

\begin{proof}
	Without loss of generality assume $x=0$, and let $\epsilon = \| y\|$. Now consider $B(0,\epsilon)$, a Euclidean ball of radius $\epsilon$ about 0. As long as $\epsilon$ is not too large, each point on $\partial B(0,\epsilon)$ is on a unique $\mathcal{L}_p^p$ geodesic curve leaving the origin; let $\gamma_b$ be the geodesic which goes through $y$ (recall it is unit speed in the direction of unit vector $b$). Note in general $b \ne u_y= \frac{y}{\|y\|}$, but these vectors are close for $\epsilon$ small.
	
For notational brevity 
we denote $\mathcal{L}_p^p(x,y), \rho(0), \nabla\rho(0), H(0)$ by $\mathcal{L},\rho_0,\nabla\rho_0,H_0$
throughout the proof.  Note since the geodesic is unit speed, $\gamma_b$ reaches $y$ at time $\mathcal{L}$, {\em i.e.,} $y = \gamma_b(\mathcal{L})$. 
	By Theorem \ref{thm:FermatGeodesics}, we have
	\begin{align}
	\label{equ:exp_of_y}
		\gamma_b(\mathcal{L}) &=  \rho_{0}^{\frac{\alpha}{2}} b \mathcal{L} + \alpha \rho_{0}^{\alpha -1} \left(\frac{1}{2}\langle b, \nabla\rho_{0}\rangle b-\frac{1}{4} \nabla\rho_{0}\right)\mathcal{L}^2 + (C_1'b+C_2'H_{0}b+C_3'\nabla\rho_{0})\mathcal{L}^3 + O(\mathcal{L}^4) \, .
	\end{align}
	Thus for
	\begin{align*}
		C_4 &= \frac{\alpha^2}{16}\rho_{0}^{2\alpha-2}\langle \nabla\rho_{0}, \nabla\rho_{0}\rangle+2\rho_{0}^{\frac{\alpha}{2}}(C_1'+C_2'\langle H_{0}b, b\rangle +C_3'\langle \nabla\rho_{0}, b\rangle )
	\end{align*}
	we have
	\begin{align*}
		\epsilon &= \| \gamma_b(\mathcal{L}) \| \\
		&= \sqrt{ \rho_{0}^{\alpha} \mathcal{L}^2 - 2\alpha \rho_{0}^{\frac{3}{2}\alpha-1}\left(\frac{1}{4}\langle b,\nabla\rho_{0}\rangle - \frac{1}{2}\langle b, \nabla\rho_{0}\rangle\right)\mathcal{L}^3 + C_4\mathcal{L}^4+O(\mathcal{L}^5) } \\
		&=\sqrt{ \rho_{0}^{\alpha} \mathcal{L}^2 +\frac{1}{2}\alpha \rho_{0}^{\frac{3}{2}\alpha-1}\langle b,\nabla\rho_{0}\rangle \mathcal{L}^3 + C_4\mathcal{L}^4+O(\mathcal{L}^5) } \\
		&= \rho_{0}^{\frac{\alpha}{2}}\mathcal{L} \sqrt{1 +\frac{1}{2}\alpha \rho_{0}^{\frac{\alpha}{2}-1}\langle b,\nabla\rho_{0}\rangle \mathcal{L} + \rho_{0}^{-\alpha}C_4\mathcal{L}^2 + O(\mathcal{L}^3) } \\
		&=\rho_{0}^{\frac{\alpha}{2}}\mathcal{L} \left(1+\frac{1}{4}\alpha \rho_{0}^{\frac{\alpha}{2}-1}\langle b,\nabla\rho_{0}\rangle \mathcal{L} + \left(\frac{C_4}{2}\rho_{0}^{-\alpha}-\frac{1}{32}\alpha^2 \rho_{0}^{\alpha-2}\langle b, \nabla\rho_{0}\rangle^2\right)\mathcal{L}^2 + O(\mathcal{L}^3) \right) \\
		&= \rho_{0}^{\frac{\alpha}{2}}\mathcal{L}  +\frac{1}{4}\alpha \rho_{0}^{\alpha-1}\langle b, \nabla\rho_{0}\rangle \mathcal{L}^2+ \left(\frac{C_4}{2}\rho_{0}^{-\frac{\alpha}{2}}-\frac{1}{32}\alpha^2 \rho_{0}^{\frac{3}{2}\alpha-2}\langle b, \nabla\rho_{0}\rangle^2\right)\mathcal{L}^3 + O(\mathcal{L}^4) 
	\end{align*}
	so that
	\begin{align}
	\label{equ:exp_of_norm_y}
	    \epsilon &= \rho_{0}^{\frac{\alpha}{2}}\mathcal{L}  +\frac{1}{4}\alpha \rho_{0}^{\alpha-1}\langle b, \nabla\rho_{0}\rangle \mathcal{L}^2+ C_4'\mathcal{L}^3 + O(\mathcal{L}^4) 
	\end{align}
	for 
	\begin{align*}
		C_4' &= \frac{1}{2}\rho_{0}^{-\frac{\alpha}{2}}\left(\frac{\alpha^2}{16}\rho_{0}^{2\alpha-2}\langle \nabla\rho_{0}, \nabla\rho_{0}\rangle+2\rho_{0}^{\frac{\alpha}{2}}(C_1'+C_2'\langle H_{0}b, b\rangle +C_3'\langle \nabla\rho_{0}, b\rangle )\right)\\
		&\qquad -\frac{1}{32}\alpha^2 \rho_{0}^{\frac{3}{2}\alpha-2}\langle b, \nabla\rho_{0}\rangle^2 \\
		&= \frac{\alpha^2}{32}\rho_{0}^{\frac{3}{2}\alpha-2}\langle \nabla\rho_{0}, \nabla\rho_{0}\rangle+(C_1'+C_2'\langle H_{0}b, b\rangle +C_3'\langle \nabla\rho_{0}, b\rangle )-\frac{1}{32}\alpha^2 \rho_{0}^{\frac{3}{2}\alpha-2}\langle b, \nabla\rho_{0}\rangle^2 \\
		&= -\frac{5\alpha^2}{96}\rho_{0}^{\frac{3}{2}\alpha-2}\langle \nabla\rho_{0},\nabla\rho_{0}\rangle+\left(\frac{13}{96}\alpha^2-\frac{\alpha}{12}\right)\rho_{0}^{\frac{3}{2}\alpha-2}\langle b,\nabla\rho_{0}\rangle^2  +\frac{\alpha}{12}\rho_{0}^{\frac{3}{2}\alpha-1}\langle H_{0}b, b\rangle \, .
	\end{align*}
	We now relate $b$ with $u_y$ to obtain an expansion independent of $b$. 
	Combining \eqref{equ:exp_of_y} and \eqref{equ:exp_of_norm_y}, we obtain:
	\begin{align*}
	    u_y &= \frac{\rho_0^{\frac{\alpha}{2}}\mathcal{L}\left(b+\alpha\rho_0^{\frac{\alpha}{2}-1}(\frac{1}{2}\langle b,\nabla\rho_{0}\rangle -\frac{1}{4}\nabla\rho_{0})\mathcal{L}+O(\mathcal{L}^2)\right)}{\rho_0^{\frac{\alpha}{2}}\mathcal{L}\left(1+\frac{\alpha}{4}\rho_0^{\frac{\alpha}{2}-1}\langle b, \nabla\rho_{0}\rangle \mathcal{L} + O(\mathcal{L}^2)\right)} \\
	    &= b + \frac{\alpha}{4}\rho_0^{\frac{\alpha}{2}-1}\langle b,\nabla\rho_{0}\rangle b\mathcal{L} - \frac{\alpha}{4}\rho_0^{\frac{\alpha}{2}-1}\nabla\rho_{0} \mathcal{L} + O(\mathcal{L}^2) \, .
	\end{align*}
	Since $b=u_y+O(\mathcal{L})$, we obtain:
	\begin{align*}
	    b &= u_y - \frac{\alpha}{4}\rho_0^{\frac{\alpha}{2}-1}\langle u_y,\nabla\rho_{0}\rangle u_y\mathcal{L} + \frac{\alpha}{4}\rho_0^{\frac{\alpha}{2}-1}\nabla\rho_{0} \mathcal{L} + O(\mathcal{L}^2) \, .
	\end{align*}
	Plugging the above into \eqref{equ:exp_of_norm_y}, we obtain:
	\begin{align}
	\label{equ:eps_from_L}
	    \epsilon &= \rho_0^{\frac{\alpha}{2}}\mathcal{L} + \frac{\alpha}{4}\rho_0^{\alpha-1}\langle u_y , \nabla\rho_{0}\rangle \mathcal{L}^2 + C_5'\mathcal{L}^3+O(\mathcal{L}^4) \, ,
	\end{align}
	where
	\begin{align*}
	    C_5' &= C_4' - \frac{\alpha^2}{16}\rho_0^{\frac{3}{2}\alpha-2} \langle u_y , \nabla\rho_{0}\rangle^2 + \frac{\alpha^2}{16}\rho_0^{\frac{3}{2}\alpha-2} \langle \nabla\rho_{0},\nabla\rho_{0}\rangle \\
	    &= \rho_0^{\frac{3}{2}\alpha-2}\left[ \frac{\alpha^2}{96} \langle \nabla\rho_{0},\nabla\rho_{0}\rangle + \left(\frac{7\alpha^2}{96}-\frac{\alpha}{12}\right)\langle u_y , \nabla\rho_{0}\rangle^2 + \frac{\alpha}{12}\rho_0\langle H_{0} u_y, u_y\rangle\right]  \, ,
	\end{align*}
	and we obtain the first statement in the theorem. 
	Rearranging \eqref{equ:eps_from_L} yields:
	\begin{align}
		\label{equ:expansion_euc_to_geo}
		O(\mathcal{L}^3) +\frac{\alpha}{4} \rho_{0}^{\frac{\alpha}{2}-1}\langle u_y, \nabla\rho_{0}\rangle \mathcal{L}^2 + \mathcal{L} -\rho_{0}^{-\frac{\alpha}{2}}\epsilon &= 0 \, .
	\end{align}
	We now solve for $\mathcal{L}$, which is the root of \eqref{equ:expansion_euc_to_geo} satisfying $\mathcal{L}\sim \epsilon$ as $\epsilon \rightarrow 0$. Expanding $\mathcal{L}(\epsilon) = c_0 + c_1\epsilon + c_2\epsilon^2+\ldots$, plugging into \eqref{equ:expansion_euc_to_geo}, and solving for the coefficients $c_i$, one obtains that any root satisfying $c_0=0$ has form:
	\begin{align*}
	    \mathcal{L} &= \rho_{0}^{-\frac{\alpha}{2}}\epsilon - \frac{\alpha}{4} \rho_{0}^{-\frac{\alpha}{2}-1}\langle u_y, \nabla\rho_{0}\rangle\epsilon^2 + O(\epsilon^3) \\
	    &=\rho_{0}^{-\frac{(p-1)}{m}}\|y\| - \frac{1}{2}\left(\frac{p-1}{m}\right) \rho_{0}^{-\frac{(p-1)}{m}-1}\left\langle u_y , \nabla\rho_{0}\right\rangle \cdot \|y\|^2 + O(\|y\|^3),
	\end{align*}
	which proves the second theorem statement. 
\end{proof}

\section{Metric Approximation}
\label{app:MetricApprox}

\begin{lem}[Fermat Paths are Local]
	\label{lem:local_paths}
	Choose $R > 0$ such that $4R \leq \mathcal{R}$ and suppose $x,y\in \mathcal{B}_z(R)$. Then $\tilde{\ell}_p^p(x,y,H_{n\rho}) = \tilde{\ell}_p^p(x,y,H_{n\rho} \cap \mathcal{B}_z(R))$ with probability $1 - \exp\left(-cn^\frac{1}{m+2p}\right)$.
\end{lem}
\begin{proof}
This is shown in the proof of \cite[Lemma 10]{Hwang2016_Shortest}, see pp. 2807, using the conclusion of  \cite[Corollary 9]{Hwang2016_Shortest}.
\end{proof}
	\nc

	\begin{lem}[Curvature Perturbation for Discrete Metric]
		\label{lem:curv_pert_dis} Suppose  $d(x,y) \leq C_{\M}$. Then
		\begin{align*}
			\tilde{\ell}_p^p(x,y,H_{n\rho} \cap \mathcal{B}_x(R)) &= \left(1\pm Cp(K+\reach^{-2})d(x,y)^2\right)\tilde{\ell}_p^p(0,u,H_{ng_x}) \  .
		\end{align*}
	\end{lem}
	\begin{proof}
		Define $u:= \T_x(y)$. Note $0 = \T_x(x)$ and $d(x,y) = \|\T_x(x)-\T_x(y)\| = \|u\|$. Let $\pi=\{0=\T_x(x_0), \T_x(x_1), \ldots, \T_x(x_L)=u \}$ be the optimal path for $\tilde{\ell}_p^p(0,u,H_{ng_x})$. By \eqref{equ:tangent_plane_dis},
		\begin{align*}
			\|x_{i+1}-x_i\| & \leq d(x_{i+1},x_i) \leq \|\T_x(x_{i+1})-\T_x(x_i)\| + CK \|\T_x(x_{i+1})-\T_x(x_i)\|^3 \\
			& \leq \|\T_x(x_{i+1})-\T_x(x_i)\|\left( 1 + CK\|\T_x(x)-\T_x(y)\|^2\right) \\
			& = \|\T_x(x_{i+1})-\T_x(x_i)\|\left( 1 + CKd(x,y)^2\right).
		\end{align*}
		Thus:
		\begin{align*}
			\tilde{\ell}_p^p(x,y,H_{n\rho} \cap \mathcal{B}_x(R)) &\leq n^{\frac{p-1}{m}} \sum \|x_{i+1}-x_i\|^p \\
			&\leq n^{\frac{p-1}{m}} \sum \|\T_x(x_{i+1})-\T_x(x_i)\|^p\left( 1 + CKd(x,y)^2\right)^p \\
			&\leq (1+CpKd(x,y)^2 +O(d(x,y)^4))\tilde{\ell}_p^p(0,u,H_{ng_x}) \ .
		\end{align*}
		Now let $\pi=\{x=x_0, x_1, \ldots, x_L=y \}$ be the optimal path for $\tilde{\ell}_p^p(x,y,H_{n\rho} \cap \mathcal{B}_x(R))$. Note \eqref{equ:tangent_plane_dis} implies
		\begin{align*}
			\| \T_x(x_{i+1}) - \T_x(x_{i})\| &= d(x_{i+1},x_i) \pm CKd(x_{i+1},x_i)^3 + O(d(x_{i+1},x_i)^5) \, .
		\end{align*} 
		Combining the above with \eqref{equ:Euc_to_geo_dis} gives
		\begin{align*}
			\| \T_x(x_{i+1}) - \T_x(x_{i})\|  &= \|x_{i+1}-x_i\|\left(1\pm C(K+\reach^{-2})\|x_{i+1}-x_i\|^2+O(\|x_{i+1}-x_i\|^4)\right) \\
			&\leq \|x_{i+1}-x_i\|\left(1\pm C(K+\reach^{-2})\|x-y\|^2+O(\|x-y\|^4)\right) \\
			&\leq \|x_{i+1}-x_i\|\left(1\pm C(K+\reach^{-2})d(x,y)^2+O(d(x,y)^4)\right) \, ,
		\end{align*}
		since the optimality of $\pi$ ensures $\|x_{i+1}-x_i\| \leq \|x-y\|$. We thus obtain:
		\begin{align*}
			\tilde{\ell}_p^p(0,u,H_{ng_x}) &\leq n^{\frac{p-1}{m}} \sum \| \T_x(x_{i+1}) - \T_x(x_{i})\|^p \\
			& \leq  n^{\frac{p-1}{m}} \sum \|x_{i+1}-x_i\|^p\left(1+ C(K+\reach^{-2})d(x,y)^2+O(d(x,y)^4)\right)^p \\
			&=n^{\frac{p-1}{m}} \sum \|x_{i+1}-x_i\|^p\left(1+ Cp(K+\reach^{-2})d(x,y)^2+O(d(x,y)^4)\right)\\
			&= \left(1+ Cp(K+\reach^{-2})d(x,y)^2+O(d(x,y)^4)\right)\tilde{\ell}_p^p(x,y,H_{n\rho} \cap \mathcal{B}_x(R)).
		\end{align*} 
		For $d(x,y) \leq C_{\M}$, we can remove the fourth order term by increasing the constant on the second order term, and we obtain
		\begin{align*}
			\frac{\tilde{\ell}_p^p(0,u,H_{ng_x})}{1+ Cp(K+\reach^{-2})d(x,y)^2} \leq \tilde{\ell}_p^p(x,y,H_{n\rho} \cap \mathcal{B}_x(R)) \leq \left(1+ Cp(K+\reach^{-2})d(x,y)^2\right)\tilde{\ell}_p^p(0,u,H_{ng_x}) 
		\end{align*}
		which proves the lemma.
	\end{proof}

	\begin{lem}[Locality of Homogeneous Paths]
		\label{lem:local_geo_homog_PPP} Suppose  \[\mathcal{C}_{p,m,\beta,K} \geq \|u\| \geq (\frac{n}{2\beta})^{-\frac{1}{m}(\frac{1}{3}-\epsilon)}.\] 
        Then 
		$\ell_p^p(0, u, H_{n\gmin}) = \ell_p^p(0,u,\overline{H}_{n\gmin})$ with probability at least $1-C_\epsilon n\exp\left(-c_\epsilon(\frac{n}{2\beta})^{\frac{2\epsilon}{3}\min\{\frac{1}{p},\frac{1}{m}\}}\right)$, and same for $H_{n\gmax}$.
	\end{lem}
	\begin{proof}
		The proof is similar to that of \cite[Thm. 7]{Hwang2016_Shortest}. For completeness, we reprove this lemma here. Suppose, for the sake of contradiction, that the optimal path leaves $B_0(2r)$ where $r := \|u\|$. Then ${\ell_p^p(0,\partial B_0(2r),H_{n\gmin}) \leq \ell_p^p(0,u,H_{n\gmin})}$, where 
		\begin{align*}
			 \ell_p^p(0,\partial B_0(2r),H_{n\gmin}) &= \min_{|v| = 2r}  \ell_p^p(0,v,H_{n\gmin}).
		\end{align*}
        Applying Proposition \ref{prop:conv_PD_homog_PPP} yields
        \begin{equation}
            (n\gmin)^{\frac{p-1}{m}} \ell_p^p(0,\partial B_0(2r),H_{n\gmin}) \leq (n\gmin)^{\frac{p-1}{m}}\ell_p^p(0,u,H_{n\gmin}) \leq \mu\|u\| + \|u\|^2= \mu r + r^2
            \label{eq:Proof_E2_Contradiction_1}
        \end{equation}
         with probability at least $1-C_\epsilon\exp\left(-c_\epsilon(n\gmin)^{\frac{2\epsilon}{3}\min\{\frac{1}{p},\frac{1}{m}\}}\right)$ if $\|u\| \geq (n\gmin)^{-\frac{1}{m}(\frac{1}{3}-\epsilon)}$.
         
	On the other hand, let $\delta = 3rn^{-\frac{1}{m}}$. Then, by  \cite[Cor. 4.2.13]{vershynin2018high}, we may find a $\delta$-net of points $v_1, \ldots, v_n$, as the covering number of $S^{m-1}(2r)$ is upper-bounded by $\left(\frac{3r}{\delta}\right)^{m}$. Again by Proposition \ref{prop:conv_PD_homog_PPP}:
    \begin{align*}
        & (n\gmin)^{\frac{p-1}{m}}\ell_p^p(0,v_i,H_{n\gmin}) \geq \mu\|v_i\| - \|v_i\|^2 = 2\mu r-4r^2 \quad \text{ for } i=1,\ldots, n \\
      \Rightarrow &  \min_{i=1,\ldots,n}  (n\gmin)^{\frac{p-1}{m}}\ell_p^p(0,v_i,H_{n\gmin})\geq \mu\|v_i\| - \|v_i\|^2 = 2\mu r-4r^2 
    \end{align*}
  with probability at least $1-C_\epsilon n\exp\left(-c_\epsilon(n\gmin)^{\frac{2\epsilon}{3}\min\{\frac{1}{p},\frac{1}{m}\}}\right)$, via a union bound. But also 
		\begin{align}
  \begin{split}
			& \min_{|v| = 2r}  \ell_p^p(0,v,H_{n\gmin}) \geq \min_{i=1,\ldots,n}  \ell_p^p(0,v_i,H_{n\gmin}) - \delta^p \\
            \Rightarrow & (n\gmin)^{\frac{p-1}{m}}\ell_p^p(0,\partial B_0(2r),H_{n\gmin}) \geq 2\mu r-4r^2 - \frac{3^pr^pg_{\min}^{\frac{p-1}{m}}}{n^{\frac{1}{m}}}.
             \label{eq:Proof_E2_Contradiction}
            \end{split}
		\end{align}
   Combining \eqref{eq:Proof_E2_Contradiction_1} and \eqref{eq:Proof_E2_Contradiction} yields
   \begin{equation*}
       5r^2 + \frac{3^pr^pg_{\min}^{\frac{p-1}{m}}}{n^{\frac{1}{m}}} \geq \mu r.
   \end{equation*}
    This yields a contradiction; indeed substituting $r := \|u\| \leq \left(Cm\right)^{-1/2}$ we arrive at
    \begin{equation*}
        5(Cm)^{-1} + \left[\frac{3^pg_{\min}^{\frac{p-1}{m}}}{n^{\frac{1}{m}}}\right](Cm)^{-p/2} \geq \mu(Cm)^{-1/2}
    \end{equation*}
    which cannot hold for $C$ small enough and $n$ large enough.
	\end{proof}

	\begin{lem}[Discrete to Continuum Approximation in the Tangent Plane]
		\label{lem:metric_approx_tangent_plane}
		Fix $\epsilon\in(0,1/(8p+6))$ and suppose 
        \begin{equation*}
            \left(\frac{n}{2\beta}\right)^{-\frac{1}{m}(\frac{1}{3}-\epsilon)}\leq \|u\| \leq \mathcal{C}_{p,m,\beta,K} \, .
        \end{equation*}
        For notational convenience define $\L = \mathcal{L}_p^p(0,u,g_x)$. Then we have 
		\begin{align*}
			| \tilde{\ell}_p^p(0,u,H_{ng_x}) - \mu\L | &\leq  \tilde{C}_1\L^2 + \tilde{C}_2\L^3 + O(\L^4)\, 
		\end{align*}
		with probability at least $1-C_\epsilon n\exp\left(-c_\epsilon(\frac{n}{2\beta})^{\frac{2\epsilon}{3}\min\{\frac{1}{m},\frac{1}{p}\}}\right)$, where 
        \begin{align*}
            \tilde{C}_1 &:= \beta^{\frac{p-1}{m}}\left( \frac{5\mu}{2}\left(\frac{p-1}{m}\right)\beta L_1 + 1\right) \ , \ \tilde{C}_2 := C_{p,d,\beta}\left(K(1+L_1) +L_1^2 + L_2 \right) \, .
        \end{align*}
	\end{lem}
	\begin{proof}
        Note throughout the proof we let $C_{p,d,\beta}$ be a constant depending on $p,d,\beta$ whose value may change line to line.
	We know  $\ell_p^p(0, u, H_{n\gmin}) = \ell_p^p(0,u,\overline{H}_{n\gmin})$ and $\ell_p^p(0, u, H_{n\gmax}) = \ell_p^p(0,u,\overline{H}_{n\gmax})$ w.h.p. by Lemma \ref{lem:local_geo_homog_PPP}. Since we can couple the PPPs so that $H_{n\gmin} \subseteq H_{ng_x} \subseteq H_{n\gmax}$, we obtain:
		\begin{align*}
			\ell_p^p(0,u,\overline{H}_{n\gmax} )&\leq  \ell_p^p(0,u,H_{ng_x})  \leq \ell_p^p(0,u,\overline{H}_{n\gmin}) \\
			\implies n^{\frac{(p-1)}{m}} \ell_p^p(0,u,\overline{H}_{n \gmax}) &\leq  n^{\frac{(p-1)}{m}} \ell_p^p(0,u,H_{ng_x}) \leq n^{\frac{(p-1)}{m}} \ell_p^p(0,u,\overline{H}_{n\gmin}).
		\end{align*}
		By applying Proposition \ref{prop:conv_PD_homog_PPP} with $q=2$:
		\begin{equation}
			\ell_p^p(0,u,\overline{H}_{\gmin n})  = \frac{1}{(n\gmin)^{\frac{(p-1)}{m}}}\left(\mu\|u\| \pm \|u\|^2 \right)
            \label{eq:Return_Here}
		\end{equation}
		for $\|u\| \geq (n\gmin)^{-\frac{1}{m}(\frac{1}{3}-\epsilon)}$ with probability at least $1-C_\epsilon\exp\left(-c_\epsilon(n\gmin)^{\frac{2\epsilon}{3}\min\{\frac{1}{m},\frac{1}{p}\}}\right)$. 
		From Theorem \ref{thm:HO_equiv}, we have:
		\begin{align*}
			\|u\| &= g_x(0)^{\frac{p-1}{m}}\mathcal{L}\pm \frac{1}{2}\left(\frac{p-1}{m}\right)g_x(0)^{\frac{2(p-1)}{m}-1}\|\nabla g_x(0)\| \mathcal{L}^2 + C\mathcal{L}^3 + O(\mathcal{L}^4) \\
			\implies \|u\|^2 &= g_x(0)^{\frac{2(p-1)}{m}}\mathcal{L}^2 \pm \left(\frac{p-1}{m}\right)g_x(0)^{\frac{3(p-1)}{m}-1}\|\nabla g_x(0)\| \mathcal{L}^3 + O(\L^4) 
		\end{align*}
		where $|C|\leq C_{p,d,\beta}(\| \nabla g_x(0) \|^2 + \|Hg_x(0)\|)$. Returning to \eqref{eq:Return_Here} we obtain:
		\begin{align*}
			n^{\frac{(p-1)}{m}} &\ell_p^p(0,u,\overline{H}_{n\gmin}) = \frac{1}{\gmin^{\frac{(p-1)}{m}}}\left(\mu\|u\| \pm \|u\|^2 \right) \\
			&\leq \mu\gmin^{\frac{-(p-1)}{m}}\left(g_x(0)^{\frac{p-1}{m}}\mathcal{L} + \frac{1}{2}\left(\frac{p-1}{m}\right)g_x(0)^{\frac{2(p-1)}{m}-1}\|\nabla g_x(0)\| \mathcal{L}^2 +\mu^{-1}g_x(0)^{\frac{2(p-1)}{m}}\mathcal{L}^2+ \tilde{C} \mathcal{L}^3 + O(\L^4)\right) \\
			&\leq \left(\frac{g_x(0)}{\gmin}\right)^{\frac{p-1}{m}}\left(\mu\mathcal{L} + \frac{\mu}{2}\left(\frac{p-1}{m}\right)g_x(0)^{\frac{(p-1)}{m}-1}\|\nabla g_x(0)\| \mathcal{L}^2 +g_x(0)^{\frac{(p-1)}{m}}\mathcal{L}^2+ \tilde{C} \mathcal{L}^3 + O(\L^4)\right)
		\end{align*}
		where 
		\begin{align*}
			\tilde{C} &\leq C_{p,d,\beta} \left(\| \nabla g_x(0) \| +\| \nabla g_x(0) \|^2 + \|Hg_x(0)\|\right).
		\end{align*}
	Since $|g_x(0) - \gmin | \leq 2\nabla g_x(0)\|u\| + 2\|Hg_x(0)\|\cdot\|u\|^2 + O(\|u\|^3)$, we have:
	\begin{align*}
		\frac{g_x(0)}{\gmin} &\leq \frac{g_x(0)}{g_x(0) - 2\|\nabla g_x(0)\|\cdot\|u\|- 2\|Hg_x(0)\|\cdot\|u\|^2 + O(\|u\|^3)} \\
		&\leq 1+ \frac{2\|\nabla g_x(0)\|\cdot\|u\|}{g_x(0)}+ \left(\frac{2\|Hg_x(0)\|}{g_x(0)} +\frac{4\|\nabla g_x(0)\|^2}{g_x(0)^2}\right)\|u\|^2 + O(\|u\|^3)\\
		\implies \left(\frac{g_x(0)}{\gmin}\right)^{\frac{p-1}{m}} &\leq 1+2\left(\frac{p-1}{m}\right)g_x(0)^{-1}\|\nabla g_x(0)\|\cdot\|u\| \\
        &\quad +C_{p,d,\beta}\left(\|\nabla g_x(0)\|^2+\|Hg_x(0)\|\right)\|u\|^2+O(\|u\|^3)
    \end{align*}
    which gives 
    \begin{align*}
        \left(\frac{g_x(0)}{\gmin}\right)^{\frac{p-1}{m}} &= 1+ 2\left(\frac{p-1}{m}\right)g_x(0)^{\frac{p-1}{m}-1}\|\nabla g_x(0)\|\mathcal{L}+C_{p,d,\beta}\left(\|\nabla g_x(0)\|^2 +\|Hg_x(0)\|\right)\mathcal{L}^2+O(\mathcal{L}^3).
    \end{align*}
            Thus:
		\begin{align*}
		&n^{\frac{(p-1)}{m}} \ell_p^p(0,u,\overline{H}_{n\gmin}) \\
            &\qquad\leq \left( 1+ 2\left(\frac{p-1}{m}\right)g_x(0)^{\frac{p-1}{m}-1}\|\nabla g_x(0)\|\mathcal{L}+C_{p,d,\beta}\left(\|\nabla g_x(0)\|^2 +\|Hg_x(0)\|\right)\mathcal{L}^2+O(\mathcal{L}^3) \right)\\
		&\quad\qquad \times \left(\mu\mathcal{L} + \frac{\mu}{2}\left(\frac{p-1}{m}\right)g_x(0)^{\frac{(p-1)}{m}-1}\|\nabla g_x(0)\| \mathcal{L}^2 +g_x(0)^{\frac{(p-1)}{m}}\mathcal{L}^2+ \tilde{C} \mathcal{L}^3 + O(\L^4) \right) \\
		&\qquad = \mu\L + \left( \frac{5\mu}{2}\left(\frac{p-1}{m}\right)g_x(0)^{\frac{(p-1)}{m}-1}\|\nabla g_x(0)\| + g_x(0)^{\frac{p-1}{m}}\right)\L^2 + \tilde{C}_2\L^3+ O(\L^4)
		\end{align*}
		where
		\begin{align*}
			\tilde{C}_2 &\leq C_{p,d,\beta} \left(\| \nabla g_x(0) \| +\| \nabla g_x(0) \|^2 + \|Hg_x(0)\|\right).
		\end{align*}
		A similar argument shows that
		\begin{align*}
			n^{\frac{(p-1)}{m}} \ell_p^p(0,u,\overline{H}_{n\gmax}) \geq \mu\L - \left( \frac{5\mu}{2}\left(\frac{p-1}{m}\right)g_x(0)^{\frac{(p-1)}{m}-1}\|\nabla g_x(0)\| + g_x(0)^{\frac{p-1}{m}}\right)\L^2 - \tilde{C}_2\L^3+ O(\L^4)\, ,
		\end{align*}
		and  we obtain
		\begin{align*}
			| \tilde{\ell}_p^p(0,u,H_{ng_x}) - \mu\mathcal{L}_p^p(0,u,g_x)  | &\leq g_x(0)^{\frac{p-1}{m}} \left( \frac{5\mu}{2}\left(\frac{p-1}{m}\right)\frac{\|\nabla g_x(0)\|}{g_x(0)} + 1\right)\L^2 + \tilde{C}_2\L^3+ O(\L^4)\, .
		\end{align*}
		Note $g_x(0)=\rho(x)$. To finish the proof we use bounds for $\|\nabla g_x(0)\|$, $\|Hg_x(0)\|$ shown in Lemma~\ref{lem:Bound_Hess_expansion} (note $\tilde{C}_2$ depends on $Hg_x(0)$). Altogether we have:
		\begin{align*}
			| \tilde{\ell}_p^p(0,u,H_{ng_x}) - \mu\mathcal{L}_p^p(0,u,g_x)  | &\leq  \rho(x)^{\frac{p-1}{m}}\left( \frac{5\mu}{2}\left(\frac{p-1}{m}\right)\frac{\|\nabla \rho(x)\|}{\rho(x)} + 1\right)\L^2 + \tilde{C}\L^3 + O(\L^4)\, .
		\end{align*}
		with 
		\begin{align*}
			\tilde{C} \leq C_{p,d,\beta}\left(K+K\| \nabla \rho(x) \| +\| \nabla \rho(x) \|^2 + \|H\rho(x)\| \right) \, .
		\end{align*}
		Since $\gmin \geq \frac{1}{\beta}(1-CmK\|u\|^2) \geq \frac{1}{2\beta}$ as long as $\|u\|$ is small enough, the above holds for $\mathcal{C}_{p,m,\beta,K} \geq \|u\| \geq (\frac{n}{2\beta})^{-\frac{1}{m}(\frac{1}{3}-\epsilon)}$ with probability at least $1-C\exp\left(-c(\frac{n}{2\beta})^{\frac{2\epsilon}{3}\min\{\frac{1}{m},\frac{1}{p}\}}\right)$. Bounding $\| \nabla \rho(x) \|\leq L_1$ and $\|H\rho(x)\|\leq L_2$ concludes the proof.
	\end{proof}

	\begin{lem}[Curvature Perturbation, Continuum Metric]
		\label{lem:curv_pert_cont}
		For $\|u\| \leq C_{\M,\rho}$, 
		\begin{align*}
			\mu\mathcal{L}_p^p(0,u,g_x) &= (1\pm CpK\|u\|^2) \mu\mathcal{L}_p^p(x,y) \ .
		\end{align*} 
	\end{lem}
	\begin{proof}
		We first note that for $\|u\| \leq C_{\M, \rho}$, the optimal $\L_p$ path $\gamma \in \M$ stays inside $\mathcal{B}_x(2r)$ and that the optimal path for $\mathcal{L}_p^p(0,u,g_x)$ stays inside $B_0(2r)$ (see for example Theorem \ref{thm:HO_equiv} and Lemma 2.2 from \cite{little2022balancing}). We thus have:
		\begin{align*}
			\mathcal{L}_p^p(0,u,g_x) &= \inf_{\T_x\gamma \in B_0(2r)} \int g_x(\T_x\gamma(t))^{(1-p)/m} |(\T_x\gamma)'(t)|\ dt \\
			&= \inf_{\gamma \in \mathcal{B}_x(2r)} \int \rho(\gamma(t))^{(1-p)/m} J_x(\T_x\gamma(t))^{(1-p)/m} \, |(\T_x\gamma)'(t)|\ dt \\
			&= \inf_{\gamma \in \mathcal{B}_x(2r)} \int \rho(\gamma(t))^{(1-p)/m} J_x(\T_x\gamma(t))^{(1-p)/m} \, |\T_x'(\gamma(t))| \cdot |\gamma'(t)|\ dt \\
			&= \inf_{\gamma \in \mathcal{B}_x(2r)} \int \rho(\gamma(t))^{(1-p)/m} (1\pm CmK\|u\|^2)^{(1-p)/m} \cdot (1\pm CK\|u\|^2) \cdot |\gamma'(t)|\ dt \\
			&= (1\pm CpK\|u\|^2) \inf_{\gamma \in \mathcal{B}_x(2r)} \int \rho(\gamma(t))^{(1-p)/m} |\gamma'(t)|\ dt \\
			&= (1\pm CpK\|u\|^2) \inf_{\gamma \in \M} \int \rho(\gamma(t))^{(1-p)/m} |\gamma'(t)|\ dt \\
			&=(1\pm CpK\|u\|^2)  \mathcal{L}_p^p(x,y).
		\end{align*}
		For the bound on $|\T_x'(\gamma(t))|$, see 1.34 of \cite{Trillos2019_Error} 
	\end{proof}

\section{Fermat Kernels and Degrees}
\label{app:FermatKernelsAndDegrees}

\corkernelcomp*

\begin{proof}
We prove the result by defining $\xi:=2\mu\beta^{\frac{(p-1)}{m}}(n\beta/2)^{-\frac{1}{m}(\frac{1}{3}-\epsilon)} \leq \frac{1}{2}h$ and considering the following 3 cases for any fixed $x_i,x_j$:
	\begin{enumerate}
		\item $\mu\mathcal{L}_p^p < \xi$ 
		\item $\xi \leq \mu\mathcal{L}_p^p \leq 2h$
		\item $\mu\mathcal{L}_p^p > 2h$
	\end{enumerate}
	Case 1: Assume $\mu\mathcal{L}_p^p<\xi$. We want to show that all kernels evaluate to 1, {\em i.e.,} $\mu\L_p^p < \widehat{h}_-$ (which also guarantees $\mu\L_p^p < \widehat{h}_+$) and that $\widetilde{\ell}_p^p < h$. Since $\mu\L_p^p < \xi \leq \frac{1}{2}h = \frac{1}{2}\frac{\widehat{h}_-}{(1-\delta)}\leq \widehat{h}_-$, the $\L_p^p$ kernels evaluate to 1. To bound $\widetilde{\ell}_p^p$, we need to further consider the following two subcases: \\
    Case 1a: $\L_p^p \leq n^{-\frac{1}{m}\left(1-\frac{2}{3p}\right)}$. In this case for $n$ large enough the bound follows by considering the straight-line path, since
    \begin{align*}
        \tilde{\ell}_p^p(x_i,x_j) &\leq n^{\frac{p-1}{m}} \| x_i - x_j \|^p
        \lesssim n^{\frac{p-1}{m}}n^{-\frac{1}{m}\left(p-\frac{2}{3}\right)} = n^{-\frac{1}{m}\left(1-\frac{2}{3}\right)}=n^{-\frac{1}{3m}} \ll n^{-\frac{1}{m}\left(\frac{1}{3}-\epsilon\right)} \sim 2\xi \, ,
    \end{align*}
    so that $\tilde{\ell}_p^p(x_i,x_j)<h$. \\
    Case 1b: $n^{-\frac{1}{m}\left(1-\frac{2}{3p}\right)} \leq \L_p^p$. This case involves discrete-to-continuum control of metric convergence at smaller scales than given in Theorem \ref{thm:metric_approx_iid}. However a repeat of the same arguments used to prove Theorem \ref{thm:metric_approx_iid}, but with choosing $q=1+\frac{1}{4p}$ when applying Proposition \ref{prop:conv_PD_homog_PPP} in the proof of Lemma \ref{lem:metric_approx_tangent_plane}, gives that for $ d(x_i,x_j) \geq 2(n\beta/2)^{-\frac{1}{m}\left(\frac{2p}{2p+1}-\epsilon\right)} $,
    \begin{align*}
    \label{equ:small_scale_metric_dev}
        |\tilde{\ell}_p^p(x_i,x_j) - \mu \L_p^p(x_i,x_j) | &= O( (\L_p^p)^{(1+\frac{1}{4p})})
    \end{align*}
    with probability at least $1-C_\epsilon n \exp\left(-c_\epsilon (\frac{n}{4\beta})^{\frac{\epsilon}{2p+1}\min\{\frac{1}{m},\frac{1}{p}\}} \right)$. We double check the lower bound is satisfied in Case 1b. Note it is sufficient to check that
    \begin{align*}
        d(x_i, x_j) \geq \frac{1}{\beta^{\frac{p-1}{m}}}n^{-\frac{1}{m}\left(1-\frac{2}{3p}\right)} \geq 2(n\beta/2)^{-\frac{1}{m}\left(\frac{2p}{2p+1}-\epsilon\right)}\, ,
    \end{align*}
    i.e. we need $1-\frac{2}{3p}< \frac{1}{1+(1/2p)}-\epsilon$. Since $\epsilon < \frac{1}{2(4p+3)} < \frac{1}{6p}$, we have
    \begin{align*}
       \frac{1}{1+(1/2p)}-\epsilon &\geq  \frac{1}{1+(1/2p)}-\frac{1}{6p}
       \geq 1 - \frac{1}{2p}-\frac{1}{6p} = 1-\frac{2}{3p}
    \end{align*}
    and the lower bound is satisfied for $n$ large enough. We can thus conclude that w.h.p.
    \begin{align*}
    \tilde{\ell}_p^p(x_i, x_j) \leq \mu\mathcal{L}_p^p(x_i, x_j) + O(\xi^{\left(1+\frac{1}{4p}\right)}) = \mu\mathcal{L}_p^p + n^{-\frac{1}{m}\left(\frac{1}{3}-\epsilon\right)\left(1+\frac{1}{4p}\right)} < 2\mu\mathcal{L}_p^p <2\xi
    \end{align*}
    so that $\tilde{\ell}_p^p(x_i,x_j)<h$. \\
Case 2: Note $d(x_i,x_j) \geq \beta^{\frac{1-p}{m}}\L_p^p(x_i,x_j) \geq \beta^{\frac{1-p}{m}} \mu^{-1}\xi = 2(n\beta/2)^{-\frac{1}{m}(\frac{1}{3}-\epsilon)}$, and so by Theorem \ref{thm:metric_approx_iid}:
	\begin{align*}
		\mu\mathcal{L}_p^p(1-\delta) &\leq \widetilde{\ell}_p^p \leq \mu\mathcal{L}_p^p(1+\delta)\\ 
		\implies \frac{\mu\mathcal{L}_p^p}{\widehat{h}_+}=\frac{\mu\mathcal{L}_p^p}{h(1+2\delta)} &\leq \frac{\widetilde{\ell}_p^p}{h} \leq \frac{\mu\mathcal{L}_p^p}{h(1-\delta)} =\frac{\mu\mathcal{L}_p^p}{\widehat{h}_-} 
	\end{align*}
	which establishes the corollary in this case. \\
	Case 3: Now assume that $\mu\mathcal{L}_p^p > 2h$. We want to show that all kernels evaluate to 0, {\em i.e.,} we need to show that $\mathcal{L}_p^p > \widehat{h}_+$ (of course this guarantees $\mathcal{L}_p^p > \widehat{h}_-$) and also that $\widetilde{\ell}_p^p > h$. The first is true since $\mathcal{L}_p^p>2h\geq h(1+2\delta)=\widehat{h}_+$. For $\widetilde{\ell}_p^p$, note that by Theorem \ref{thm:metric_approx_iid}
	\begin{align*}
		\widetilde{\ell}_p^p &> \mu \L_p^p - C_1(\L_p^p)^2 - C_2(\L_p^p)^3 > \frac{\mu}{2} \L_p^p > h
	\end{align*}
	whenever $\L_p^p\leq C_3$ for a fixed constant $C_3$. But when $\L_p^p > C_3$, it is clear that $\widetilde{\ell}_p^p > \frac{\mu}{2}\L_p^p$ from Theorem 1 in \cite{Hwang2016_Shortest} (or from the proof of Proposition \ref{prop:conv_PD_homog_PPP}). 
 
 Finally, to conclude that \eqref{equ:kernel_sandwich} holds for all pairs of points we take a union bound, so that \eqref{equ:kernel_sandwich} holds with probability at least $1-C_\epsilon n^3\exp\left(-c_\epsilon (\frac{n}{4\beta})^{\frac{\epsilon}{2p+1}\min\{\frac{1}{m},\frac{1}{p}\}} \right)$ (note the probability from Case 1 dominates the probability from Case 2, since we applied Proposition \ref{prop:conv_PD_homog_PPP} with $q=1+\frac{1}{4p}$ instead of $q=2$).
 
\end{proof}

\degapprox*

\begin{proof}
   From the proof of Lemma 18 in \cite{Trillos2019_Error}, if we have a discrete metric $\widetilde{d}$ which approximates a geodesic $d$ by:
   \begin{align*}
       \eta\left(\frac{d(x_i,x_j)}{h(1-\delta)}\right) \leq \eta\left(\frac{\widetilde{d}(x_i,x_j)}{h}\right) &\leq \eta\left(\frac{d(x_i,x_j)}{h(1+\delta)}\right)
   \end{align*}
   for all $x_i, x_j$, then for $h \leq C_\M$ (see Assumption 3 in \cite{Trillos2019_Error}) 
   \begin{align*}
       \left| d_i - \rho(x_i) \right| &\leq C\left( L_{\rho}h + \beta m K h^2 + \beta m \delta + \beta m \eta(0)\omega_m \frac{\epsilon}{h}\right)
   \end{align*}
   where $L_\rho$ is the Lipschitz constant of the density associated with the geodesic, $\beta$ bounds this density, $K$ is the maximal sectional curvature associated with the geodesic, and $\epsilon$ is the $\infty$-Wasserstein distance between $\nu$ (measure from which points were sampled) and $\nu_n$ (empirical measure). Note if $d=\mathcal{L}_p^p$ then data is sampled from density $\rho^p$, and $L_\rho,\beta, K,\delta$ all depend on $p$. Specifically we obtain that for $d=\mathcal{L}_p^p$ and $\widetilde{d}= \frac{\widetilde{\ell}_p^p}{\mu}$
    \begin{align*}
       \left| d_i - \rho^p(x_i) \right| &\leq C\left( L_{\rho^p}h + \beta^p d K_p h^2 + \beta^p d \delta + \beta^p d \eta(0)\omega_d \frac{\epsilon}{h}\right)
   \end{align*}
   where $K_p$ is computed in Theorem \ref{thm:New_Sec_Curv}, $L_{\rho^p}$ is the Lipschitz constant of $\rho^p$, and we can choose $\delta = 4C_1h+8C_2\mu h^2$ by Corollary \ref{cor:kernel_comp}. We may take $L_{\rho^p} = C\beta^{p-1}L_1$ as for all $x,y \in \M$
   \begin{equation*}
       |\rho^p(x) - \rho^p(y)| \leq |\rho(x) - \rho(y)|\sum_{j=0}^{p-1}|\rho^{p-j-1}(x)||\rho^j(y)| \leq \left(L_1\right)\left(C\beta^{p-1}\right).
   \end{equation*}
\end{proof}

\section{De-Poissonization: Proof of Theorem \ref{thm:metric_approx_iid}}
\label{app:DePPP}

\begin{proof}
To utilize the results of Theorem \ref{thm:metric_approx}, we consider a coupling process of 2 random PPPs as follows. We have two infinite sequences with draws from $\rho$:
\begin{align*}
    x_1, x_2, x_3, &\ldots \\
    y_1, y_2, y_3, &\ldots
\end{align*}
Now we let $N_{-}$ be a Poisson random variable with mean $n_{-} = n-n^{\frac{1+\kappa}{2}}$, and $N_\text{inc}$ a Poisson random variable with mean $2n^{\frac{1+\kappa}{2}}$; note $N_{-}+N_\text{inc}$ is Poisson with mean $n_+ = n+n^{\frac{1+\kappa}{2}}$. This process defines two PPPs:
\begin{align*}
    H_{n_-\rho} &:= \{ x_1, \ldots, x_{N_{-}} \}, \\
    H_{n_+\rho} &:= \{ x_1, \ldots, x_{N_{-}}, y_1, \ldots, y_{N_{\text{inc}}} \}.
\end{align*}
We define the following high probability events:
\begin{align*}
    \Omega_1 &:= \{ N_{-} \leq n \}, \\ 
    \Omega_2 &:= \{ n \leq N_{-}+N_{\text{inc}} \}. 
\end{align*}
We now let $X_n$ consist of the first $N_{-} \wedge n$ points from the $x_i$ sequence, with an additional $n-N_{-}$ points taken from the $y_i$ sequence if $n>N_{-}$; $X_n$ consists of $n$ iid samples from $\rho$.
Note that on $\Omega_1$, $X_n$ gets to use all of the points available to $H_{n_-\rho}$, and possibly more; on $\Omega_2$, $H_{n_+\rho}$ gets to use all of the points available to $X_n$, and possibly more. 
For notational brevity, we let $\zeta = C_1\L^2+C_2\L^3$ where $C_1, C_2$ are the constants from Theorem \ref{thm:metric_approx} and define the following events:
\begin{align*}
    A &:= \{ n_{-}^{\frac{p-1}{m}} \ell_p^p(X_n) < \mu \L + \zeta \}, \\
    A' &:= \{ n_{-}^{\frac{p-1}{m}} \ell_p^p(H_{n_{-}\rho}) < \mu \L + \zeta \},\\
    B &:= \{ n_{+}^{\frac{p-1}{m}} \ell_p^p(X_n) > \mu \L - \zeta\}, \\
    B' &:= \{ n_{+}^{\frac{p-1}{m}} \ell_p^p(H_{n_{+}\rho}) > \mu \L - \zeta \}.
\end{align*}
We then obtain:
\begin{align*}
    P(A) &\geq P(A \cap \Omega_1) \geq P(A' \cap \Omega_1) = P(A') - P(A' \cap \Omega_1^C) \geq P(A') - P(\Omega_1^C), \\ 
    P(B) &\geq P(B \cap \Omega_2) \geq P(B' \cap \Omega_2) = P(B') - P(B' \cap \Omega_2^C) \geq P(B') - P(\Omega_2^C) \, . 
\end{align*}
By Theorem \ref{thm:metric_approx}
\begin{align*}
P(A') &\geq 1-Cn_{-}\exp\left(-c(n_{-}/2\beta)^{\kappa}\right) \geq 1-Cn\exp\left(-c\left(n/4\beta\right)^{\kappa}\right)\\
P(B') &\geq 1-Cn_{+}\exp\left(-c(n_{+}/2\beta)^{\kappa}\right) \geq 1-Cn\exp\left(-c(n/2\beta)^{\kappa}\right)
\end{align*}
since $n\geq 5$ and $2(n\beta/2)^{-\frac{1}{m}\left(\frac{1}{3} - \epsilon\right)}\leq d(x,y) \leq C_{\M,\rho}$.

We now bound $P(\Omega_1^C), P(\Omega_2^C)$. Note that when $X=\text{Poisson}(\lambda)$, the following Poisson tail bounds can be derived from a Chernoff bound argument:
\begin{align*}
    P(X \geq x) \leq \left(\frac{\lambda}{x}\right)e^{x-\lambda} \text{ for } x > \lambda \quad,\quad P(X \leq x) \leq \left(\frac{\lambda}{x}\right)e^{x-\lambda} \text{ for } x < \lambda \, .
\end{align*}
Combining the first inequality with $\ln(1+x)\leq \frac{2x}{2+x}$ for $-1<x\leq 0$ and $\frac{1}{1-x} \geq 1+x$ for $|x|\leq 1$ gives $P(N_{-} \geq n) \leq e^{-\frac{1}{2}n^{\kappa}}$. Combining the second inequality with $\ln(1+x)\leq \frac{x}{\sqrt{1+x}}$ for $x\geq 0$ and $\frac{1}{\sqrt{1+x}} \leq 1-\frac{x}{4}$ for $0\leq x\leq 1$ gives $P(N_{-}+N_{\text{inc}} \leq n)\leq e^{-\frac{1}{4}n^{\kappa}}$. We thus obtain:
\begin{align*}
    P(\Omega_1^C) &\leq e^{-\frac{1}{2}n^{\kappa}} \quad, \quad P(\Omega_2^C) \leq e^{-\frac{1}{4}n^{\kappa}}
\end{align*}
so that
\begin{align*}
   P(A \cap B) &\geq P(A) - P(B^C) \geq 1-Cn\exp\left(-c\left(n/4\beta\right)^{\kappa}\right) \, .
\end{align*}
Note that on $A \cap B$, we have:
\begin{align*}
     n^{\frac{p-1}{m}} \ell_p^p(X_n)  &< \left(\frac{1}{1-n^{\frac{\kappa-1}{2}}}\right)^{\frac{p-1}{m}}\left(\mu \L + \zeta\right) \leq \left(1+2n^{\frac{\kappa-1}{2}}\right)^{\frac{p-1}{m}}\left(\mu \L + \zeta\right)
\end{align*}
since $n^{\frac{\kappa-1}{2}} \leq \frac{1}{2}$ (true since $\kappa \leq \frac{1}{21}, n\geq 5$). We utilize the inequality $(1+x)^q \leq 1+ x(1 \vee \lceil q \rceil^2)$ for $0 \leq x \leq \lceil q \rceil ^{-1}$. If $q \leq 1$, then $(1+x)^q \leq 1 + x$ and the inequality clearly holds. If $q > 1$, then by the Binomial Theorem
\begin{align*}
    (1+x)^{\lceil q \rceil} = \sum_{k=0}^{\lceil q \rceil} {\lceil q \rceil \choose k} x^k = 1 + \sum_{k=1}^{\lceil q \rceil} {\lceil q \rceil \choose k} x^k \leq 1 + \lceil q \rceil^2 x
\end{align*}
as long as ${\lceil q \rceil \choose k} x^k \geq {\lceil q \rceil \choose k+1} x^{k+1}$ for $1 \leq k\leq \lceil q \rceil -1$, which is guaranteed by $x \leq \lceil q \rceil ^{-1}$. We thus obtain that for $2n^{\frac{\kappa-1}{2}} \leq \left\lceil\frac{p-1}{m}\right\rceil^{-1}$:
\begin{align*}
    n^{\frac{p-1}{m}} \ell_p^p(X_n) &\leq \left[1+2n^{\frac{\kappa-1}{2}}\left(1 \vee \left\lceil\frac{p-1}{m}\right\rceil^2\right)\right]\left(\mu\L + \zeta\right) \\
    &\leq \mu\L + \zeta + 2n^{\frac{\kappa-1}{2}}\left(1 \vee \left\lceil\frac{p-1}{m}\right\rceil^2\right)3\mu \L
\end{align*}
since $\mu\L+\zeta \leq 3\mu\L$ for $d(x,y)\leq C_{\M,\rho}$. A similar argument gives
\begin{align*}
 n^{\frac{p-1}{m}} \ell_p^p(X_n) &\geq \mu\L - \zeta - n^{\frac{\kappa-1}{2}}\left(1 \vee \left\lceil\frac{p-1}{m}\right\rceil^2\right)3\mu \L   
\end{align*}
so that on $A \cap B$ we have
\begin{align*}
|n^{\frac{p-1}{m}} \ell_p^p(X_n) - \mu\L | &\leq \zeta + 2n^{\frac{\kappa-1}{2}}\left(1 \vee \left\lceil\frac{p-1}{m}\right\rceil^2\right)3\mu \L \, .
\end{align*}
In particular, 
\begin{align*}
|n^{\frac{p-1}{m}} \ell_p^p(X_n) - \mu\L | &\leq C_1\L + C_2(\L^3 + \L n^{\frac{\kappa-1}{2}}) \, ,
\end{align*}
where we update $C_{p,d,\beta}$ in the definition of $C_2$ as needed.

Finally, we observe that the lower bound on $d(x,y)$ implies $\L^2 \geq C_{p,d,\beta}n^{-\frac{2}{m}\left(\frac{1}{3}-\epsilon\right)}$. Since $n^{-\frac{2}{m}\left(\frac{1}{3}-\epsilon\right)}$ dominates $n^{\frac{\kappa-1}{2}}$ whenever $3m-4+10\epsilon \geq 0$ which is guaranteed by $m\geq 2$, we obtain the theorem statement. 
\end{proof}

\section{Additional Experimental Plots}
\label{sec:Additional Plots}


\begin{figure}[htbp!]
	\centering
\begin{subfigure}[t]{0.32\textwidth}
		\centering
		\includegraphics[width=\textwidth]{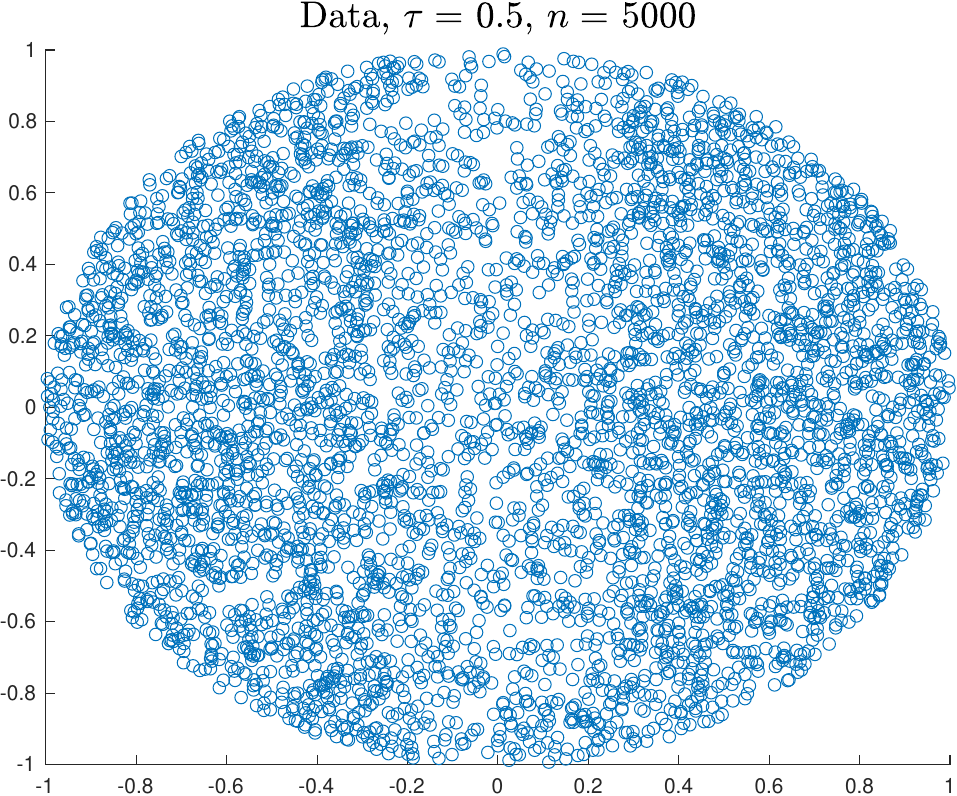}
\end{subfigure}
\begin{subfigure}[t]{0.32\textwidth}
		\centering
		\includegraphics[width=\textwidth]{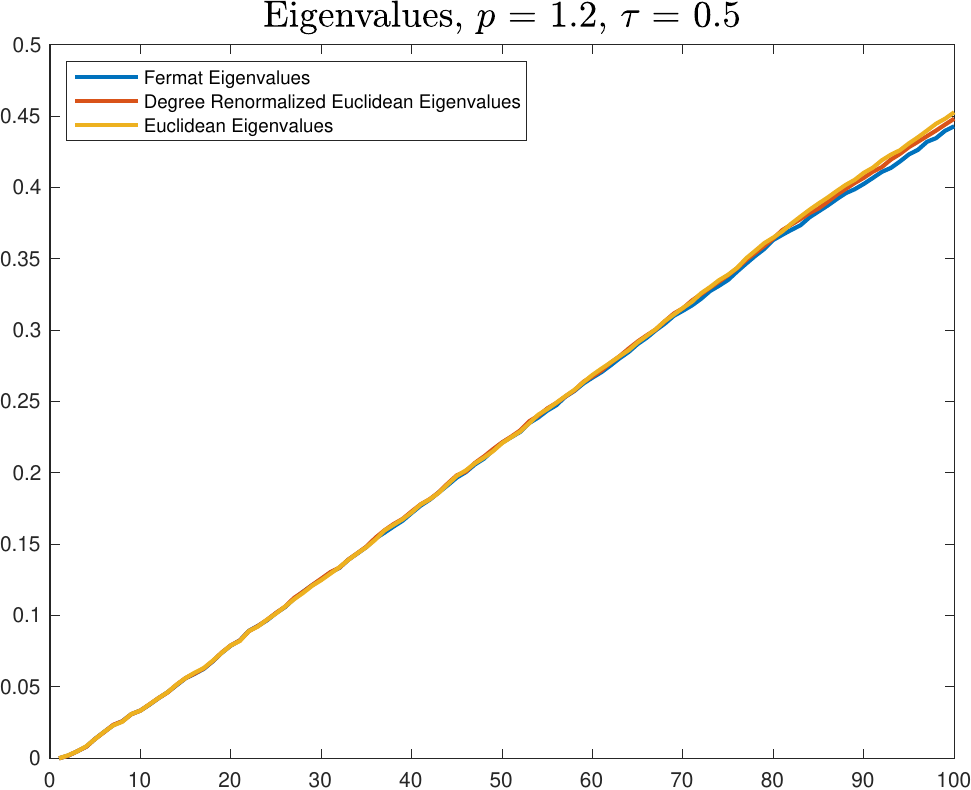}
\end{subfigure}
\begin{subfigure}[t]{0.32\textwidth}
		\centering
		\includegraphics[width=\textwidth]{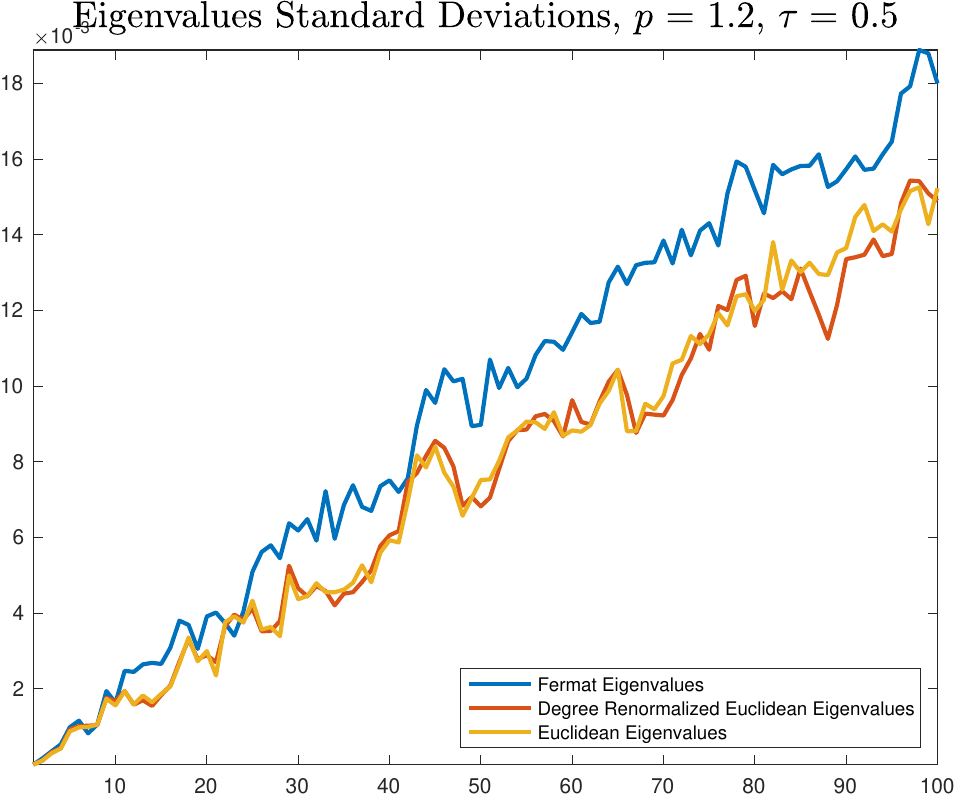}
\end{subfigure}

\begin{subfigure}[t]{0.32\textwidth}
		\centering
		\includegraphics[width=\textwidth]{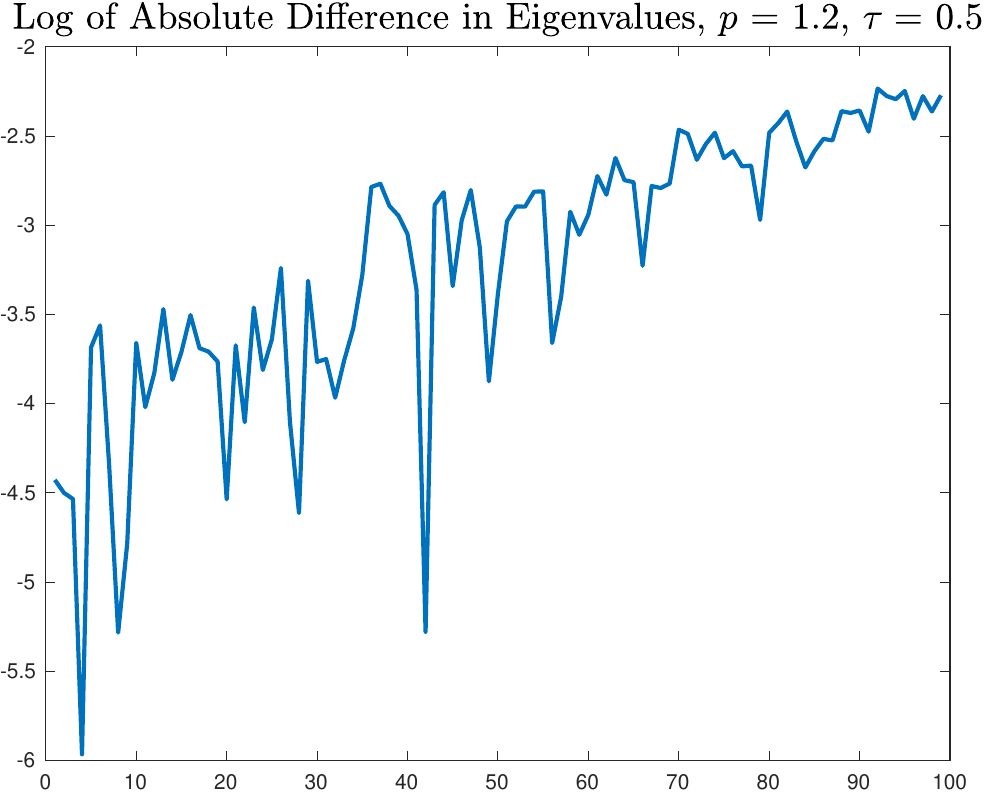}
\end{subfigure}
  \begin{subfigure}[t]{0.32\textwidth}
		\centering
		\includegraphics[width=\textwidth]{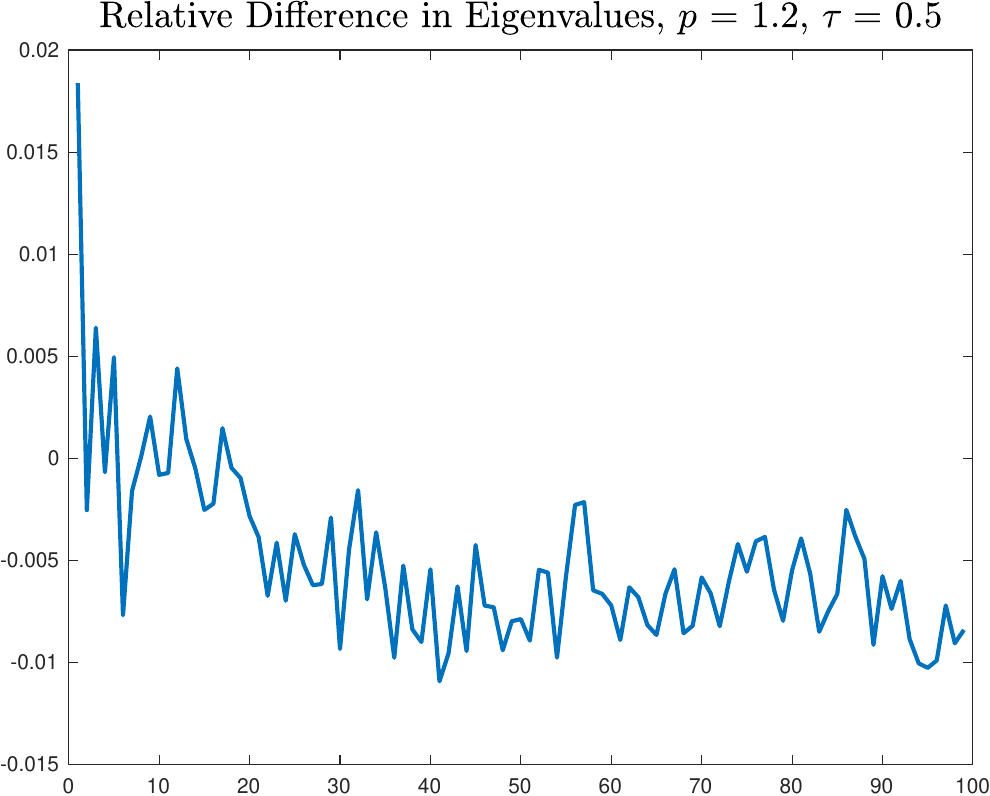}
\end{subfigure}
\begin{subfigure}[t]{0.32\textwidth}
		\centering
		\includegraphics[width=\textwidth]{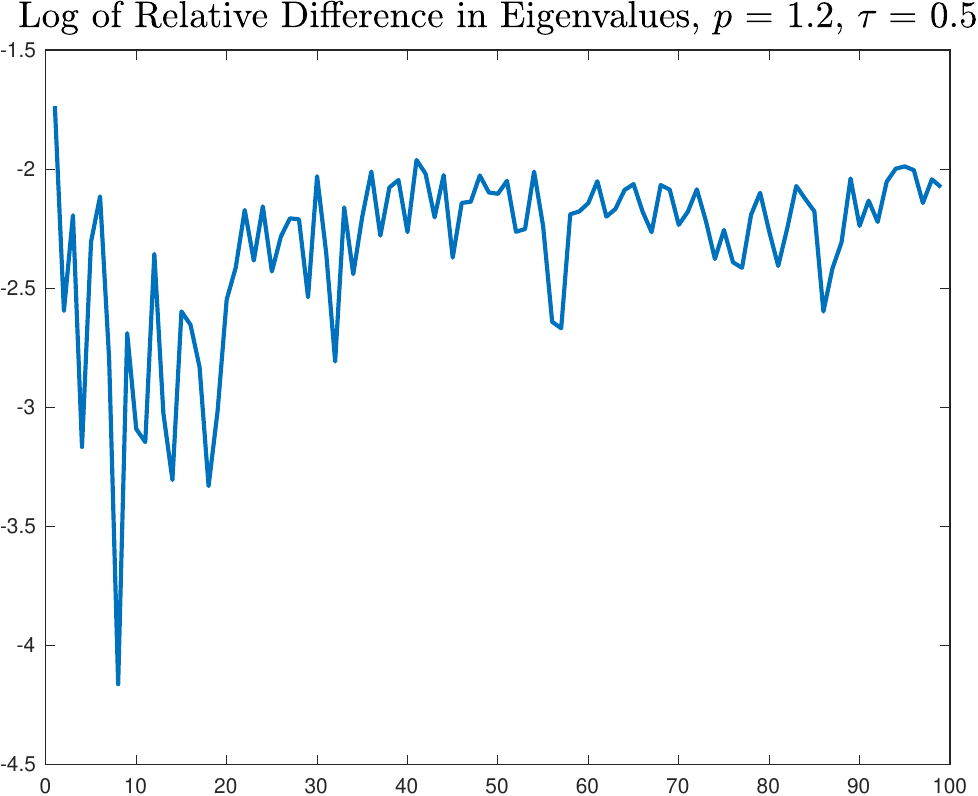}
\end{subfigure}
	\caption{$p=1.2$, $\tau=.5$.  Runtime for Fermat Laplacian: $ 168.46\pm 10.70$s.  Runtime for Rescaled Euclidean Laplacian: $7.38\pm1.10$s.}
	\label{fig:BallGap_p=1.2_thresh=.5}
\end{figure}


\begin{figure}[htbp!]
	\centering
\begin{subfigure}[t]{0.32\textwidth}
		\centering
		\includegraphics[width=\textwidth]{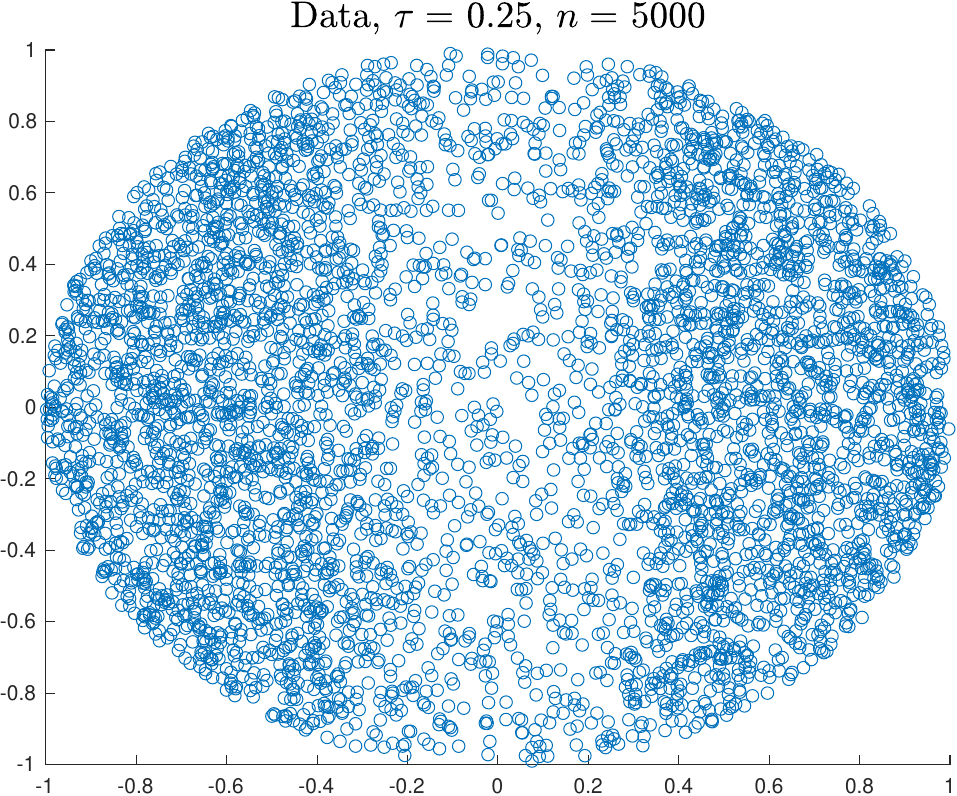}
\end{subfigure}
\begin{subfigure}[t]{0.32\textwidth}
		\centering
		\includegraphics[width=\textwidth]{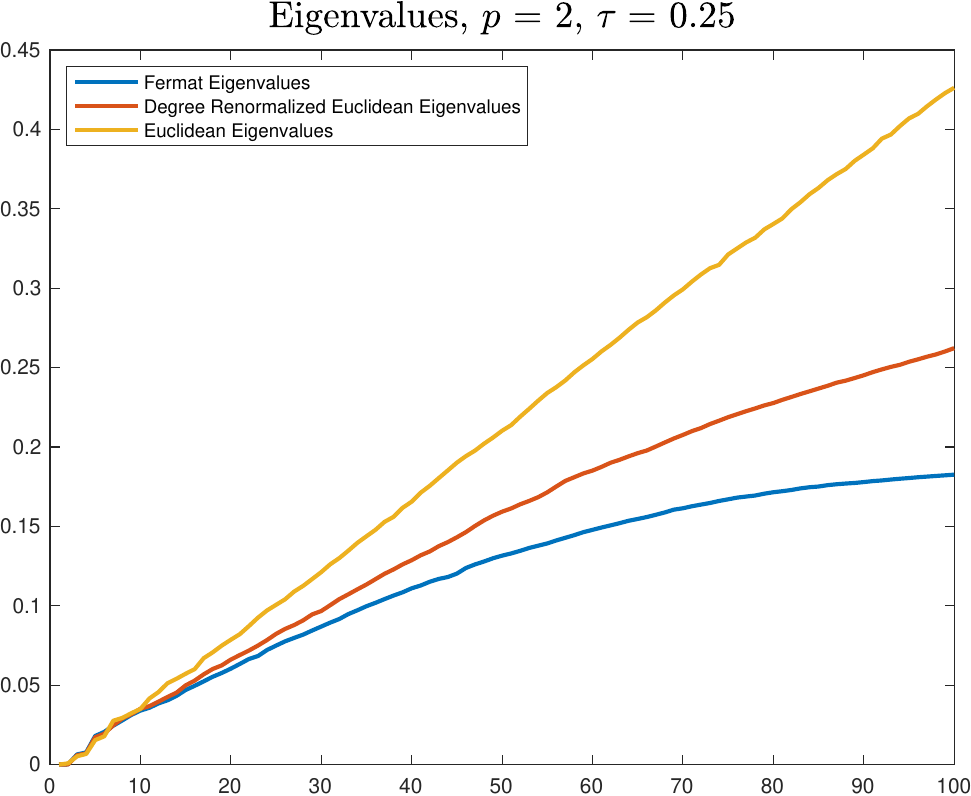}
\end{subfigure}
\begin{subfigure}[t]{0.32\textwidth}
		\centering
		\includegraphics[width=\textwidth]{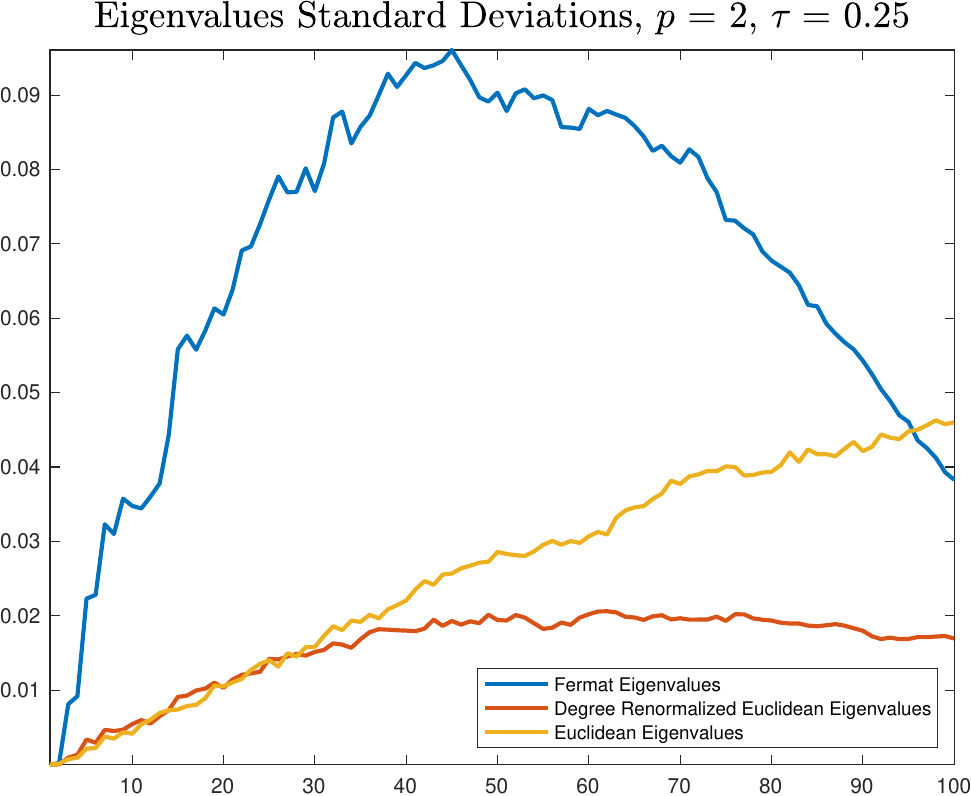}
\end{subfigure}
\begin{subfigure}[t]{0.32\textwidth}
		\centering
		\includegraphics[width=\textwidth]{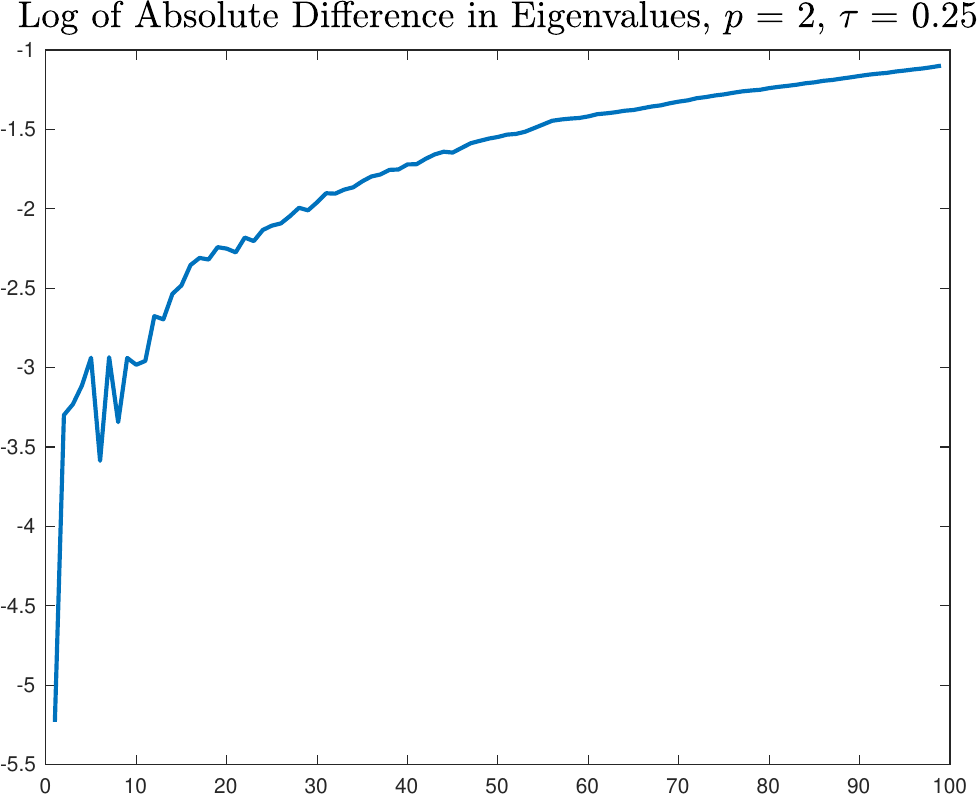}
\end{subfigure}
  \begin{subfigure}[t]{0.32\textwidth}
		\centering
		\includegraphics[width=\textwidth]{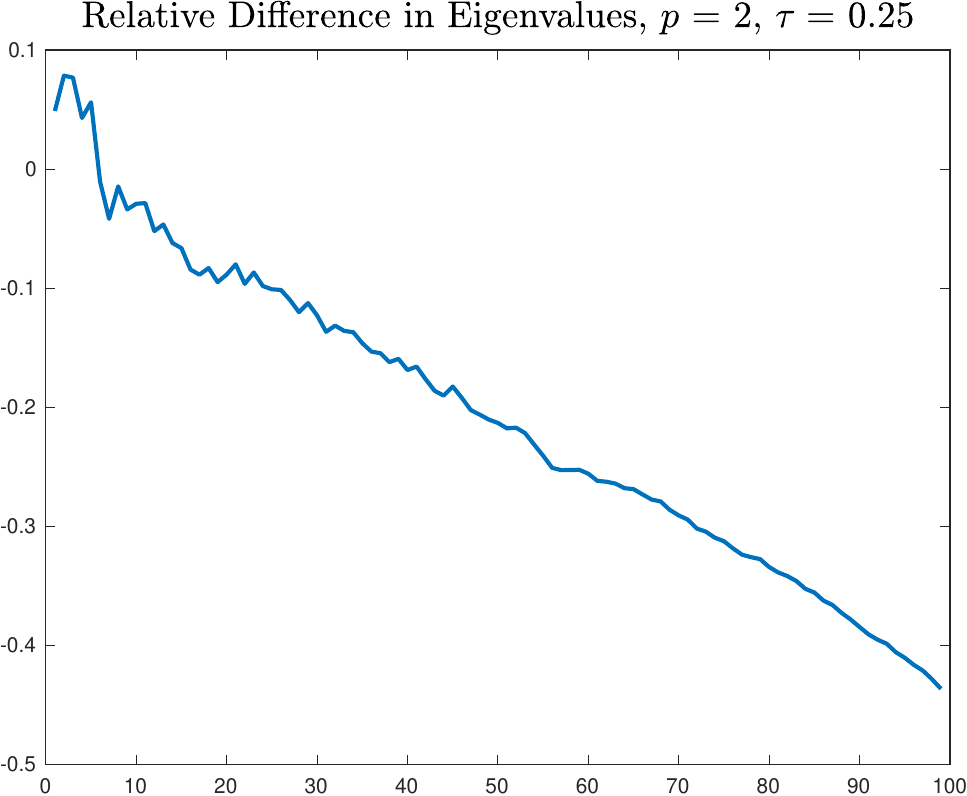}
\end{subfigure}
\begin{subfigure}[t]{0.32\textwidth}
		\centering
		\includegraphics[width=\textwidth]{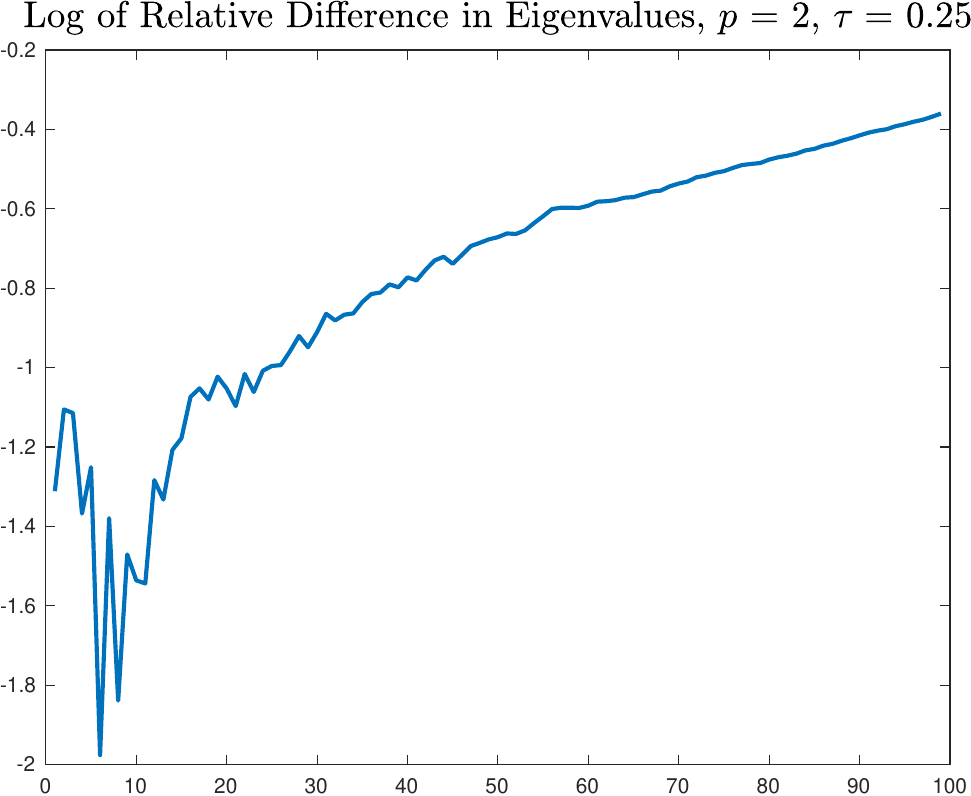}
\end{subfigure}
	\caption{$p=2$, $\tau=.1$.  Runtime for Fermat Laplacian: $212.52\pm 70.49$s.  Runtime for Rescaled Euclidean Laplacian: $1.33\pm.24$s.}
	\label{fig:BallGap_p=2_thresh=.25}
\end{figure}


\begin{figure}[h]
	\centering
\begin{subfigure}[t]{0.32\textwidth}
		\centering
		\includegraphics[width=\textwidth]{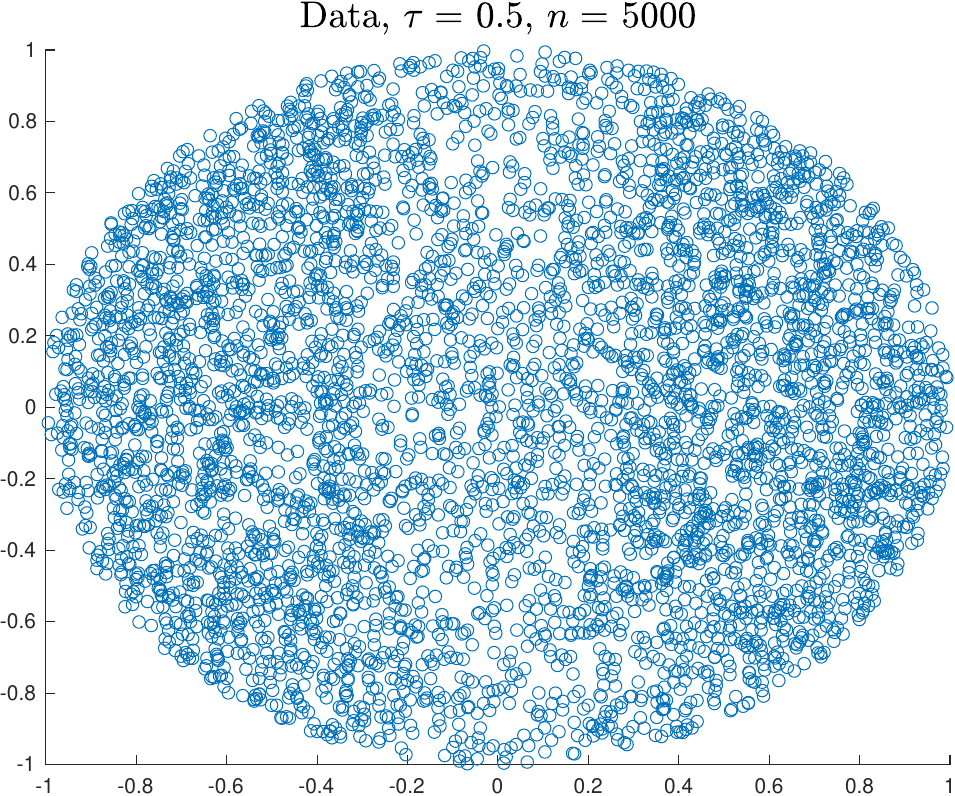}
\end{subfigure}
\begin{subfigure}[t]{0.32\textwidth}
		\centering
		\includegraphics[width=\textwidth]{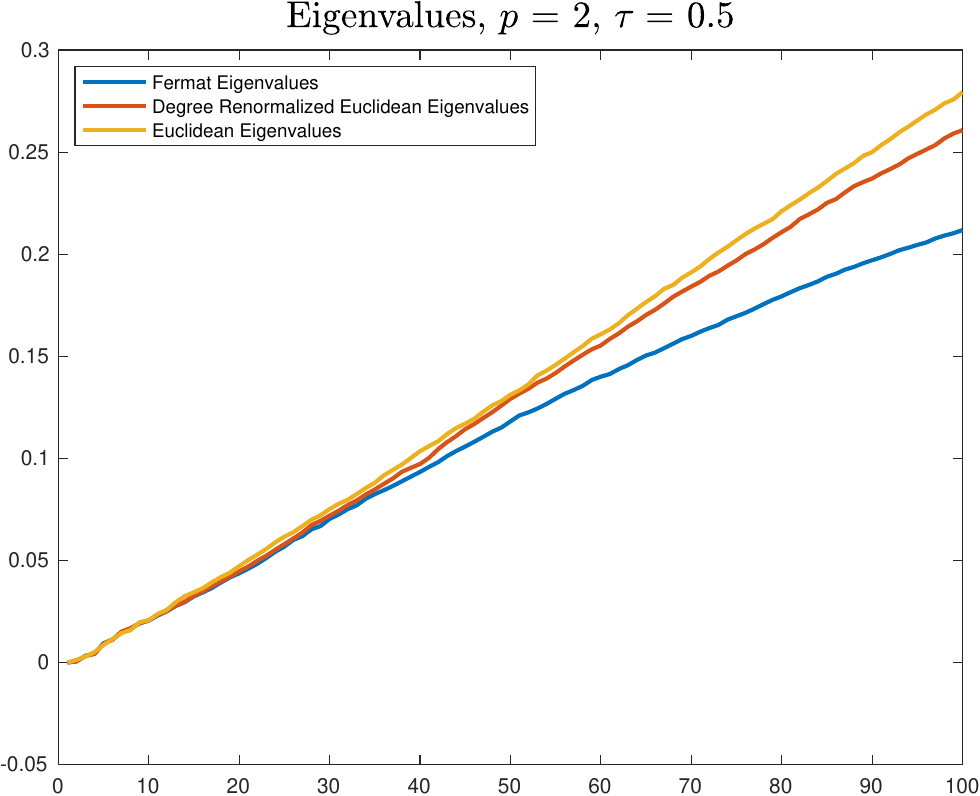}
\end{subfigure}
\begin{subfigure}[t]{0.32\textwidth}
		\centering
		\includegraphics[width=\textwidth]{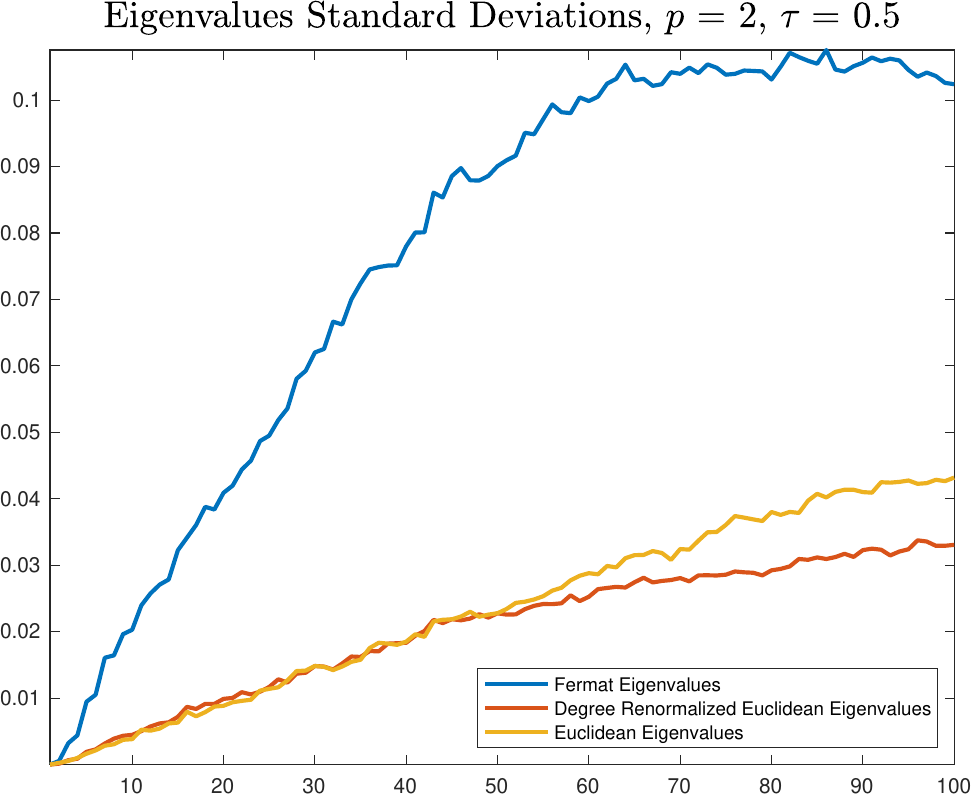}
\end{subfigure}
\begin{subfigure}[t]{0.32\textwidth}
		\centering
		\includegraphics[width=\textwidth]{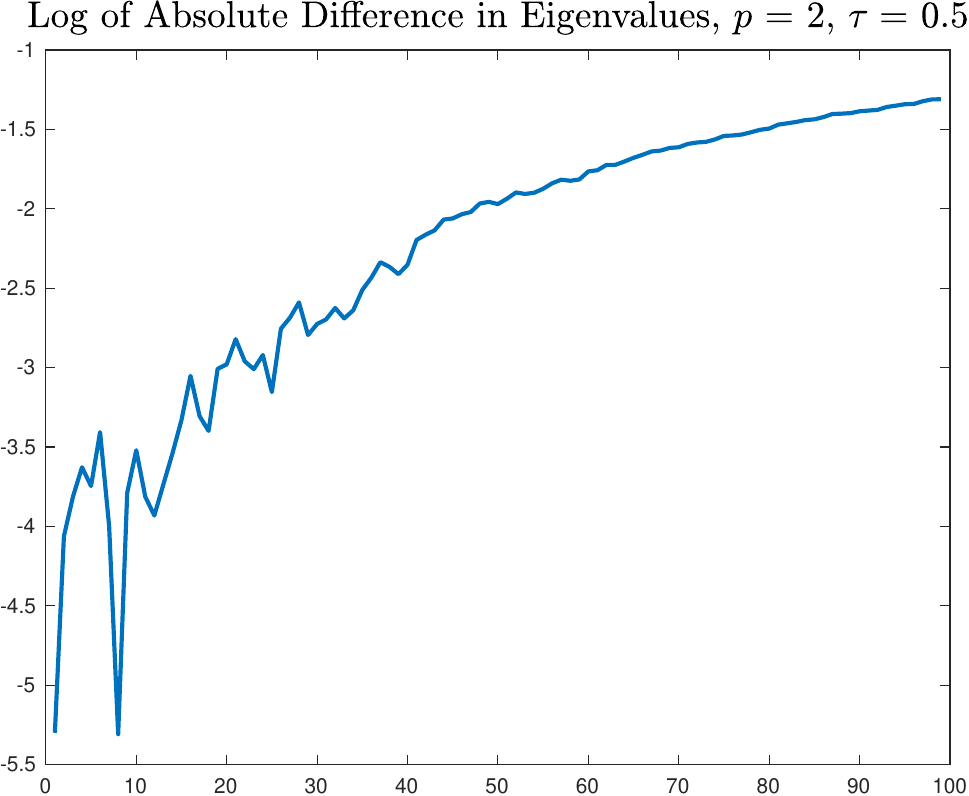}
\end{subfigure}
  \begin{subfigure}[t]{0.32\textwidth}
		\centering
		\includegraphics[width=\textwidth]{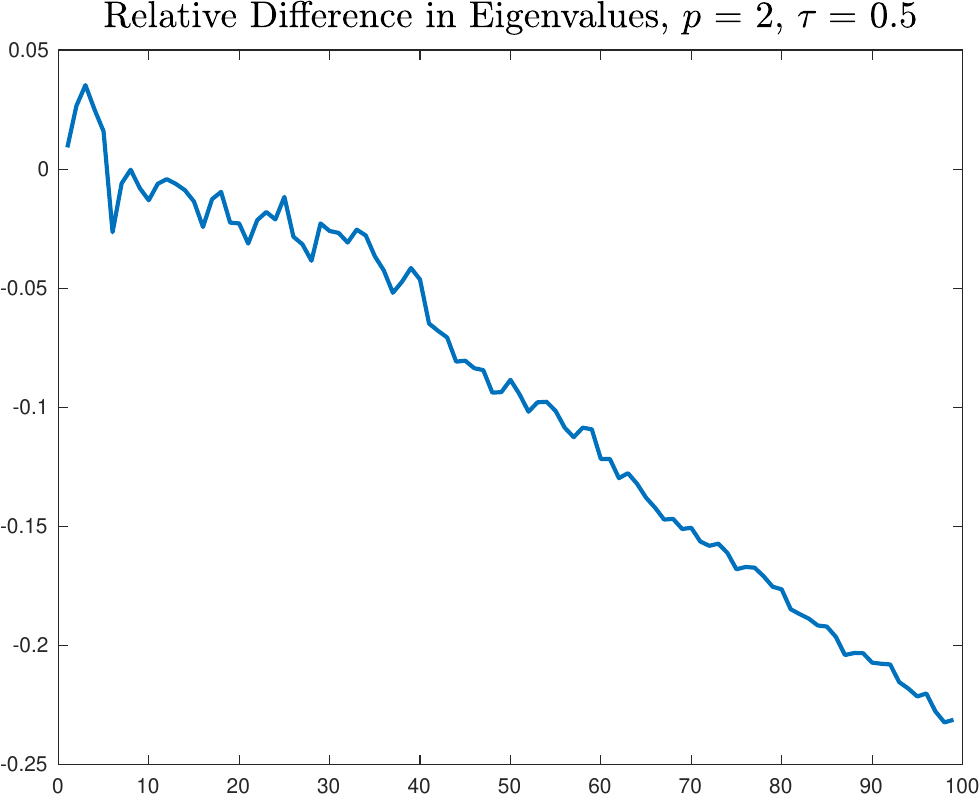}
\end{subfigure}
\begin{subfigure}[t]{0.32\textwidth}
		\centering
		\includegraphics[width=\textwidth]{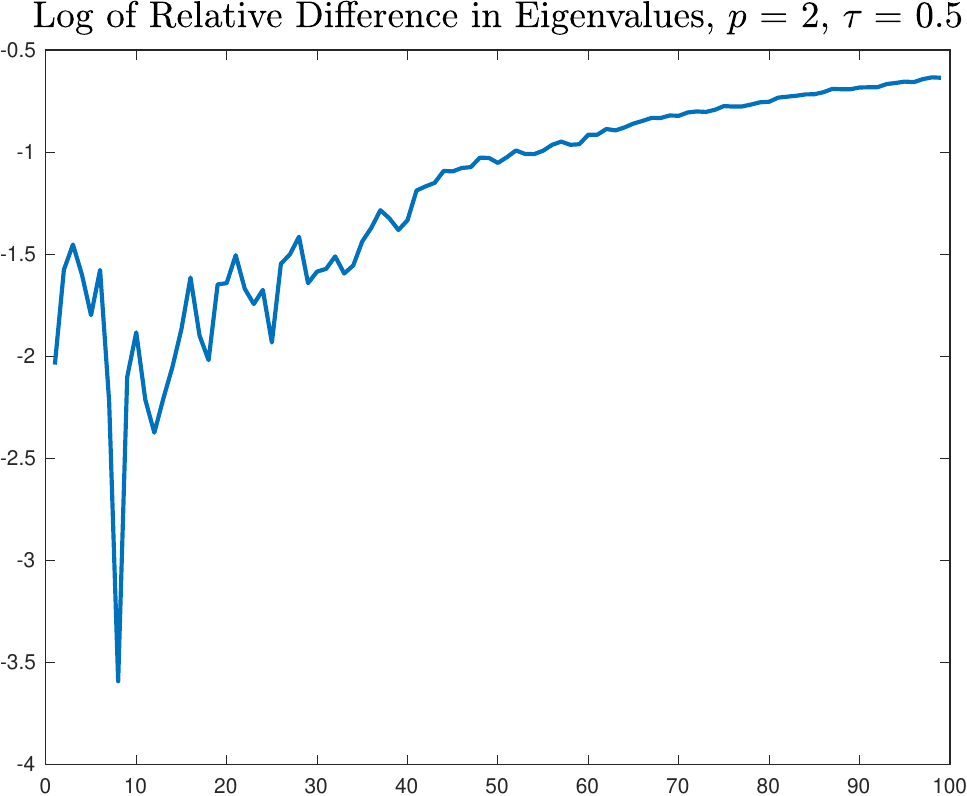}
\end{subfigure}
	\caption{$p=2$, $\tau=.5$.  Runtime for Fermat Laplacian: $83.62\pm10.50$s.  Runtime for Rescaled Euclidean Laplacian: $1.24\pm .16$s.}
	\label{fig:BallGap_p=2_thresh=.5}
\end{figure}


\begin{figure}[t]
	\centering
\begin{subfigure}[t]{0.32\textwidth}
		\centering
		\includegraphics[width=\textwidth]{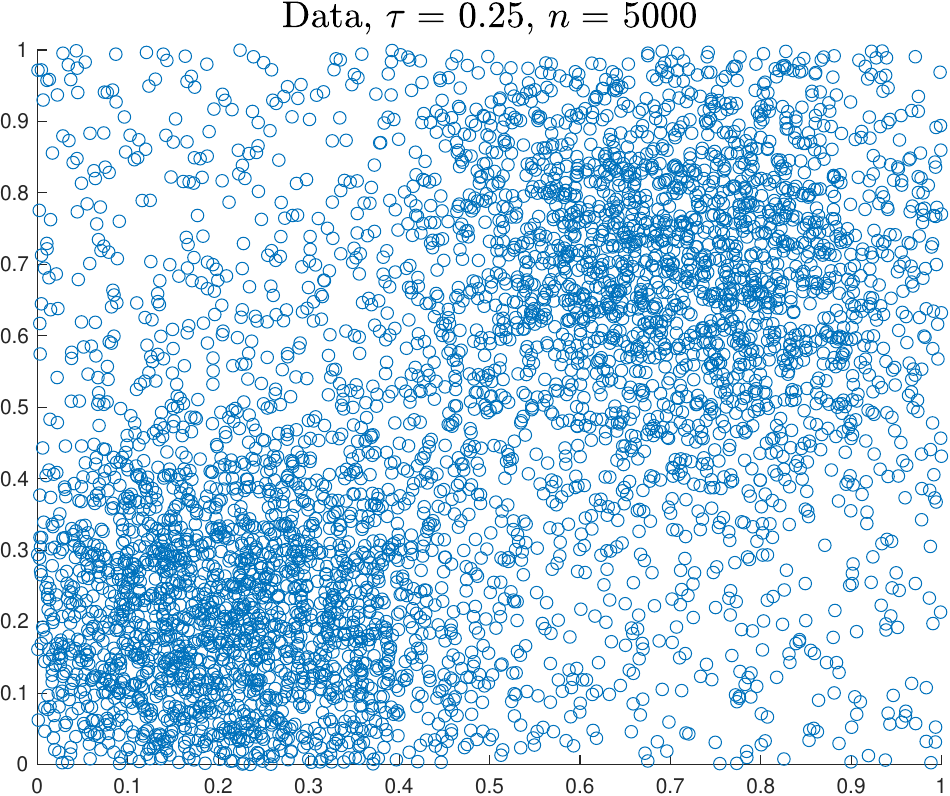}
\end{subfigure}
\begin{subfigure}[t]{0.32\textwidth}
		\centering
		\includegraphics[width=\textwidth]{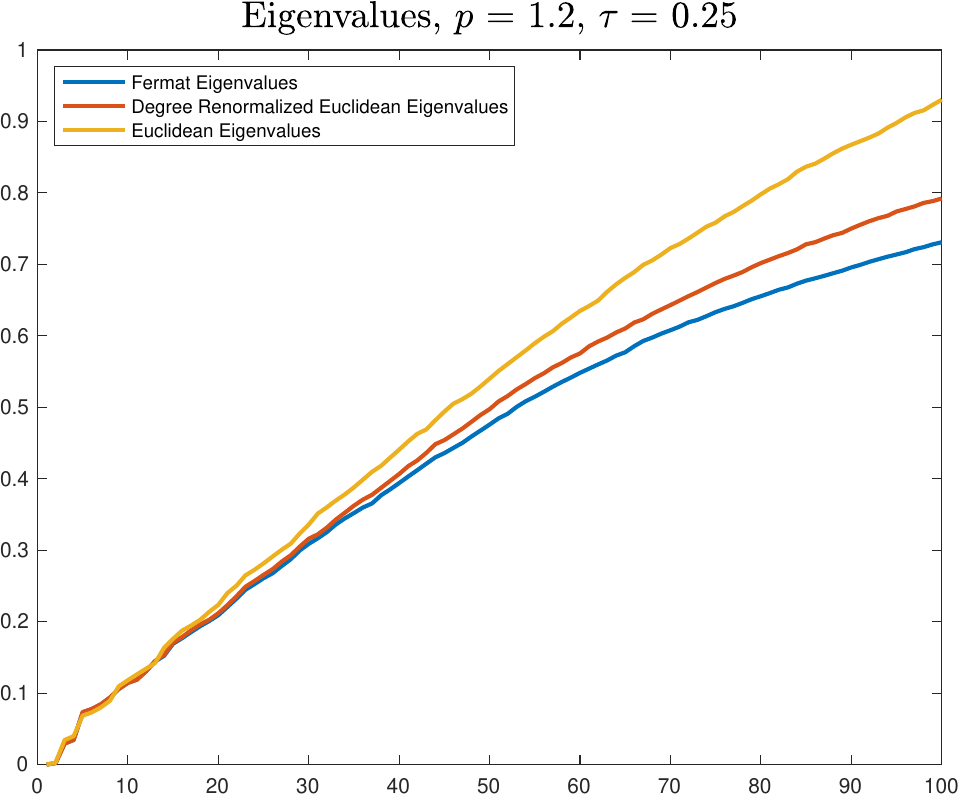}
\end{subfigure}
\begin{subfigure}[t]{0.32\textwidth}
		\centering
		\includegraphics[width=\textwidth]{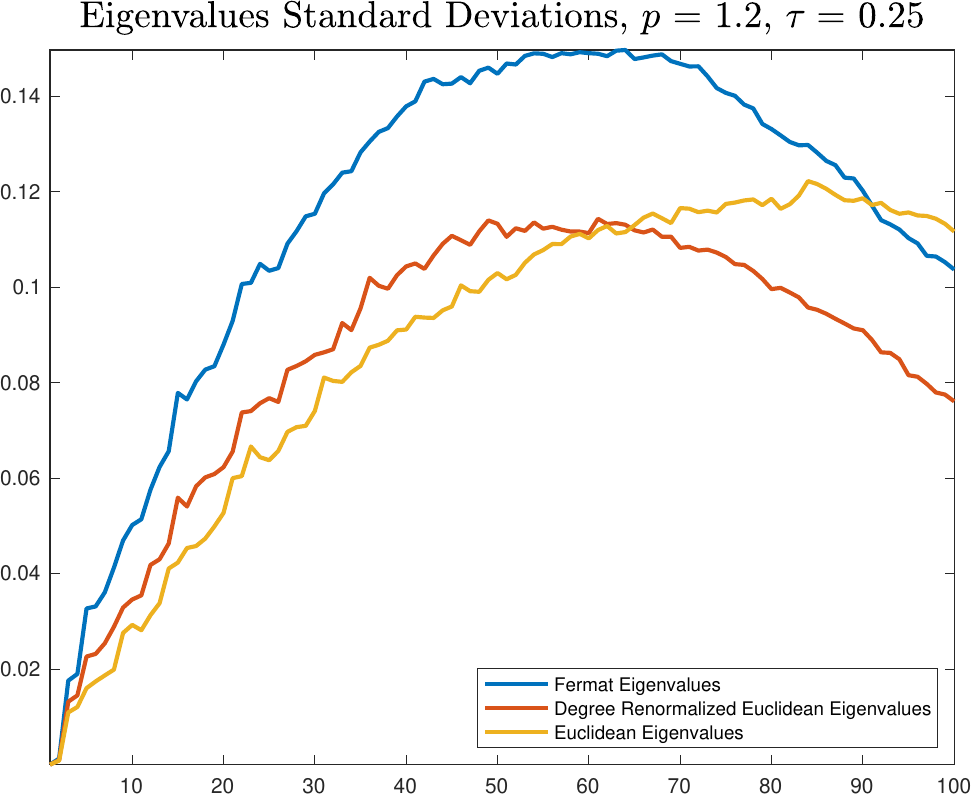}
\end{subfigure}

\begin{subfigure}[t]{0.32\textwidth}
		\centering
		\includegraphics[width=\textwidth]{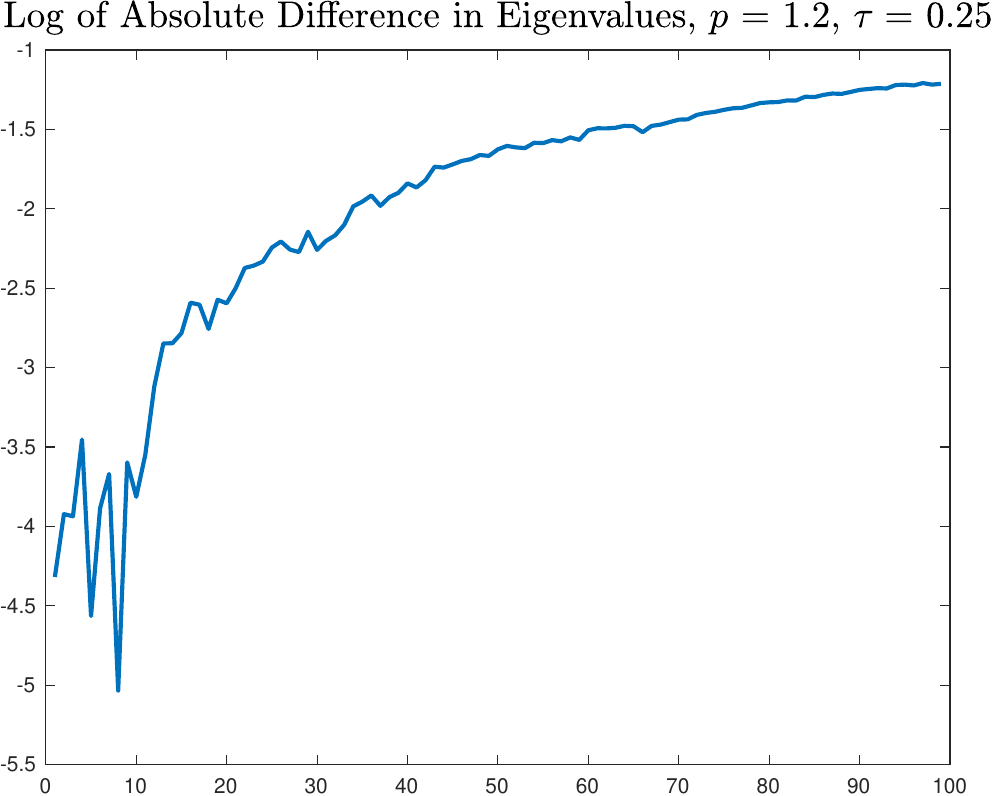}
\end{subfigure}
\begin{subfigure}[t]{0.32\textwidth}
		\centering
		\includegraphics[width=\textwidth]{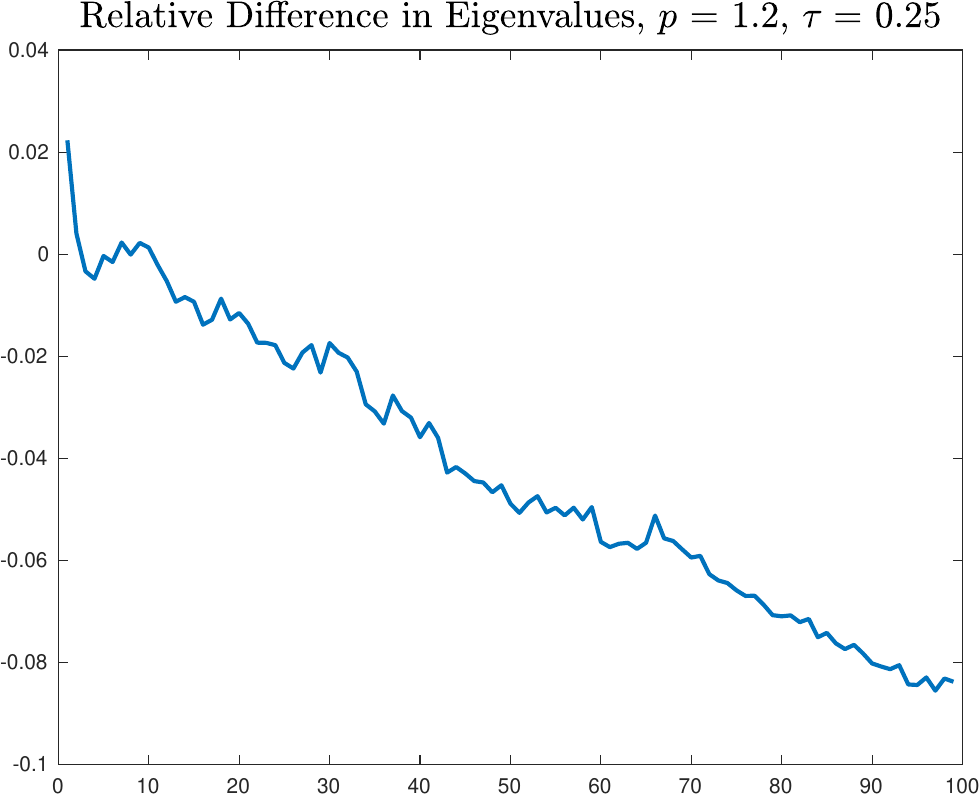}
\end{subfigure}
\begin{subfigure}[t]{0.32\textwidth}
		\centering
		\includegraphics[width=\textwidth]{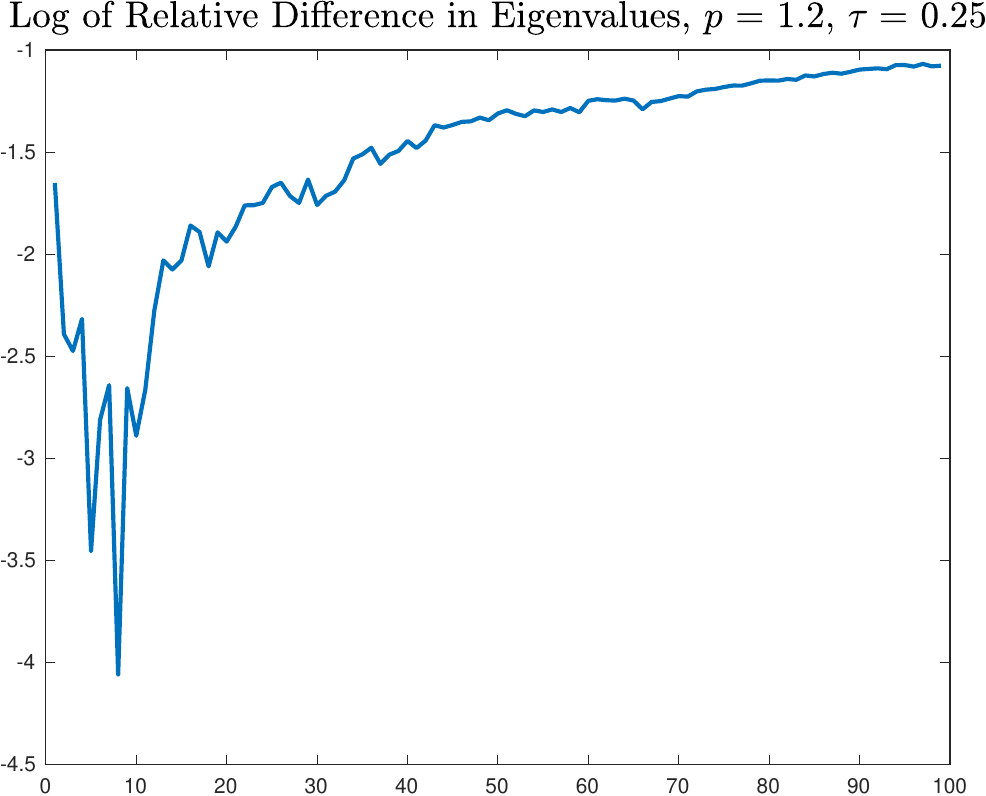}
\end{subfigure}
	\caption{$p=1.2$, $\tau=.25$.  Runtime for Fermat Laplacian: $170.10\pm 58.73$s.  Runtime for Rescaled Euclidean Laplacian: $1.27\pm.05$s.}
	\label{fig:GaussianMixture_p=1.2_thresh=.1}
\end{figure}


\begin{figure}[h]
	\centering
\begin{subfigure}[t]{0.32\textwidth}
		\centering
		\includegraphics[width=\textwidth]{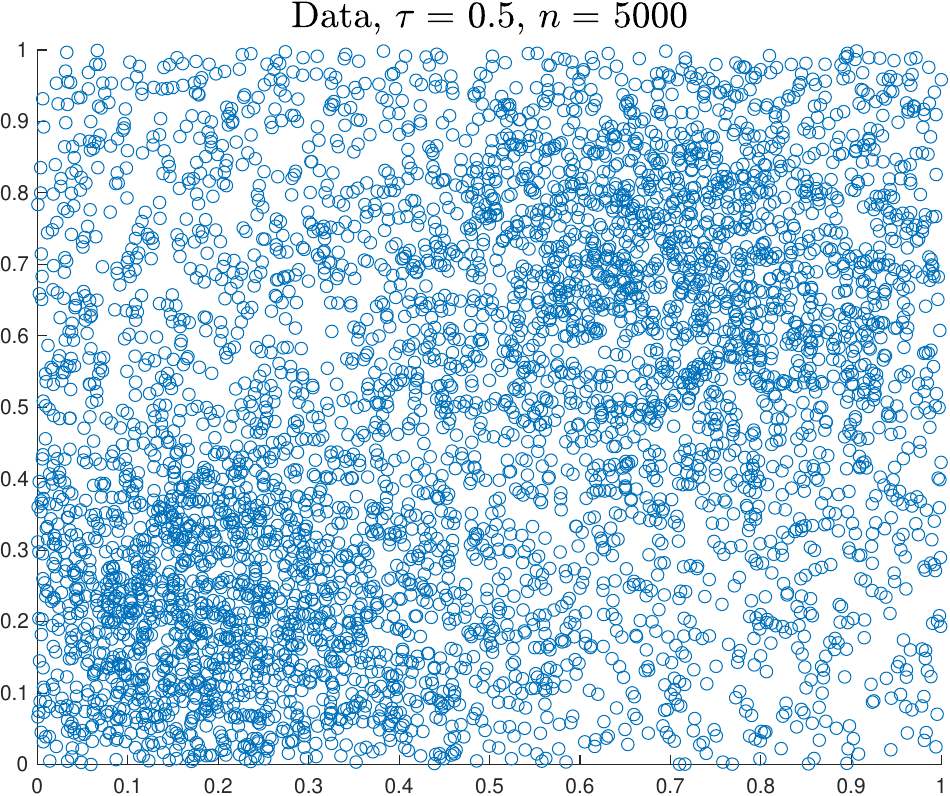}
\end{subfigure}
\begin{subfigure}[t]{0.32\textwidth}
		\centering
		\includegraphics[width=\textwidth]{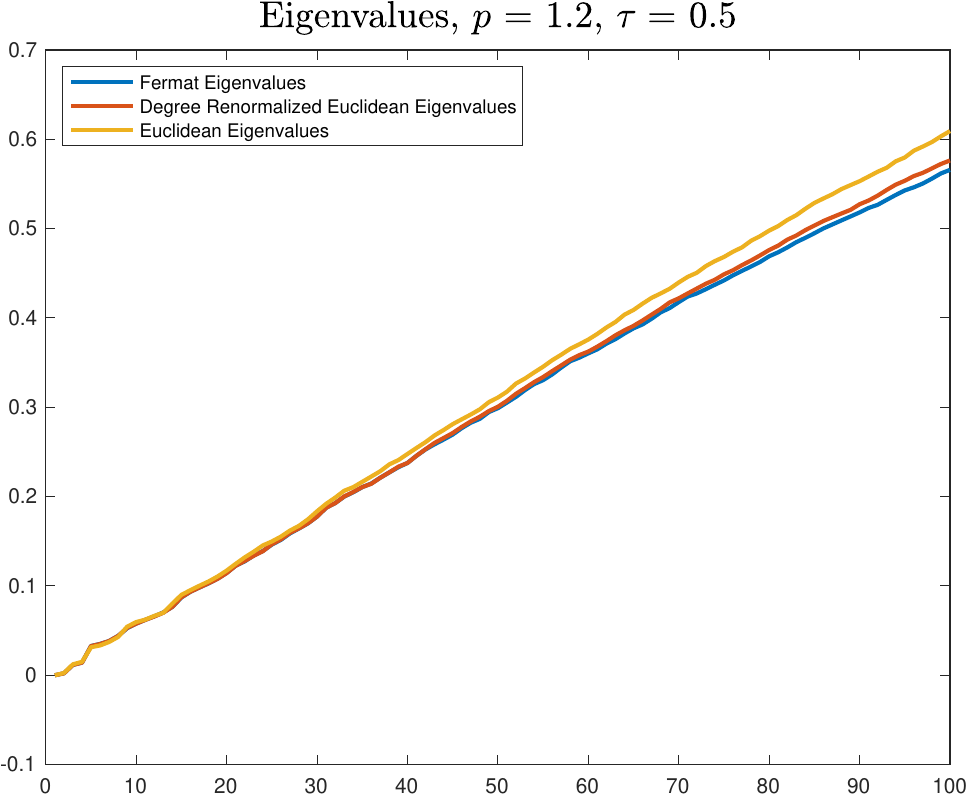}
\end{subfigure}
\begin{subfigure}[t]{0.32\textwidth}
		\centering
		\includegraphics[width=\textwidth]{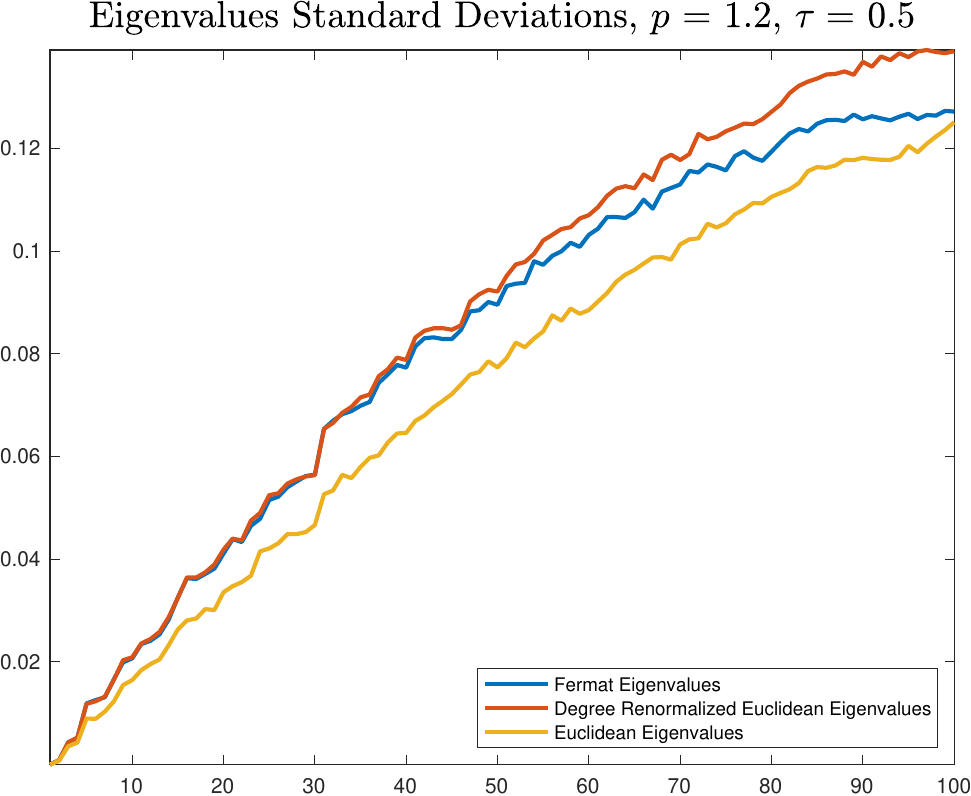}
\end{subfigure}
\begin{subfigure}[t]{0.32\textwidth}
		\centering
		\includegraphics[width=\textwidth]{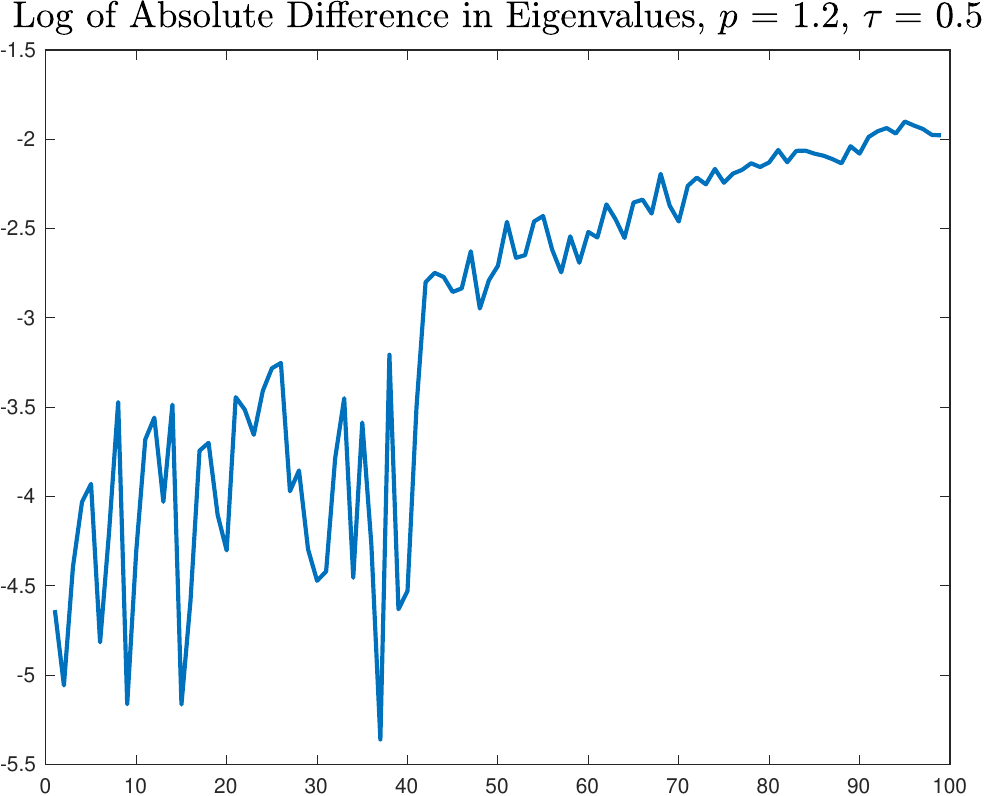}
\end{subfigure}
  \begin{subfigure}[t]{0.32\textwidth}
		\centering
		\includegraphics[width=\textwidth]{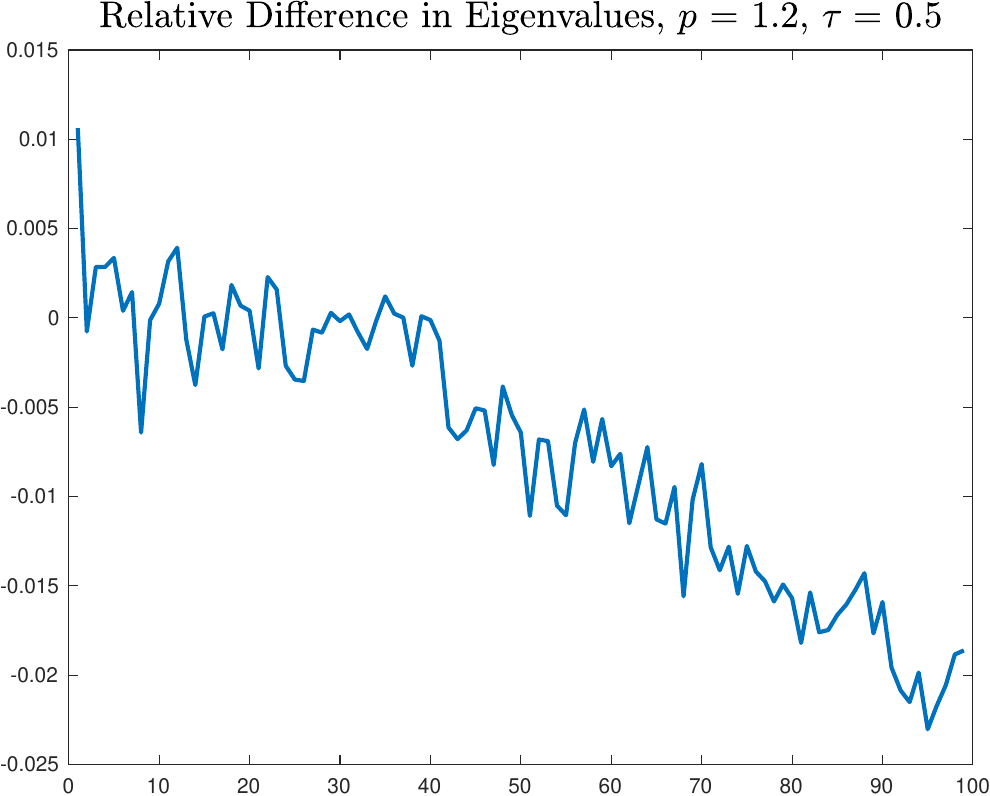}
\end{subfigure}
\begin{subfigure}[t]{0.32\textwidth}
		\centering
		\includegraphics[width=\textwidth]{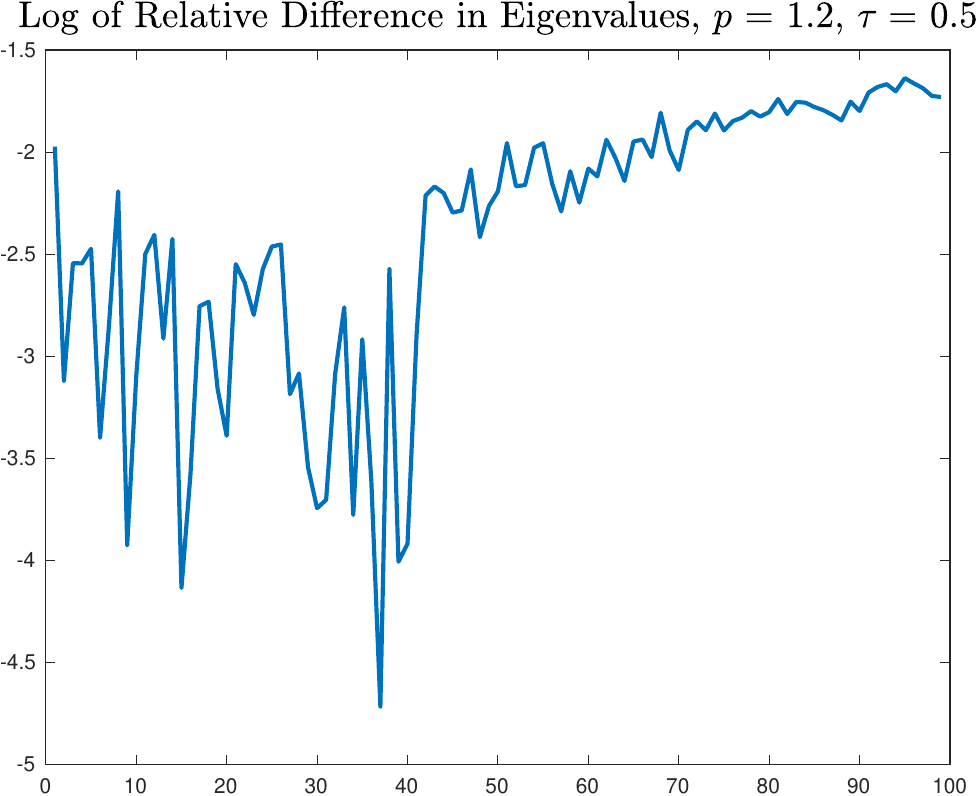}
\end{subfigure}
	\caption{$p=1.2$, $\tau=.5$.  Runtime for Fermat Laplacian: $58.72\pm1.51$s.  Runtime for Rescaled Euclidean Laplacian: $1.04\pm.05$s.}
	\label{fig:GaussianMixture_p=1.2_thresh=.5}
\end{figure}


\begin{figure}[h]
	\centering
\begin{subfigure}[t]{0.32\textwidth}
		\centering
		\includegraphics[width=\textwidth]{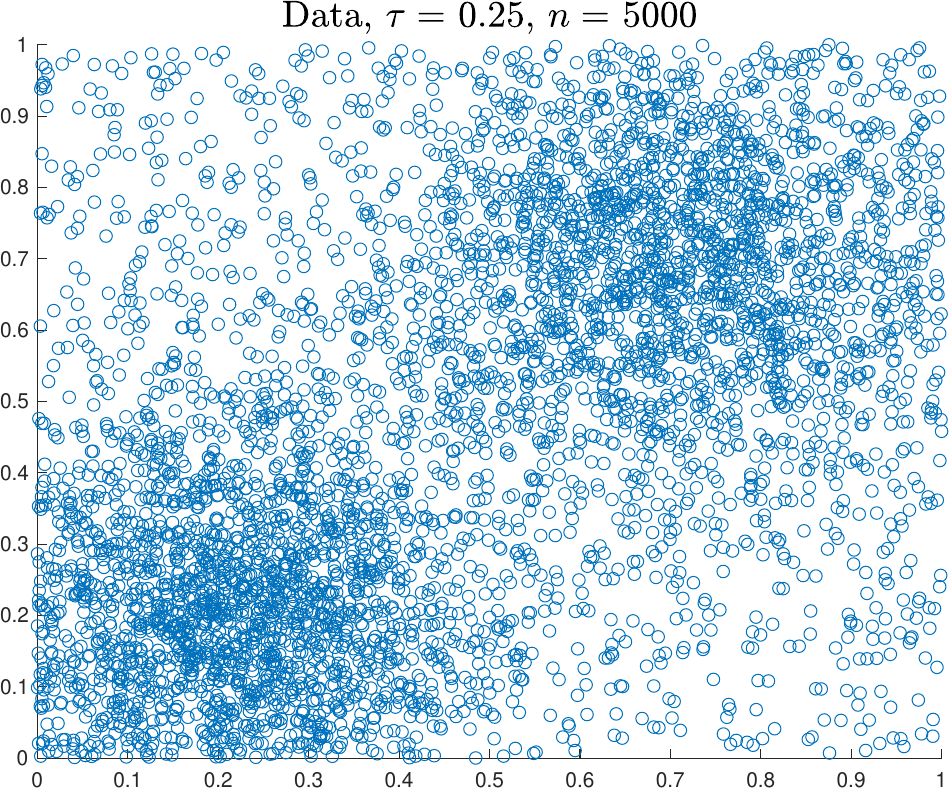}
\end{subfigure}
\begin{subfigure}[t]{0.32\textwidth}
		\centering
		\includegraphics[width=\textwidth]{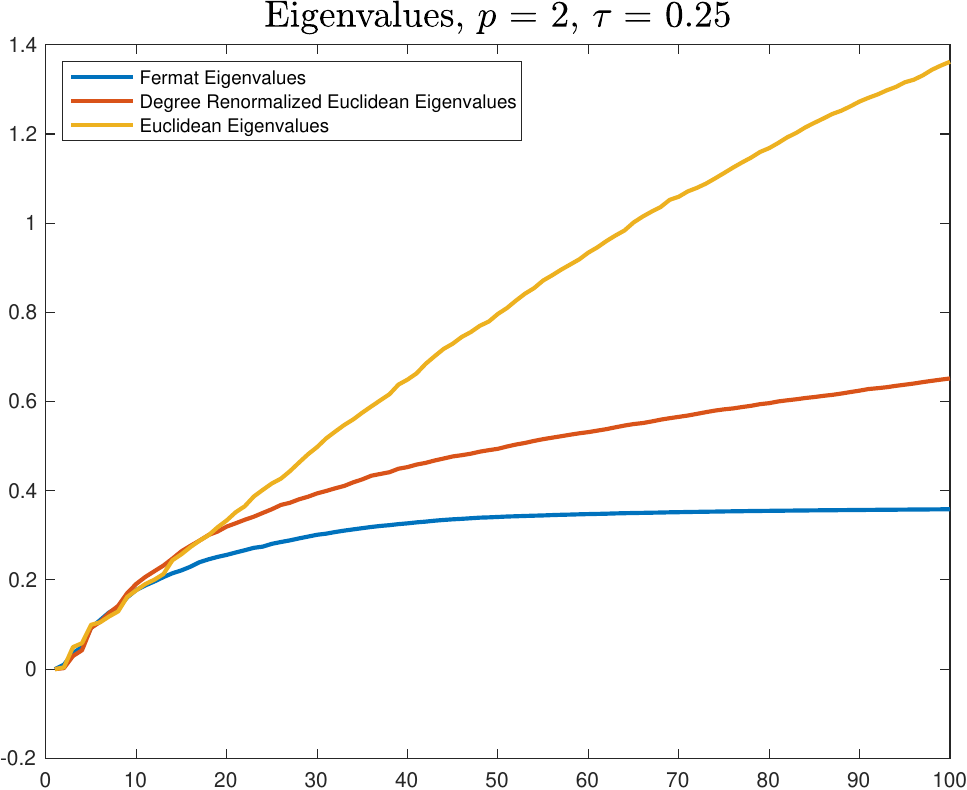}
\end{subfigure}
\begin{subfigure}[t]{0.32\textwidth}
		\centering
		\includegraphics[width=\textwidth]{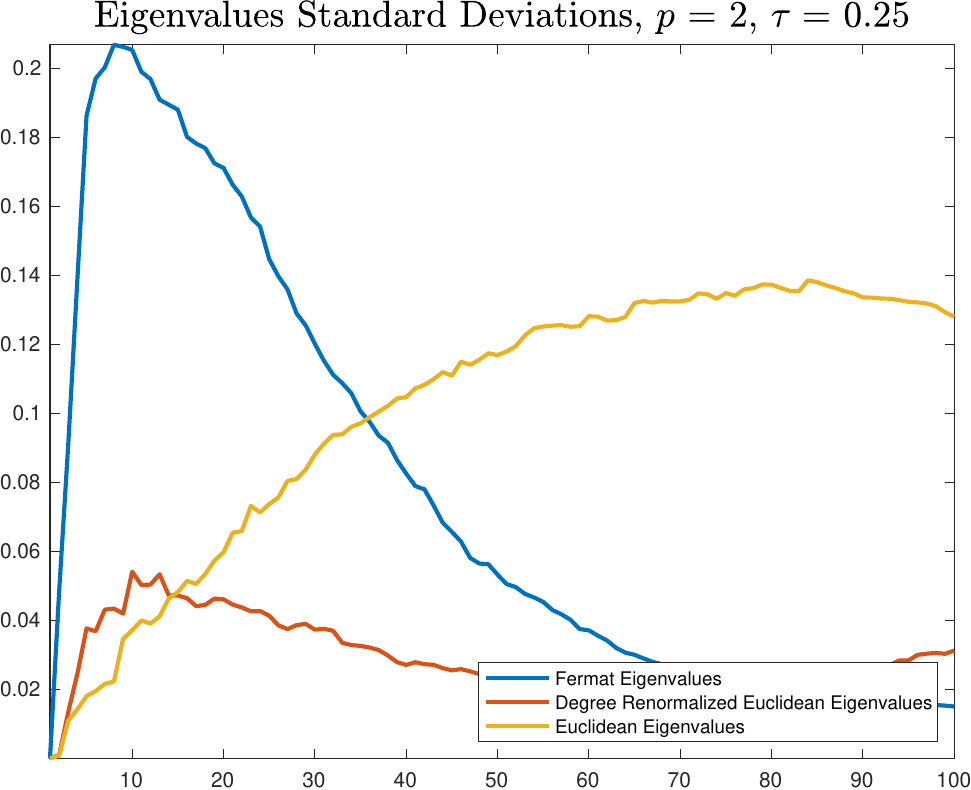}
\end{subfigure}
\begin{subfigure}[t]{0.32\textwidth}
		\centering
		\includegraphics[width=\textwidth]{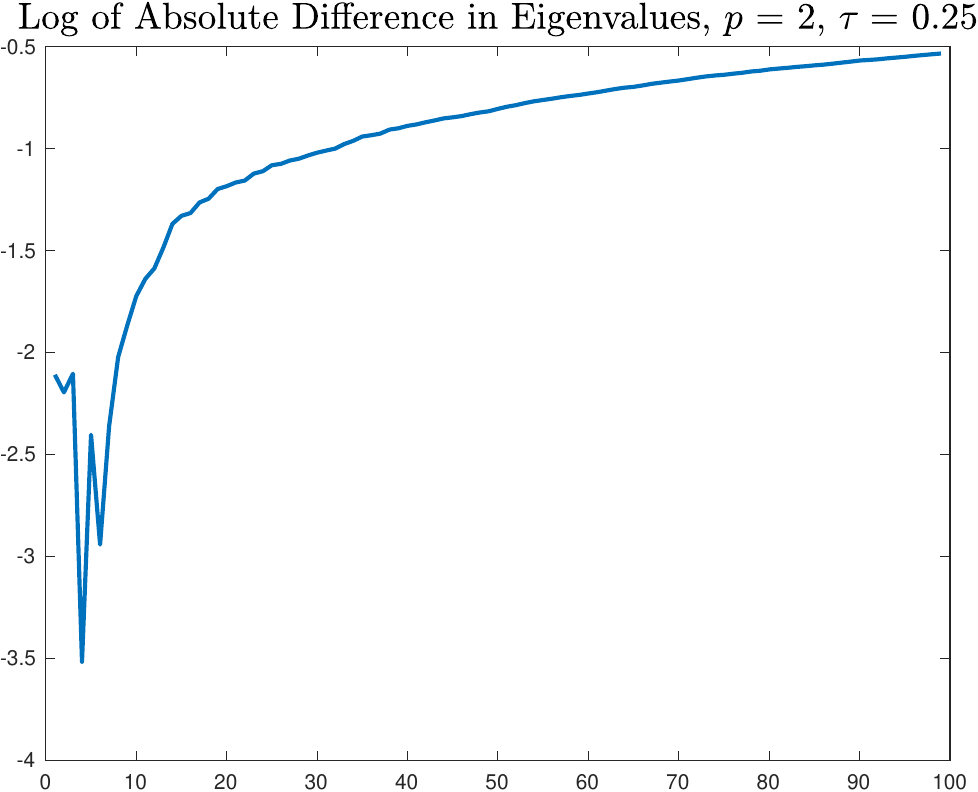}
\end{subfigure}
  \begin{subfigure}[t]{0.32\textwidth}
		\centering
		\includegraphics[width=\textwidth]{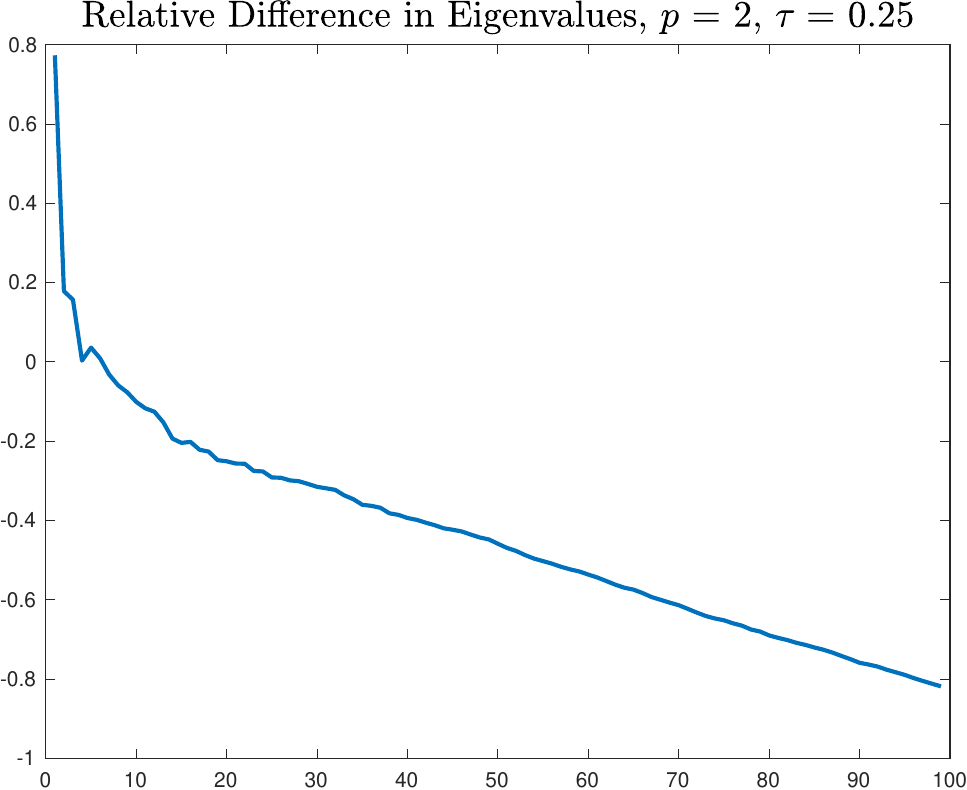}
\end{subfigure}
\begin{subfigure}[t]{0.32\textwidth}
		\centering
		\includegraphics[width=\textwidth]{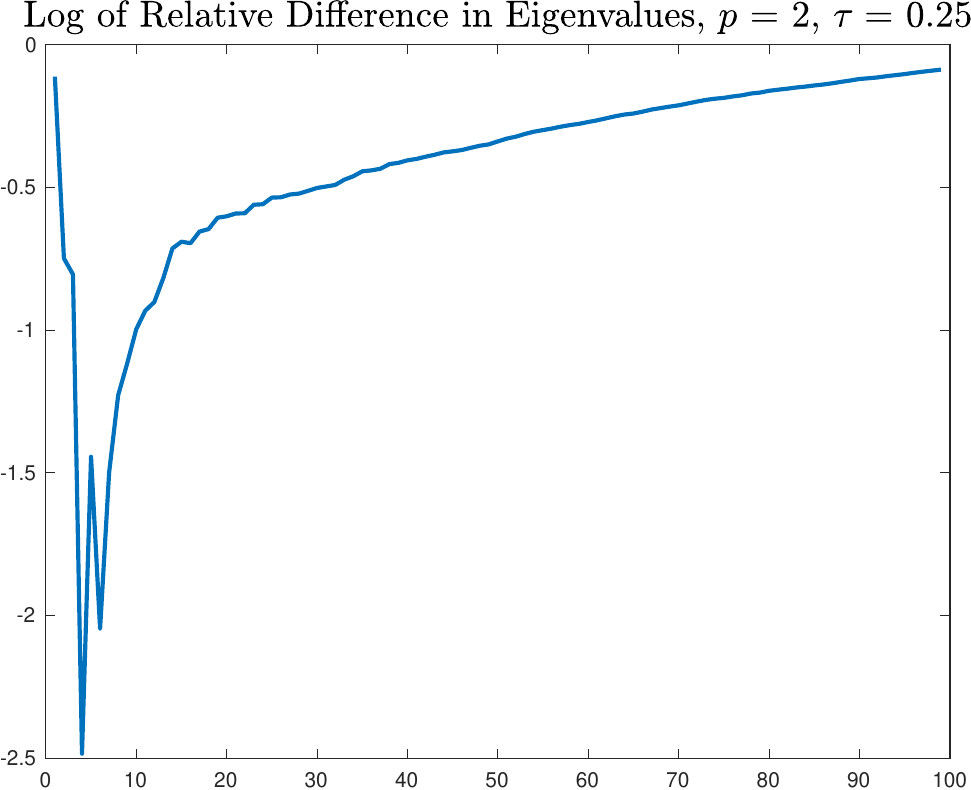}
\end{subfigure}
	\caption{$p=2$, $\tau=.25$.  Runtime for Fermat Laplacian: $219.76\pm19.41$s.  Runtime for Rescaled Euclidean Laplacian: $.60\pm.28$s.}
	\label{fig:GaussianMixture_p=2_thresh=.25}
\end{figure}

\end{document}